\documentclass{article} 
\usepackage{iclr2027_conference,times}

\usepackage{amsmath,amsfonts,bm}

\def\eqref#1{equation~\ref{#1}}

\def\1{\bm{1}}

\DeclareMathAlphabet{\mathsfit}{\encodingdefault}{\sfdefault}{m}{sl}
\SetMathAlphabet{\mathsfit}{bold}{\encodingdefault}{\sfdefault}{bx}{n}

\usepackage{hyperref}
\usepackage{url}
\usepackage[table]{xcolor}
\usepackage[textsize=tiny]{todonotes}
\usepackage{booktabs, pifont}
\usepackage[most]{tcolorbox}
\usepackage{wrapfig,lipsum,booktabs}
\tcbuselibrary{breakable}
\usepackage{makecell}

\newtcolorbox[auto counter]{promptbox}[1][]{
  enhanced, breakable,
  colback=blue!4!white, colframe=blue!45!black,
  coltitle=white, colbacktitle=blue!45!black,
  fonttitle=\bfseries, title={#1},
  fontupper=\scriptsize\ttfamily,
  boxrule=0.7pt, arc=2pt, outer arc=2pt,
  left=6pt, right=6pt, top=5pt, bottom=5pt,
}
\newtcolorbox[auto counter]{inputbox}[1][]{
  enhanced, breakable,
  colback=black!3!white, colframe=black!55,
  coltitle=white, colbacktitle=black!55,
  fonttitle=\bfseries, title={#1},
  fontupper=\scriptsize\ttfamily,
  boxrule=0.7pt, arc=2pt, outer arc=2pt,
  left=6pt, right=6pt, top=5pt, bottom=5pt,
}
\newtcolorbox[auto counter]{outputbox}[1][]{
  enhanced, breakable,
  colback=purple!4!white, colframe=purple!55!black,
  coltitle=white, colbacktitle=purple!55!black,
  fonttitle=\bfseries, title={#1},
  fontupper=\scriptsize\ttfamily,
  boxrule=0.7pt, arc=2pt, outer arc=2pt,
  left=6pt, right=6pt, top=5pt, bottom=5pt,
}

\makeatletter
\newcommand{\promptlabel}[1]{%
  \phantomsection
  \protected@edef\@currentlabel{\thetcbcounter}%
  \label{#1}%
}
\makeatother

\title{Training Reasoning Models to Generate Diverse Scientific Ideas}

\author{Manya Wadhwa$^{\spadesuit}$\thanks{Work done while at Salesforce AI Research.} , Pranav Narayanan Venkit$^{\diamondsuit}$, Hiroaki Hayashi$^{\diamondsuit}$, Yu Li$^{\diamondsuit}$,\\
\textbf{Tuhin Chakrabarty}$^{\clubsuit}$, \textbf{Chien-Sheng Wu}$^{\diamondsuit}$ \\
\\
$^{\spadesuit}$New York University, $^{\clubsuit}$Stony Brook University, $^{\diamondsuit}$Salesforce AI Research \\
\url{mw4141@nyu.edu}
}

\iclrfinalcopy 
\begin{document}

\maketitle

\begin{abstract}

One of the first steps of a research pipeline is generating ideas. Large Language Models (LLMs) are increasingly used to automate this step, but most of this work optimizes the quality of individual ideas, while overlooking diversity. In our work, we study document-grounded ideation, focusing on increasing the diversity of LLM generated ideas and ensuring they are high quality. Given prior papers, a model must propose a set of ideas that are each high-quality and clearly distinct from one another, while also being grounded in the input documents. To optimize for this, we propose training with a set-level reinforcement learning objective. In our setup, in a single generation, the model writes $K$ structured idea cards grounded in the source papers, and the set is scored \textit{jointly}, where each idea's quality is weighted by its distance from the ideas written before it. This teaches the model to steer each new idea away from its predecessors. We show that our trained 8B model achieves higher creative utility (a measure combining quality and diversity) than Claude Opus 4.8 and GPT-6 Astra on our contamination-free benchmark. In a pilot human study with four experts, our sets are judged as promising as Claude Opus 4.8's and contain marginally more unique ideas. When a coding agent runs proof-of-concept experiments on the ideas, our model's success@$k$ is comparable to Claude Opus 4.8's, so we see that the added diversity does not cost any executability\footnote{Code and data: \url{https://github.com/ManyaWadhwa/diverge_ideation}}.

\end{abstract}

\section{Introduction}
\label{sec:intro}   

The first step of any research cycle is generating ideas worth testing, before an experiment can tell you which one was right. Current work with Large Language Models (LLMs) on idea generation has focused on one of two things: producing a single idea and scoring it against an existing, held-out idea \citep{giants}, or building end-to-end systems that carry one idea through execution and judge whether that execution succeeded \citep{kirgis2026can,ning2026one,falck2026training,garikaparthi2026researchgym,si2601towards}. Either way, an idea is generated and judged on its own, never alongside other candidates it could be weighed against.

Looking at how humans do research, a researcher facing a body of literature brainstorms several directions off the same shared context, and then selects among them using some signal~\citep{platt1964strong,klahr1988dual}, rather than just committing to one idea. We draw a connection between ideation and creative problem solving. The brainstorm-then-select process is how creative thinking/problem solving can also be formalized, \textit{divergent thinking} explores a space broadly for many candidates, and \textit{convergent thinking} narrows it through selection \citep{boden2004creative, guilford1967nature}. The way current LLMs are used for ideation can look different. Because each idea is generated and judged on its own, multiple candidates come from repeated independent sampling, with no shared context to explore deliberately and no pool to converge over. Any diversity among them is incidental rather than designed \citep{multi-answer-rl,vpo}. Work on autonomous research agents has also flagged diversity as a shortcoming \citep{liu2026benchmarking,garikaparthi2026researchgym}. Also previous attempts at training an ideation model against execution signal has been shown to collapse onto a single dominant idea rather than a spread of them \citep{si2601towards}.

\begin{figure}[t]
    \centering
    \includegraphics[scale=0.26,trim=1cm 23cm 10cm 1cm]{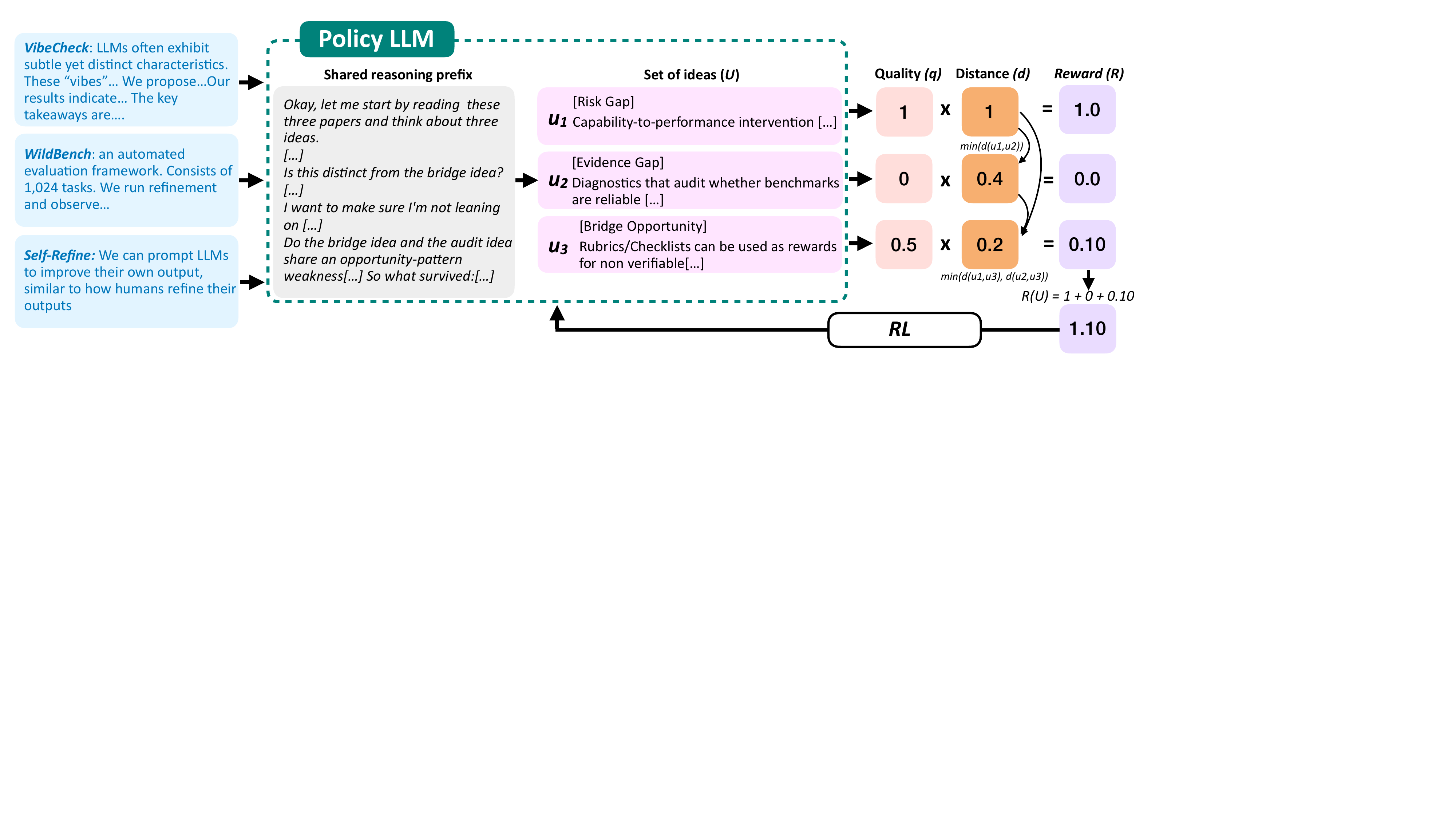}
    \caption{Overview of our training pipeline. Given a set of documents, the policy generates a set of ideas $U=\{u_1,\dots,u_k\}$; for reasoning models, the whole set is generated after a single shared reasoning prefix. Each idea $u_i$ is scored by its quality $q_i$ and its distance $d_i$ to the closest earlier idea ($d_1=1$), giving a per-item reward $r_i = q_i \cdot d_i$. The set reward $R(U)=\sum_i r_i$ is used as the RL signal to update the policy, encouraging sets that are both high-quality and diverse.}
    \label{fig:intro_fig}
\end{figure}

In our work, we close this gap by treating idea generation as a creative search problem. Given a corpus of prior documents, an agent should propose a set of $K$ ideas, each grounded in specific source documents, that are high-quality and differ from each other by design. We study this for document-grounded ideation \citep{wang2024scimon}. We combine quality and diversity into one set-level training signal by adapting the \textit{Creative Utility} formulation~\citep{wadhwa2026create} where each idea earns credit for its own quality and for its distance from the ideas emitted before it, so a redundant idea earns little no matter how well-written it is (as illustrated in Figure~\ref{fig:intro_fig}).

To achieve this, we train an ideation model (Qwen3-8B~\citep{qwen3technicalreport}) in two steps. We first do a supervised fine-tuning step to initialize the policy on well-formed idea sets, followed by GRPO~\citep{shao2024deepseekmath} against a set-level objective (adapted from the creative utility formulation). We then report this utility measure on a contamination-free test set that we put together. We also look at two existing benchmarks, GiantsBench \citep{giants} and IdeaSeed \citep{chen2026measuring}. Our tested baselines include the untrained base model and its SFT checkpoint, strong closed-source agents, GIANTS~\citep{giants}, and DARLING~\citep{li2025darling}, an alternative quality-diversity objective that we retrain in our setting.

We show that our 8B trained model achieves the highest creative utility on our test set, outperforming strong closed-source agents and competing quality-diversity objectives such as DARLING. A small pilot human study with four expert annotators finds that our sets are comparable to Claude Opus 4.8's in the number of promising ideas, while containing more mutually distinct ones. Finally, when we execute each set of ideas in a fixed sandbox with a coding agent, our model is comparable to Claude Opus 4.8 on success@$k$, showing that we are able to increase diversity without compromising important aspects of ideation like executability.

\textbf{Contributions} Our contributions are (1) a set-level RL objective for training that balances quality and diversity; (2) an instantiation of this objective for document-grounded ideation; and (3) an 8B trained ideation model that outperforms strong closed-source agents and models trained on competing objectives on creative utility, and whose ideas are comparably promising in a small pilot human and execution study.

\section{Training with a Set-Level Reward}
\label{sec:reward}
\subsection{Background}
\label{sec:background}
We start by describing the standard reinforcement learning setting for language models we build on. A policy parameterized as an LM $\pi_\theta$ maps a prompt $x$ to a distribution over textual outputs $y$. Given a reward function $R(x,y)\in\mathbb{R}$, training aims to maximize expected reward:
\begin{equation}
\label{eq:base_rl}
\max_\theta\ \mathbb{E}_{x\sim D,\ y\sim\pi_\theta(\cdot\mid x)}\big[R(x,y)\big].
\end{equation}
We optimize Equation~\ref{eq:base_rl} with Group Relative Policy Optimization (GRPO) \citep{shao2024deepseekmath}: for each $x$, we sample $G$ rollouts $\{y_1,\dots,y_G\}\sim\pi_\theta(\cdot\mid x)$, score each with $R(x,y_g)$, and update the policy using each rollout's advantage relative to the group's mean reward, with no learned critic.

\subsection{Set-Level objective} 
\label{sec:general_objective}

In our setting, each prompt $x$ is answered by a set of $K$ outputs $U=\{u_1,\dots,u_K\}$ written in a single generation, after a shared reasoning prefix. This set is the rollout $y$ in Eq.~\ref{eq:base_rl}, so GRPO scores it with one scalar and computes one advantage per set. All $K$ outputs therefore share that single credit signal. Our goal is a set that is high-quality and whose outputs are mutually distinct. Then we train the model using a \textit{set-level} objective using the definition below:
\begin{equation}
\label{eq:reward}
R_{set}(x,U)=\sum_{i=1}^{K} q(u_i)\cdot{\min_{j<i} d(u_i,u_j)}, \qquad \min_{j<i} d(u_i,u_j) := 1 \ \text{if } i=1
\end{equation}

For each item in the generation, we multiply its quality by how far it is from its nearest predecessor and aggregate over the set to get the reward. A redundant but high quality item earns almost nothing (Fig~\ref{fig:intro_fig}).

The objective has two components: a quality function $q(u_i)\in[0,1]$ that scores an output for the prompt, and a distance $d(\cdot,\cdot)\in[0,1]$ between two outputs. Any task with a quality signal and a notion of diversity fits, such as instruction following, creative writing, or ideation; we instantiate both for document-grounded ideation in Section~\ref{sec:instantiation}.  


\textbf{Relation to Creative Utility.} Eq.~\ref{eq:reward} is adapted from Creative Utility (CU) \citep{wadhwa2026create}, which scores a set of ideas by the quality each one adds beyond the ideas already selected. We make two design choices in adapting it to a training reward. First, CU applies a patience discount ($\gamma$) to later ideas; we drop it for training, so every idea in the set is weighted equally and a model cannot fall back on generating weak ideas later in the set. Second, CU takes the best ordering of the set; we instead use the order in which the model writes them. Our prompt asks the model to list ideas from most to least promising, and it lets the reward be computed directly from a single rollout. At evaluation we report CU as is (Section~\ref{sec:experiments}). 


\textbf{Relation to multi-answer RL.} Our proposed objective can be viewed as a generalized form of the set-level objective introduced by \citet{multi-answer-rl}. Their formulation however has no notion of distance and instead only checks for coverage against an established ground truth answer set. Their objective can be defined as a special case of Eq~\ref{eq:reward}, by setting $d(u_i, u_j)=1$ and $q$ to a coverage metric.



\section{Document Grounded Ideation}
\label{sec:ideation}
As mentioned in Section~\ref{sec:intro}, we study the task of document-grounded ideation i.e. given a body of related literature, an agent needs to generate a set of ideas worth pursuing next, and we evaluate the set of ideas for its quality and diversity.

\textbf{Task definition.} An instance is a corpus of documents $D = \{d_1, \dots, d_n\}$ (each a scientific paper, given as a summary). An agent must produce a set of $K$ ideas $U = \{u_1, \dots, u_K\}$, where each idea $u_i$ is a structured \emph{idea card} with four fields: a \textit{problem gap}, a \textit{core claim}, a \textit{method}, and an \textit{inspiration} field that names the specific source documents the idea draws on and how.

\textbf{Why a set of $K$?}  We make the set the primary unit, generating all $K$ ideas in one pass so each is written conditioned on those already emitted. The agent can actively avoid repeating itself, steering later ideas toward directions its earlier ones left uncovered, rather than relying on independent draws to spread apart by chance \citep{multi-answer-rl, vpo,zhang2026verbalized}.



See Appendix \ref{app:data_format} for definitions of the idea card sub fields and examples of ideas.

\subsection{Reward Instantiation}
\label{sec:instantiation}
We instantiate quality ($q$) and distance($d$) in Equation \ref{eq:reward} for document-grounded ideation as follows:

\textbf{Quality:} A part of ideation is to come up with directions that are worth pursuing next, so we define quality as long-term scientific impact \citep{tong2026ai}. While there are other properties of \textit{good} ideas such as fluency or soundness, these are not sufficient to decide if the idea is worth pursuing next. We therefore treat soundness as a separate axis, which we only probe with a small execution check (Section~\ref{sec:autoresearch}), rather than the training target. We anchor long-term impact to the gold idea $g$, a reference idea whose value is already established, so the judgment reflects what the field found worth doing. Section ~\ref{sec:data} describes how we obtain $g$. We model impact as a pairwise judgment between the generated idea $u$ and the gold idea $g$:


\begin{equation}
\label{eq:quality}
q(u)=I(u,g)=
\begin{cases}
1 & \text{if } u \succ g,\\
0.5 & \text{if } u \sim g,\\
0 & \text{if } g \succ u.
\end{cases}
\end{equation}
where $I(u,g)$ is the pairwise impact judgment, and $u \succ g$ means the judge deems $u$ higher-impact than $g$, and $u \sim g$ means it cannot separate them. 



\textbf{Distance.} Distance captures whether two ideas represent different directions. Two ideas can share a problem while having completely different methods. We operationalize this metric by scoring semantic distance field-wise over the \textit{problem gap}, \textit{claim}, and \textit{method} fields, using an LLM-judged similarity rubric, and averaging with equal weight across the three fields. 
\begin{equation}
\label{eq:distance}
d(u_i,u_j)=\tfrac{1}{3}\big(d_{\text{problem}}(u_i,u_j) +d_{\text{claim}}(u_i,u_j)+d_{\text{method}}(u_i,u_j)\big),
\end{equation}
where each $d_{\text{field}}\in[0,1]$ is a per-field distance derived from a 1--5 LLM-judged similarity rubric and normalized to be in the $[0,1]$ range.

For impact reward we use an off-the-shelf reward model, \texttt{SciJudge-30B}~\citep{tong2026ai} and for distance we use \texttt{gpt-oss-120b}~\citep{openai2025gptoss120bgptoss20bmodel}. Details around the prompts and human validation are in the Appendix~\ref{app:metrics}.


\subsection{Data}
\label{sec:data}
We build a training set and a contamination-free test set from published papers. Our data construction pipeline has two steps: (a) gathering the source and target documents (b) generating data for supervised fine-tuning. Each instance pairs a \emph{target} paper (or gold idea) with its \emph{source documents}.

\textbf{Gathering documents.} We take each paper in the training split of GiantsBench \citep{giants} as a target and retrieve its full text and citation graph, including the in-text context of each citation. An LLM (\texttt{claude-sonnet-4-6}) selects the source documents from the target's references, keeping those that motivated the paper, exposed the gap it addresses, supplied its conceptual foundation, or were directly extended \citep{cohan2019structural}. Each source is summarized into a source card in the idea-card schema, and the cards form the input corpus $D$. The target's text is summarized into an idea card and held out as the gold idea $g$. We drop instances with two or fewer source documents or no gold idea. For training we focus on the Language domain. The test set is built using the same pipeline but with target papers 2026 (Section\ref{sec:experiments}). The test split also spans 16 domains (distribution in Figure~\ref{fig:domain_distribution}, Appendix~\ref{app:data}).


\textbf{SFT data construction} The SFT stage teaches the idea-card format and a thinking pattern. A teacher LLM (\texttt{claude-opus-4.8}) first proposes $K$ ideas from the source documents alone. Because we cannot access a closed model's reasoning, we then use \emph{hindsight} prompting, where the teacher sees the sources and its own ideas and writes a first-person exploration that arrives at exactly those ideas, floating and rejecting candidates, hedging over the sources and raising questions. Each instance gets a different opening statement so traces do not collapse into one template. We train on 7,172 instances and use $1{,}740$ for testing. Appendix~\ref{app:data} has details about the train and test splits.


\subsection{Training Details}  Our proposed training setup has two steps: (a) light supervised fine-tuning with next token prediction to teach the model the format of our task as well as the thinking pattern, and, (b) GRPO as the base RL algorithm. We use Qwen3-8B~\citep{qwen3technicalreport} as the base model with a maximum response length of 16k tokens, and we set $K=3$. We repeat these two steps for two setups of training: (i) \textit{with think}: where we optimize for the thinking traces that we constructed in the previous section and (ii) \textit{no think}: where we only optimize for the final answer block and have an empty think block.

We also show the recipe generalization on Olmo-3-7B-Think-SFT~\citep{olmo2025olmo3}. The hyperparameters, SFT variants, Olmo-3 details and compute details are in Appendix~\ref{app:training}.

\section{Experimental Setup}
\label{sec:experiments}

\textbf{Baselines.} We compare a model trained with our proposed objective against four groups of systems. 

\emph{General models.} We evaluate strong closed-source models (Claude Opus 4.8 and GPT-6 Astra) and a representative open-source model (Nemotron-120B~\citep{nvidia_nemotron_3_2025}), prompted zero-shot with our set-generation prompt. 

\emph{Competing objectives.} \textbf{DARLING}~\citep{li2025darling} is the closest approach to ours that optimizes quality and diversity together, but across the rollouts of a single-idea policy rather than within the generated set. We replicate the algorithm for our task and train it with $K=1$, so the only difference from our method is where diversity is enforced (Appendix~\ref{app:darling}). \textbf{Impact-only RL} uses our set-level RL setup with the diversity term removed (i.e., $d=1$ in Eq.~\ref{eq:reward}), and isolates what the diversity term contributes. \textbf{Impact+diversity, no SFT} uses our full reward but starts GRPO directly from the base model, and isolates what the SFT warm-start contributes. These are trained from the same base model and training corpus as our method.

\emph{Base and SFT.} \textbf{Qwen3-8B (base)} is the untrained base model, prompted for $K$ ideas directly. \textbf{SFT} is the checkpoint after supervised initialization and before GRPO. Both are the starting points for our method.

\emph{Proposed.} \textbf{Proposed} is SFT followed by GRPO with our set-level reward (Eq.~\ref{eq:reward}). For SFT and Proposed we report two variants: \emph{think} and \emph{no think}.

\textbf{Benchmarks.} We evaluate the above on three datasets.
\begin{enumerate}
    \item \textbf{Ours (contamination-free).} To measure generalization free of contamination (target papers from 2026), we evaluate on the test-set as explained in Section~\ref{sec:data}. We have $1{,}740$ evaluation instances across 16 domains. 
    \item \textbf{GiantsBench} \citep{giants}. The input for this dataset is two documents. We evaluate on its held-out test split which has 2{,}374 instances and is disjoint from the training split we use for our data (Section~\ref{sec:data}). We compare our model against GIANTS-4B, which was trained on the GiantsBench train split (Appendix~\ref{app:benchmarks}). 
    \item \textbf{IdeaSeed} \citep{chen2026measuring}. A dataset of 11k pairs of published papers, each with a human-written proposal and its related-work context, which acts as input. We use this benchmark to test whether a trained model keeps the breadth of idea types that the research community pursues. We report this on a sample of 500 instances (Appendix~\ref{app:benchmarks}).
\end{enumerate}

\textbf{Metrics.}  Our primary metric is \textbf{Creative Utility (CU)}~\citep{wadhwa2026create}: 
\begin{equation}
\mathrm{CU}(U)=\max_{\tau}\;\frac{\sum_{i=1}^{|U|}\gamma^{\,i-1}\,q\big(u_{\tau(i)}\big)\min_{j<i}d\big(u_{\tau(i)},u_{\tau(j)}\big)}{\sum_{i=1}^{|U|}\gamma^{\,i-1}},
\end{equation}

Here $\tau$ ranges over orderings of the set $U$, $q$ is quality, $d$ is distance (with $\min_{j<1}d:=1$), and $\gamma\in(0,1]$ is a patience discount on later ideas. The denominator normalizes CU to $[0,1]$. The maximum over orderings is combinatorial, so, following ~\cite{wadhwa2026create}, we evaluate it with the greedy order. We use $\gamma=0.9$ and $K=3$.

We also report two components of the reward separately. \textbf{Quality ($q$)} is the average length-controlled win rate~\citep{dubois2404length} of an idea against the gold idea (Eq.~\ref{eq:quality}) as a strict win-rate, judged with \texttt{GPT-5.4-mini} (Appendix~\ref{app:metrics-quality}). \textbf{Distance ($d$)} is the mean pairwise distance $d(u_i,u_j)$ within the set (Eq.~\ref{eq:distance}), judged with \texttt{gemini-3.5-flash}. We  also report \textbf{grounding} ($g$), an LLM judge's assessment of how crucial and faithful each idea's declared source support is to what the documents actually claim. Grounding is judged with \texttt{gemini-3.5-flash}. 

For \textbf{IdeaSeed}, we compare the model's population-level idea-type distribution against the human idea-type distribution, per axis. We report (i) \textbf{normalized entropy $H_{norm}$ } and (ii) \textbf{total variation distance} (TVD).

We detail the prompts and inference parameters for all LLM judges in Appendix~\ref{app:metrics}.

\begin{table}[t]
\centering
\small
\caption{Results on our test set. We report Creative Utility ($CU$) with $\gamma=0.9$. Win Rate is the per-idea impact against the gold idea, corrected for length. Avg.\ Distance is the mean pairwise distance between the ideas in a set. Values in parentheses are standard deviations across instances. Proposed (no think) has the highest creative utility; Proposed (think) is on par with DARLING and the no-SFT ablation.}
\label{tab:final-si-n1-main}
\begin{tabular}{l|c|cc}
\toprule
Model & Creative Utility (CU) $\uparrow$ & LC Win Rate ($q$) $\uparrow$ & Avg. Distance ($d$) $\uparrow$ \\
\midrule
\multicolumn{4}{c}{General models} \\
\midrule
Claude Opus 4.8 & $0.38_{(0.22)}$ & 0.21 & $0.80_{(0.07)}$ \\
\rowcolor{gray!20}
GPT-6 Astra & $0.28_{(0.24)}$ & 0.15 & $0.86_{(0.06)}$ \\
Nemotron 120B & $0.14_{(0.18)}$ & 0.05 & $0.79_{(0.10)}$ \\
\midrule
\addlinespace[2pt]
\multicolumn{4}{c}{Competing objectives} \\
\midrule
DARLING & $0.52_{(0.17)}$ & 0.38 & $0.67_{(0.13)}$ \\
\rowcolor{gray!20}
Impact-only RL & $0.38_{(0.18)}$ & 0.43 & $0.38_{(0.17)}$ \\
Impact+diversity, no SFT & $0.51_{(0.26)}$ & 0.30 & $\mathbf{0.92_{(0.07)}}$ \\
\addlinespace[2pt]
\midrule
\multicolumn{4}{c}{Base and SFT} \\
\midrule
Qwen3-8B & $0.03_{(0.10)}$ & 0.01 & $0.80_{(0.11)}$ \\
\rowcolor{gray!20}
SFT (no think) & $0.21_{(0.19)}$ & 0.07 & $0.77_{(0.09)}$ \\
SFT (think) & $0.16_{(0.18)}$ & 0.06 & $0.76_{(0.10)}$ \\
\midrule
\addlinespace[2pt]
\multicolumn{4}{c}{Proposed (SFT + RL)} \\
\midrule
\rowcolor{blue!10}
Proposed (no think) & $\mathbf{0.64_{(0.23)}}$ & \textbf{0.50} & $0.88_{(0.07)}$ \\
\rowcolor{blue!10}
Proposed (think) & $0.51_{(0.23)}$ & 0.34 & $0.83_{(0.08)}$ \\
\bottomrule
\end{tabular}
\vspace{-1em}
\end{table}

\section{Results}
\label{sec:main_results}

\textbf{Proposed (no think) achieves the highest Creative Utility}  Table~\ref{tab:final-si-n1-main}, shows that the proposed model trained with the set-level reward and no thinking, \textit{Proposed (no think)}, achieves the highest creative utility and win-rate, with high distance between ideas. This utility increases from 0.03 (base model), to 0.21 (SFT no think) to 0.64 post our proposed training (no think). We also beat state of the art closed source models, where Claude Opus 4.8 achieves a CU of only 0.38 and GPT-6 Astra achieves 0.28. The thinking variant of our proposed training, \textit{Proposed (think)} achieves a utility of 0.51, which still beats both closed-source models but is lower than the no-think variant and on par with DARLING (0.52) and the no-SFT ablation (0.51). We discuss this trade-off at the end of this section. The high creative utility is because \textit{both} impact against gold (win-rate) and idea diversity are high for our proposed model. Full table in Appendix~\ref{app:results}.

\textbf{Proposed training recipe beats different competing objectives} Training with a set level objective for only quality (\textit{Impact-only RL}) gives a good win-rate (43\%), but it collapses the set. Its average distance falls to 0.38. This shows that training for quality only in a set can get us good ideas, but the model ends up focusing on only one good instead of trying to get \textit{multiple} good ideas. For DARLING, we see that it reaches a utility of 0.52 against 0.64 for Proposed (no think), but has a lower distance (0.67 against 0.88). Showing that rewarding within-set generation for diversity is more effective than across-rollout diversity. 

\begin{wrapfigure}{r}{0.5\textwidth}
\centering
\includegraphics[scale=0.25, trim= 0 18cm 18cm 0]{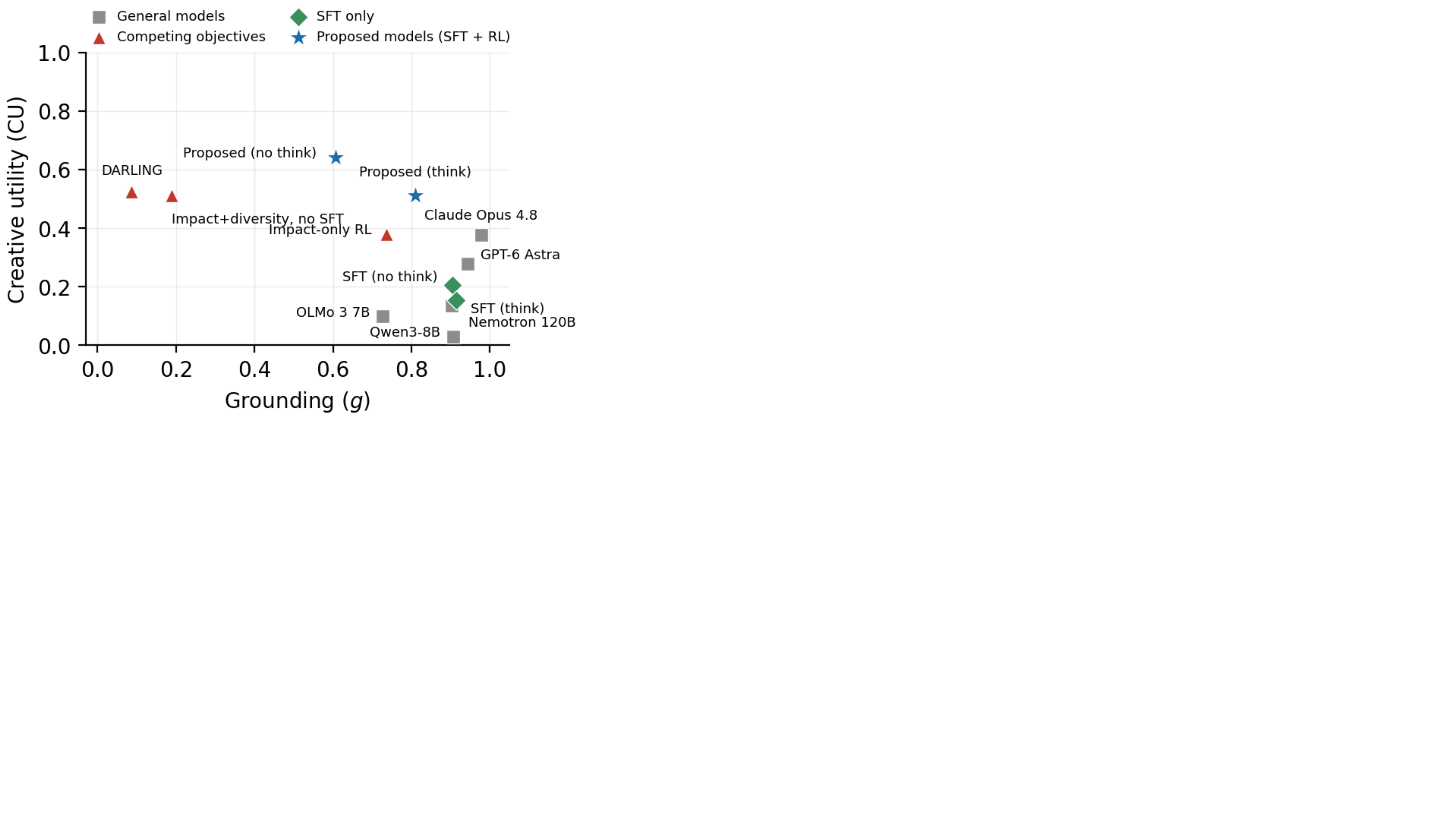}
\caption{Creative Utility (CU) against average grounding ($g$) for the systems also mentioned in Table~\ref{tab:final-si-n1-main}. The highest-CU systems are less grounded than Claude and the SFT-only models.}
\label{fig:grounding}
\end{wrapfigure}

\textbf{SFT keeps the model grounded} To test the SFT warm-start, we train the same base model with our objective but skip SFT (\textit{Impact+diversity, no SFT}). This baseline matches Proposed (think) at 0.51 CU, but does worse on grounding as seen in Figure~\ref{fig:grounding}. Grounding for the no SFT baseline falls to $g=0.19$, against 0.81 with SFT (Figure~\ref{fig:grounding}). Without the warm-start, the policy gets its diversity by drifting away from its source documents. It also stays 0.13 CU below Proposed (no think). So SFT is what lets the set-level reward raise utility without losing grounding.

\textbf{Thinking helps retain grounding} Proposed (think) has lower utility than Proposed (no think) (0.51 vs.\ 0.64). Primary reason for this is a lower win rate (0.34 vs.\ 0.50), but we see that thinking gets more grounding (0.81 vs.\ 0.61; Figure~\ref{fig:grounding}). The grounding gap appears only after RL, since SFT-only think and no-think models are equally grounded (0.91 vs.\ 0.90). Thinking thus seems to help the policy \emph{retain} grounding as RL pushes for utility. We hypothesize two reasons why thinking may not raise utility. First, the reward scores only the final ideas, and the impact signal is a judged comparison against a gold idea that cannot be verified at intermediate steps. Thus the steps might not have enough causal influence on the reward~\citep{sprague2025to}. Second, designing thinking traces suitable for creative ideation is an open-problem. In our work, we tried one approach that was hindsight generation. While we instruct the teacher model to propose negatives, verify etc. there may be instances where there is no `search' for candidates beyond what the model generated originally; we also tried distilling forward-thinking traces but it showed the same trend (Appendix~\ref{app:results}). We leave the problem of designing synthetic or appropriate thinking traces for creative/open-ended problems to future work.

\begin{wraptable}{r}{0.4\textwidth}
  \centering
  \small
  \renewcommand{\tabcolsep}{1mm}
  \caption{GiantsBench results on 2{,}374 test papers. }
  \begin{tabular}{lcc}
    \toprule
    Model & Win Rate $\uparrow$ & Avg.\ distance $\uparrow$ \\
    \midrule
    GIANTS-4B  & 18.3\% & 0.025 \\
    Proposed & \textbf{68.3\%} & \textbf{0.939} \\
    \bottomrule
  \end{tabular}
  \label{tab:giantsbench}
\end{wraptable}

\textbf{We beat Giants-4B on GiantsBench} Table~\ref{tab:giantsbench} shows the length-controlled win-rate and the average distance on GiantsBench. Our model achieves a win-rate of 68.3\% against the gold idea, compared to 18.3\% for GIANTS-4B. We want to note that GIANTS was trained on a `similarity'  to gold-idea metric, which does not lead to the model generating ideas better than the gold idea (18.3\% win rate) and also leads to collapse of ideas. GIANTS-4B's average pairwise distance is 0.025, against 0.939 for ours, since sampling it three times yields near-identical restatements of the same insight.

\begin{wraptable}{l}{0.5\textwidth}
  \centering
  \small
  \setlength{\tabcolsep}{6pt}
  \caption{Taxonomy diversity on IdeaSeed (500x3 ideas), Method Paradigm
  axis. 
  }
  \begin{tabular}{lcc}
    \toprule
    Method & $H_{\mathrm{norm}}$ $\uparrow$ & TVD vs human $\downarrow$ \\
    \midrule
    Qwen3-8B & 0.77 & 0.27 \\
    Claude Opus 4.8 & 0.77 & 0.23 \\
    DARLING & 0.01 & 0.72 \\ 
    \midrule
    Proposed (no think) & 0.65 & 0.43 \\
    Proposed (think) & 0.73 & 0.30 \\
    \midrule
    Human proposals & 0.87 & -- \\
    \bottomrule
  \end{tabular}
  \label{tab:ideaseed-entropy-main}
\end{wraptable}

\textbf{IdeaSeed: quality gains without type collapse} In Table~\ref{tab:ideaseed-entropy-main}, we report population level entropy metrics on the `method' axis. Every model is more concentrated than human proposals ($H_{\mathrm{norm}}=0.87$). We see that after training, our models do not collapse at the population level. Proposed (think) reaches $0.73$ (TVD $0.30$), close to Claude Opus 4.8 and Qwen3-8B ($0.77$), while Proposed (no think) is lower at $0.65$ (TVD $0.43$). The prompted baselines are somewhat ahead of us on this axis, which suggests RL may have cost some population level type diversity, though our thinking proposal recovers much of it. Since our objective targets diversity within each set and does not directly optimize the spread of idea types across instances, improving population-level diversity is an open direction for future work. DARLING on the other hand collapses onto a single Evidence Gap / Empirical Mapping pair. Results on the Opportunity axis and the per-label distributions are in Appendix~\ref{app:results}.

Appendix~\ref{app:results} reports how the proposed recipe generalizes to a different model family, per-domain results, and how win rate and diversity change with idea position.

\section{Downstream Validation of Idea Sets}
To validate that the ideas generated by our model are actually promising for researchers, we conduct two experiments: (a) human study with domain experts and, (b) execution-based evaluation using coding agents.

\subsection{Human Study: Portfolio Quality and Diversity} \label{sec:userstudy}

\label{sec:human}
We conduct a pilot user study to evaluate if the ideas are usable, promising and unique to humans. To recruit users, we set a minimum criterion that they have a PhD in the field of NLP and at least four published academic papers in top-tier conferences. We onboard four participants for the task, that meet our established criteria. Each user gives us a preference for the topics within NLP/ML they are comfortable with (self-declared expertise), and we match them with a subset of queries from our test set that fall under those topics. Each user gets assigned 15 instances and they blind evaluate sets of ideas from three systems: Claude Opus 4.8, Qwen3-8B and our Proposed (no think) model. Overall, each user gives judgments on 45 sets of ideas. 

 For each set of ideas, our experts make three judgments: (1) which ideas are worth pursuing as early-stage research directions (\textit{promising}); (2) ignoring quality, the largest subset of ideas that pursue substantively different directions (\textit{unique}); and (3) the same subset, restricted to ideas marked worth pursuing (\textit{Unique Promising}). Two ideas count as the same direction when carrying out one would largely address the other. Generally, a higher score is better for all these metrics.

\begin{wraptable}{r}{0.6\textwidth}
  \centering
  \small
  \renewcommand{\tabcolsep}{1mm}
   \caption{Human study results with 4 annotators, 15 papers each. We see our proposed model generates ideas that are as promising and diverse as Claude-Opus-4.8. Proposed is significantly better than Qwen3-8B on all three metrics.}
  \begin{tabular}{lccc}
    \toprule
    System & Promising & Unique & Unique Promising \\
    \midrule
Qwen3-8B & $1.52_{0.83}$ & $2.52_{0.62}$ & $1.42_{0.81}$ \\
Claude Opus 4.8 & $\mathbf{2.20_{0.73}}$ & $2.72_{0.52}$ & $2.03_{0.76}$ \\
Proposed & $2.17_{0.72}$ & $\mathbf{2.83_{0.38}}$ & $\mathbf{2.05_{0.75}}$ \\
    \bottomrule
  \end{tabular}
  \label{tab:human}
\end{wraptable}

\paragraph{Results} Table~\ref{tab:human} reports aggregate judgments across 4 annotators and 60 instances. Our model's portfolios contain 2.17 promising ideas and 2.83 unique ideas on average, against 2.20 and 2.72 for Claude Opus 4.8 and 1.52 and 2.52 for Qwen3-8B. Combining these, we see that our proposed model and Claude Opus 4.8 are considered similarly promising.  Appendix~\ref{app:results} breaks these results down by topics of the target paper and users.

\subsection{Can ideas be executed?} 
\label{sec:autoresearch}

Having established the quality and diversity of ideas in the previous sections, we check if the generated ideas hold up when carried towards execution. This is like a held-out quality check for the model trained on the impact signal. To test this, we take 15 instances and execute sets of 3 ideas from Claude Opus 4.8 and our no-think and think models. Every idea frozen into a card independently with the source hidden and handed to a Claude Code agent (Claude Opus 5) in a fixed sandbox with a 90-minute, \$30 budget and no GPU, identical for every idea. The agent designs and runs a proof-of-concept test for early signal on the claim, then self-reports one verdict: \emph{supports}, \emph{refutes} or, \emph{inconclusive}. \emph{Supports} requires a positive result covering most or all of the claim. 

\begin{wraptable}{l}{0.4\textwidth}
\centering
\small
\renewcommand{\tabcolsep}{1mm}
\caption{Execution results of 132 ideas by a coding agent.}
\begin{tabular}{lrrr}
\toprule
Arm & $k=1$ & $k=2$ & $k=3$ \\
\midrule
Claude Opus-4.8 & 13\% & 20\% & 27\% \\
Proposed (no think) & 11\% & 22\% & 33\% \\
Proposed (think) & 10\% & 19\% & 29\% \\
\bottomrule
\end{tabular}
\label{tab:autoresearch}
\end{wraptable}

Table~\ref{tab:autoresearch} reports success@$k$. For $k=3$ it is the fraction of instances (15; 14 for think) for which at least one of the 3 ideas from a system gets a \emph{supports} verdict from the agent. For $k<3$ we average that indicator over all size-$k$ subsets of the instance's 3 ideas, then over instances. Our models are comparable to Claude Opus 4.8 at every $k$. We find no evidence of a quality loss relative to Claude Opus 4.8. All success rates increase with $k$, so a set of ideas is more useful than a single one. 

\textbf{What a run looks like.} Given a frozen card, the agent first turns the claim's properties into concrete measurements: it picks datasets, controls and thresholds. It then runs a first proof of concept, and revises the plan after seeing its own results. In the example in Appendix~\ref{app:autoresearch}, we show a memory-benchmark idea from the think model. The first run showed that one planned probe could not support the claim's
``shared failure'' conjunct. The agent replaced the probe with three pre-registered ones and fixed its decision rules before running them. The final verdict is \emph{supports}. A qualitative read of this run suggests the verdict is somewhat stronger than the evidence in places (Appendix~\ref{app:autoresearch}), so we treat verdicts as early signals for a researcher to follow up rather than as findings. Table~\ref{tab:autoresearch_examples} shows additional examples of claims, verdicts along with the initial plan tried by the coding agent. 

\section{Related Work}

\paragraph{Quality and diversity in LLMs.} A large body of work has quantified the quality-diversity trade-off in LLMs, across creative writing, open-ended generation, and idea generation \citep{zhang2025noveltybench,shypula2025evaluating,chen2026measuring,wadhwa2026create,padmakumar2024does,jiang2026artificial}. A common finding is that RL and preference tuning collapse the policy onto a few modes, trading diversity for quality \citep{kirk2024understanding,cui2025entropy}. Training-time fixes reward diversity across independent rollouts of a prompt \citep{li2025darling}, or, in preference optimization, select diverse preference pairs \citep{lanchantin2025diverse}. More recent work rewards a \emph{set} of outputs directly, either to cover a verifiable answer set \citep{multi-answer-rl} or to trade off a fixed reward vector \citep{vpo}. All of these need an answer set or a fixed reward space to define coverage. In open-ended ideation neither exists, since whether two ideas are distinct is itself a judgment made in the context of the others in the set. Our objective addresses this by scoring each idea against the ones already written in the same generation.

\paragraph{Research ideation.} Literature-grounded ideation systems generally start from a small seed and build ideas outward from it.  SciMON~\citep{wang2024scimon} revises an idea until it is sufficiently novel relative to prior papers, ResearchAgent~\citep{baek2025researchagent} refines a proposal through rounds of reviewer feedback, and Nova generates hundreds of candidates from a seed paper and clusters them to select a diverse final set \citep{hu2024nova}. Revision, critique, and clustering are all ways to obtain diversity, but they act \emph{after} generation on candidates written without regard to one another, and the generator is never trained to write ideas that differ from each other. On the other hand, GIANTS~\citep{giants} trains a model with RL to anticipate a downstream paper's core insight from its parent papers, and \citet{tong2026ai} train a judge and an ideator on community feedback such as citations, both rewarding one idea at a time.

A separate line grounds ideation in execution, using the outcomes of implementing and running ideas as the reward. \citet{si2601towards} find that reinforcement learning from execution signal suffers from mode collapse: the model raises its average reward by converging on a few simple ideas. Because each idea is rewarded in isolation, nothing discourages repeating the same safe idea. We differ on both axes. Our input is a whole body of documents instead of a single seed, and the model writes a set of ideas in one generation, trained on a reward in which each idea is scored against the ones already written. We also use execution as an evaluation, checking whether sets of ideas succeed downstream (Section~\ref{sec:autoresearch}).

\section{Conclusion}

We present a set-level RL objective for document-grounded ideation for training to generate diverse, high quality ideas. A model is prompted/taught to generate $K$ ideas in one rollout and each idea's quality is weighted by its distance from the ideas before it. Our best 8B model (no think) achieves higher creative utility than strong closed-source models and competing quality-diversity objectives, while the think variant beats the closed-source models and is on par with DARLING. In small pilot human and execution studies its sets are comparable to Claude Opus 4.8's, so we see no evidence that diversity costs quality. Future work includes grounding-aware rewards, diversity across a population, and reasoning traces designed for creative ideation.

\section*{Limitations} Our study currently focuses on scientific ideation based on context and parametric knowledge, instead of focusing on retrieval. This is partially because we are testing the set up of training with a new set style generation where the aim is to increase diversity while also improving the quality. In future work we aim to include retrieval in this ideation process. We also train on english only, natural language processing scientific papers.

\section*{Acknowledgments}
This work was done while at an internship at Salesforce AI Research. Thanks to TAUR Lab at NYU for review and feedback.

\bibliography{iclr2027_conference}

@article{multi-answer-rl,
  title={{Reaching Beyond the Mode: Reinforcement Learning for Distributional Reasoning in Language Models}},
  author={Puri, Isha and Damani, Mehul and Shenfeld, Idan and Ghassemi, Marzyeh and Andreas, Jacob and Kim, Yoon},
  year={2026},
  journal={arXiv preprint arXiv:2603.24844},
  eprint={2603.24844},
  archivePrefix={arXiv},
  primaryClass={cs.LG}
}

@article{vpo,
  title={{Vector Policy Optimization: Training for Diversity Improves Test-Time Search}},
author={Ryan Bahlous-Boldi and Isha Puri and Idan Shenfeld and Akarsh Kumar and Mehul Damani and Sebastian Risi and Omar Khattab and Zhang-Wei Hong and Pulkit Agrawal},
booktitle={COLM 2026 Workshop on Efficient Reasoning},
year={2026},
url={https://openreview.net/forum?id=1kkO8IjjYM}
}

@article{cui2025entropy,
  title={{The Entropy Mechanism of Reinforcement Learning for Reasoning Language Models}},
  author={Cui, Ganqu and Zhang, Yuchen and Chen, Jiacheng and Yuan, Lifan and Wang, Zhi and Zuo, Yuxin and Li, Haozhan and Fan, Yuchen and Chen, Huayu and Chen, Weize and Liu, Zhiyuan and Peng, Hao and Bai, Lei and Ouyang, Wanli and Cheng, Yu and Zhou, Bowen and Ding, Ning},
  year={2025},
  journal={arXiv preprint arXiv:2505.22617},
  eprint={2505.22617},
  archivePrefix={arXiv},
  primaryClass={cs.LG}
}

@article{li2025darling,
  title={{Jointly Reinforcing Diversity and Quality in Language Model Generations}},
  author={Li, Tianjian and Zhang, Yiming and Yu, Ping and Saha, Swarnadeep and Khashabi, Daniel and Weston, Jason and Lanchantin, Jack and Wang, Tianlu},
  year={2025},
  journal={arXiv preprint arXiv:2509.02534},
  eprint={2509.02534},
  archivePrefix={arXiv},
  primaryClass={cs.CL}
}

@article{giants,
  title={{{GIANTS}: Generative Insight Anticipation from Scientific Literature}},
  author={He-Yueya, Joy and Singh, Anikait and Gao, Ge and Li, Michael Y. and Yang, Sherry and Finn, Chelsea and Brunskill, Emma and Goodman, Noah D.},
  journal={arXiv preprint arXiv:2604.09793},
  year={2026}
}

@article{tong2026ai,
  title={{AI Can Learn Scientific Taste}},
  author={Tong, Jingqi and Li, Mingzhe and Li, Hangcheng and Yang, Yongzhuo and Mou, Yurong and Ma, Weijie and Xi, Zhiheng and Chen, Hongji and Liu, Xiaoran and Cheng, Qinyuan and others},
  journal={arXiv preprint arXiv:2603.14473},
  year={2026}
}

@article{kirgis2026can,
  title={{Can AI agents conduct open-ended AI research? Early evidence from two case studies}},
  author={Kirgis, Peter and Kapoor, Sayash and Schwartz, Andrew and Rabanser, Stephan and Africa, David and Voudouris, Konstantinos and Nguyen, Viet and Pilditch, Toby and Dubois, Magda and Coppock, Harry and others},
  journal={arXiv preprint arXiv:2607.27191},
  year={2026}
}

@article{ning2026one,
  title={{One Run Is Not an Idea: The Implementation Lottery in Automated Research}},
  author={Ning, Jingjie and Zhong, Shanshan and Li, Xiaochuan and Zeng, Ji and Xiong, Chenyan},
  journal={arXiv preprint arXiv:2607.26587},
  year={2026}
}

@article{falck2026training,
  title={{Training AI Scientists to Replicate Research}},
  author={Falck, Damon and Sabri, Samer and Surina, Anja and Foster, Thom and Sims, Anya and Devlin, Sam and Rogers, Dylan and Collins, Tantum and Aleksiev, Kaloyan and Kirsch, Louis and others},
  journal={arXiv preprint arXiv:2608.13331},
  year={2026}
}

@article{garikaparthi2026researchgym,
  title={{ResearchGym: Evaluating Language Model Agents on Real-World AI Research}},
  author={Garikaparthi, Aniketh and Patwardhan, Manasi and Cohan, Arman},
  journal={arXiv preprint arXiv:2602.15112},
  year={2026}
}

@article{klahr1988dual,
  title={{Dual Space Search during Scientific Reasoning}},
  author={Klahr, David and Dunbar, Kevin},
  journal={Cognitive science},
  volume={12},
  number={1},
  pages={1--48},
  year={1988},
  publisher={Wiley Online Library}
}

@article{wadhwa2026create,
  title={{CREATE: Testing LLMs for Associative Creativity}},
  author={Wadhwa, Manya and Roy, Tiasa Singha and Lederman, Harvey and Li, Junyi Jessy and Durrett, Greg},
  journal={arXiv preprint arXiv:2603.09970},
  year={2026}
}

@article{shao2024deepseekmath,
  publtype={informal},
  author={Zhihong Shao and Peiyi Wang and Qihao Zhu and Runxin Xu and Junxiao Song and Mingchuan Zhang and Y. K. Li and Y. Wu and Daya Guo},
  title={DeepSeekMath: Pushing the Limits of Mathematical Reasoning in Open Language Models},
  year={2024},
  cdate={1704067200000},
  journal={CoRR},
  volume={abs/2402.03300},
  url={https://doi.org/10.48550/arXiv.2402.03300}
}

@article{chen2026measuring,
  title={{Measuring the Gap Between Human and LLM Research Ideas}},
  author={Chen, Ziyu and Zhao, Yilun and Cohan, Arman},
  journal={arXiv preprint arXiv:2607.01233},
  year={2026}
}

@article{platt1964strong,
  title={{Strong Inference: Certain systematic methods of scientific thinking may produce much more rapid progress than others}},
  author={Platt, John R.},
  journal={Science},
  volume={146},
  number={3642},
  pages={347--353},
  year={1964},
  publisher={American Association for the Advancement of Science}
}

@article{liu2026benchmarking,
  title={{Benchmarking AI Agents for Addressing Scientific Challenges Across Scales}},
  author={Liu, Tianyu and Wang, Allen Xin and Panescu, Antonia and Chen, Lisa Xinyi and Long, Wenxin and Wei, Xinyu and Jing, Yueqian and Zeng, Ziyao and Chen, Jihang and Jiang, Sihan and others},
  journal={arXiv preprint arXiv:2606.12736},
  year={2026}
}

@inproceedings{si2601towards,
title={{Towards Execution-Grounded Automated {AI} Research}},
author={Chenglei Si and Zitong Yang and Yejin Choi and Emmanuel Candes and Diyi Yang and Tatsunori Hashimoto},
booktitle={Forty-third International Conference on Machine Learning (ICML)},
year={2026},
note={arXiv:2601.14525},
url={https://openreview.net/forum?id=j0PagZWEFo}
}

@inproceedings{cohan2019structural,
  title={{Structural Scaffolds for Citation Intent Classification in Scientific Publications}},
  author={Cohan, Arman and Ammar, Waleed and Van Zuylen, Madeleine and Cady, Field},
  booktitle={Proceedings of the 2019 conference of the North American chapter of the Association for Computational Linguistics: human language technologies, volume 1 (long and short papers)},
  pages={3586--3596},
  year={2019}
}

@article{kinney2023semantic,
  title={{The Semantic Scholar Open Data Platform}},
  author={Kinney, Rodney and Anastasiades, Chloe and Authur, Russell and Beltagy, Iz and Bragg, Jonathan and Buraczynski, Alexandra and Cachola, Isabel and Candra, Stefan and Chandrasekhar, Yoganand and Cohan, Arman and others},
  journal={arXiv preprint arXiv:2301.10140},
  year={2023}
}

@article{lin1991divergence,
  title={{Divergence measures based on the Shannon entropy}},
  author={Lin, Jianhua},
  journal={IEEE Transactions on Information theory},
  volume={37},
  number={1},
  pages={145--151},
  year={1991},
  publisher={IEEE}
}

@book{boden2004creative,
  title={{The Creative Mind: Myths and Mechanisms}},
  author={Boden, Margaret A},
  year={2004},
  publisher={Routledge}
}

@article{guilford1967nature,
  title={{The Nature of Human Intelligence.}},
  author={Guilford, Joy Paul},
  year={1967},
  publisher={McGraw-Hill}
}

@inproceedings{wang2024scimon,
  title={{{SciMON}: Scientific Inspiration Machines Optimized for Novelty}},
  author={Wang, Qingyun and Downey, Doug and Ji, Heng and Hope, Tom},
  booktitle={Proceedings of the 62nd annual meeting of the association for computational linguistics (volume 1: long papers)},
  pages={279--299},
  year={2024}
}

@misc{qwen3technicalreport,
      title={{Qwen3 Technical Report}}, 
      author={Qwen Team},
      year={2025},
      eprint={2505.09388},
      archivePrefix={arXiv},
      primaryClass={cs.CL},
      url={https://arxiv.org/abs/2505.09388}, 
}

@inproceedings{
zhang2025noveltybench,
title={{NoveltyBench: Evaluating Creativity and Diversity in Language Models}},
author={Yiming Zhang and Harshita Diddee and Susan Holm and Hanchen Liu and Xinyue Liu and Vinay Samuel and Barry Wang and Daphne Ippolito},
booktitle={Second Conference on Language Modeling},
year={2025},
url={https://openreview.net/forum?id=XZm1ekzERf}
}

@misc{
lanchantin2025diverse,
title={{Diverse Preference Optimization}},
author={Jack Lanchantin and Angelica Chen and Shehzaad Dhuliawala and Ping Yu and Jason E Weston and Sainbayar Sukhbaatar and Ilia Kulikov},
year={2026},
url={https://openreview.net/forum?id=XF9rcsXd1q}
}

@inproceedings{padmakumar2024does,
  title={{Does Writing with Language Models Reduce Content Diversity?}},
  author={Padmakumar, Vishakh and He, He},
  booktitle={International Conference on Learning Representations},
  volume={2024},
  pages={642--669},
  year={2024}
}

@article{jiang2026artificial,
  title={{Artificial Hivemind: The Open-Ended Homogeneity of Language Models (and Beyond)}},
  author={Jiang, Liwei and Chai, Yuanjun and Li, Margaret and Liu, Mickel and Fok, Raymond and Dziri, Nouha and Tsvetkov, Yulia and Sap, Maarten and Choi, Yejin},
  journal={Advances in Neural Information Processing Systems},
  volume={38},
  year={2025}
}

@inproceedings{
shypula2025evaluating,
title={{Evaluating the Diversity and Quality of {LLM} Generated Content}},
author={Alexander Shypula and Shuo Li and Botong Zhang and Vishakh Padmakumar and Kayo Yin and Osbert Bastani},
booktitle={Second Conference on Language Modeling},
year={2025},
url={https://openreview.net/forum?id=O7bF6nlSOD}
}

@inproceedings{hu2024nova,
    title = {{``{NOVA}: An Iterative Planning Framework for Enhancing Scientific Innovation with Large Language Models"}},
    author = "Hu, Xiang  and
      Fu, Hongyu  and
      Wang, Jinge  and
      Wang, Yifeng  and
      Li, Zhikun  and
      Xu, Renjun  and
      Lu, Yu  and
      Jin, Yaochu  and
      Pan, Lili  and
      Lan, Zhenzhong",
    editor = "Che, Wanxiang  and
      Nabende, Joyce  and
      Shutova, Ekaterina  and
      Pilehvar, Mohammad Taher",
    booktitle = "Findings of the Association for Computational Linguistics: ACL 2025",
    month = jul,
    year = "2025",
    address = "Vienna, Austria",
    publisher = "Association for Computational Linguistics",
    url = "https://aclanthology.org/2025.findings-acl.1099/",
    doi = "10.18653/v1/2025.findings-acl.1099",
    pages = "21330--21359",
    ISBN = "979-8-89176-256-5"
}

@inproceedings{baek2025researchagent,
  title={{ResearchAgent: Iterative Research Idea Generation over Scientific Literature with Large Language Models}},
  author={Baek, Jinheon and Jauhar, Sujay Kumar and Cucerzan, Silviu and Hwang, Sung Ju},
  booktitle={Proceedings of the 2025 conference of the nations of the Americas chapter of the association for computational linguistics: human language technologies (volume 1: long papers)},
  pages={6709--6738},
  year={2025}
}

@misc{olmo2025olmo3,
title={{Olmo 3}},
author={Team Olmo and Allyson Ettinger and Amanda Bertsch and Bailey Kuehl and David Graham and David Heineman and Dirk Groeneveld and Faeze Brahman and Finbarr Timbers and Hamish Ivison and Jacob Morrison and Jake Poznanski and Kyle Lo and Luca Soldaini and Matt Jordan and Mayee Chen and Michael Noukhovitch and Nathan Lambert and Pete Walsh and Pradeep Dasigi and Robert Berry and Saumya Malik and Saurabh Shah and Scott Geng and Shane Arora and Shashank Gupta and Taira Anderson and Teng Xiao and Tyler Murray and Tyler Romero and Victoria Graf and Akari Asai and Akshita Bhagia and Alexander Wettig and Alisa Liu and Aman Rangapur and Chloe Anastasiades and Costa Huang and Dustin Schwenk and Harsh Trivedi and Ian Magnusson and Jaron Lochner and Jiacheng Liu and Lester James V. Miranda and Maarten Sap and Malia Morgan and Michael Schmitz and Michal Guerquin and Michael Wilson and Regan Huff and Ronan Le Bras and Rui Xin and Rulin Shao and Sam Skjonsberg and Shannon Zejiang Shen and Shuyue Stella Li and Tucker Wilde and Valentina Pyatkin and Will Merrill and Yapei Chang and Yuling Gu and Zhiyuan Zeng and Ashish Sabharwal and Luke Zettlemoyer and Pang Wei Koh and Ali Farhadi and Noah A. Smith and Hannaneh Hajishirzi},
year={2025},
eprint={2512.13961},
archivePrefix={arXiv},
primaryClass={cs.CL},
url={https://arxiv.org/abs/2512.13961},
}

@misc{openai2025gptoss120bgptoss20bmodel,
      title={{gpt-oss-120b and gpt-oss-20b Model Card}}, 
      author={OpenAI},
      year={2025},
      eprint={2508.10925},
      archivePrefix={arXiv},
      primaryClass={cs.CL},
      url={https://arxiv.org/abs/2508.10925}, 
}

@misc{nvidia_nemotron_3_2025,
  title  = {{NVIDIA Nemotron 3: Efficient and Open Intelligence}},
  author = {{NVIDIA}},
  year   = {2025},
  url    = {https://arxiv.org/abs/2512.20856},
  note   = {White Paper}
}

@inproceedings{
dubois2404length,
title={{Length-Controlled AlpacaEval: A Simple Debiasing of Automatic Evaluators}},
author={Yann Dubois and Percy Liang and Tatsunori Hashimoto},
booktitle={First Conference on Language Modeling},
year={2024},
url={https://openreview.net/forum?id=CybBmzWBX0}
}

@inproceedings{
sprague2025to,
title={{To CoT or not to CoT? Chain-of-thought helps mainly on math and symbolic reasoning}},
author={Zayne Rea Sprague and Fangcong Yin and Juan Diego Rodriguez and Dongwei Jiang and Manya Wadhwa and Prasann Singhal and Xinyu Zhao and Xi Ye and Kyle Mahowald and Greg Durrett},
booktitle={The Thirteenth International Conference on Learning Representations},
year={2025},
url={https://openreview.net/forum?id=w6nlcS8Kkn}
}

@article{sheng2024hybridflow,
  title   = {{HybridFlow: A Flexible and Efficient RLHF Framework}},
  author  = {Guangming Sheng and Chi Zhang and Zilingfeng Ye and Xibin Wu and Wang Zhang and Ru Zhang and Yanghua Peng and Haibin Lin and Chuan Wu},
  year    = {2024},
  journal = {arXiv preprint arXiv: 2409.19256}
}

@article{liu2025understanding,
  title={{Understanding R1-Zero-Like Training: A Critical Perspective}},
  author={Liu, Zichen and Chen, Changyu and Li, Wenjun and Qi, Penghui and Pang, Tianyu and Du, Chao and Lee, Wee Sun and Lin, Min},
  journal={arXiv preprint arXiv:2503.20783},
  year={2025}
}

@article{zheng2023judging,
  title={{Judging LLM-as-a-Judge with MT-Bench and Chatbot Arena
}},
  author={Zheng, Lianmin and Chiang, Wei-Lin and Sheng, Ying and Zhuang, Siyuan and Wu, Zhanghao and Zhuang, Yonghao and Lin, Zi and Li, Zhuohan and Li, Dacheng and Xing, Eric and others},
  journal={Advances in neural information processing systems},
  volume={36},
  pages={46595--46623},
  year={2023}
}

@inproceedings{kirk2024understanding,
  title={{Understanding the Effects of RLHF on LLM Generalisation and Diversity}},
  author={Kirk, Robert and Mediratta, Ishita and Nalmpantis, Christoforos and Luketina, Jelena and Hambro, Eric and Grefenstette, Edward and Raileanu, Roberta},
  booktitle={International Conference on Learning Representations},
  volume={2024},
  pages={20620--20653},
  year={2024}
}

@inproceedings{
zhang2026verbalized,
title={{Verbalized Sampling: How to Mitigate Mode Collapse and Unlock {LLM} Diversity}},
author={Jiayi Zhang and Simon Yu and Derek Chong and Anthony Sicilia and Michael Tomz and Christopher D Manning and Weiyan Shi},
booktitle={Forty-third International Conference on Machine Learning},
year={2026},
url={https://openreview.net/forum?id=zloIrd77G5}
}
\bibliographystyle{iclr2027_conference}

\appendix

\section*{AI Use Statement}


The training data for our proposed method was generated using AI. We used Claude Opus-4.8 for it. In addition, code written with generative AI was used to conduct and carry out the analyses here. However, the research hypotheses in this work and the design of the experiments were both human-originated. AI tools were used for assistance with revisions, but not for writing in a substantive way. The authors take responsibility for the final content of this work.

\section{Dataset}
\label{app:data}

\subsection{Data Format}
\label{app:data_format}

In this Section we explain different aspects of the data curation process, including document acquisition, formats etc.

\paragraph{Idea Format} Each idea the model generates is a JSON object in the four-field schema of Section~\ref{sec:ideation}. This includes: (\textit{problem\_gap}, \textit{core\_claim}, \textit{method}, \textit{inspiration}). The definitions of the four fields are given below (as per the inference prompt):
\begin{itemize}
    \item \textit{problem gap}: The gap the idea addresses, drawn from across the source documents -- an open question, tension, or unresolved difficulty that connects them and that none resolves on its own. State it as one clear gap, NOT as your specific solution to it. BEGIN this field with the single best-fitting OPPORTUNITY PATTERN label in square brackets (from the taxonomy below), then the prose gap (1 to 2 sentences). Pick the ONE dominant pattern even if several apply; if NONE genuinely fits, use ``[Other]'' rather than forcing a poor match. Example: ``[Bridge Opportunity] Work on X and on Y each..but neither connects..''.

\item \textit{core claim}:  The couple of sentences the whole idea lives or dies on -- ``we claim that X.'' It must be something that could turn out to be wrong: a definite, testable assertion / hypothesis / insight, not a topic or a direction. Do not just restate the gap, and do not describe what you would do about it. From these sentences alone a reader should know exactly what the idea is betting on (1-2 sentences).

\item \textit{method}: At a high level, how the claim would be turned into research -- the kind of move (combining ideas, extending to a new setting, deriving formally, building an artifact, etc.) and how it addresses the gap. A proposed approach, not a step-by-step protocol. Begins with one bracketed Method Paradigm label.

\item \textit{inspiration}: Three to five sentences arguing how the idea arises FROM the source documents -- the relationship between them, not a per-document restatement. Must (1) name each cited source's limitation and (2) justify why connecting them is a non-obvious leap. Refers to sources by their numbered `Document N' label.
\end{itemize}

 \textit{problem gap} and \textit{method} are each required to open with one bracketed taxonomy label (an \emph{opportunity pattern} and a \emph{method paradigm}, respectively) drawn from the fixed 7+7 taxonomy given in full in the ideation prompt \citep{chen2026measuring} (Appendix~\ref{app:prompts}). The example idea card in Appendix~\ref{app:prompts} shows how an idea card is structured.

\subsection{Data Curation: Gathering Documents}
\label{app:document_collection}

We describe the dataset construction in detail in this section. As mentioned in Section~\ref{sec:data}, our dataset curation has two steps: (a) gathering the documents (b) putting together the SFT dataset. In this subsection we discuss the gathering documents. 

Each instance in our dataset has a \emph{target} or \emph{gold} idea which is derived from a published paper. For each gold idea, we also store a subset of its references that we consider as ``core'' to its generation. These references are inputs to our proposed method. We use the GiantsBench \citep{giants} training split to get the target paper ids. We detail the generation process and each of the steps below.

\paragraph{Pipeline.} For every target ID we proceed through the following stages:
\begin{enumerate}
    \item \textbf{Target full text.} We fetch the target's arXiv HTML (falling back to ar5iv for older papers) and parse it into \textit{\{title, abstract, sections, tables, references, full text\}}. 
    \item \textbf{Reference edges.} We pull the target's full reference list as structured edges from the Semantic Scholar API~\citep{kinney2023semantic}, with per-edge citation intents, an \texttt{isInfluential} flag, and the cited paper's metadata (title, abstract, year, venue, open-access PDF link). 
    \item \textbf{Importance classification.} An LLM (\texttt{claude-sonnet-4-6}) reads the target's title, abstract, and body once, together with every reference and all of its collected mention snippets, and decides which references are \emph{sources} i.e. the small subset that motivated the paper, exposed a gap it targets, supplied its conceptual foundation, or is directly extended. We get this classification from prior work done on citation importance~\citep{cohan2019structural}. A reference can be a \emph{source} or an \emph{other}. Each \emph{source} receives one or more of four labels (\texttt{motivates}, \texttt{exposes\_gap}, \texttt{foundation}, \texttt{extends}), a short rationale, and verbatim supporting quotes. The full prompt is given in Prompt~\ref{prompt:source_select}.
    \item \textbf{Reference full text.} For every source reference, we get the paper text through the process described in Step 1. 
    \item \textbf{Summarization.} We summarize the target and source documents using the idea card format in Appendix~\ref{app:data_format}. For the source documents, we add two additional fields i.e. \textit{results} and \textit{takeaways}. The target's own contribution is extracted separately as a \emph{gold idea} used only inside the reward (the impact comparison, Section~\ref{sec:instantiation}) and is never shown to the policy. 
    \item \textbf{Consolidation.} We drop targets with two or fewer source documents or no gold idea, and then join everything into one record per target. The \textit{source} references become the input, and the target paper becomes the \textit{gold} document.
\end{enumerate}

\begin{table}[t]
\centering
\small
\renewcommand{\arraystretch}{1.3}
\caption{Representative source citations, one per functional label. The in-text context is the sentence in the target paper that cites the source; the label is the role assigned by the importance classifier}
\begin{tabular}{@{}p{0.60\linewidth}p{0.24\linewidth}l@{}}
\toprule
\textbf{In-text context} & \textbf{Source} & \textbf{Label} \\
\midrule
``Recent work \textbf{Gilardi et al. ( 2023 )}; He et al. ( 2023a ); Ding et al. ( 2022 ) has looked at the annotation accuracy for LLMs and found them to be at par with crowd-worker annotators.'' & ChatGPT outperforms crowd workers for text-annotation tasks & \texttt{motivates} \\
``\textbf{MuRAG Chen et al. ( 2022 )} requires 3 million image-text pairs for pre-training to achieve decent results.'' & MuRAG: Multimodal Retrieval-Augmented Generator for Open Question Answering over Images and Text & \texttt{exposes\_gap} \\
``The global strategy involves using a trained image caption model to generate text descriptions of the macro scenes of the image \textbf{Changpinyo et al. ( 2022 )}; Yang et al. ( 2022b )'' & All You May Need for VQA are Image Captions & \texttt{foundation} \\
``This paper is an extended work of an earlier version presented at SOMET 2022 \textbf{(Shirafuji et al., 2022 )}'' & Prompt Sensitivity of Language Model for Solving Programming Problems & \texttt{extends} \\
\bottomrule
\end{tabular}
\label{tab:sparker_examples}
\end{table}

\begin{wrapfigure}{r}{0.4\textwidth}
\vspace{-2em}
\centering
\includegraphics[scale=0.20, trim= 0 8cm 27cm 0]{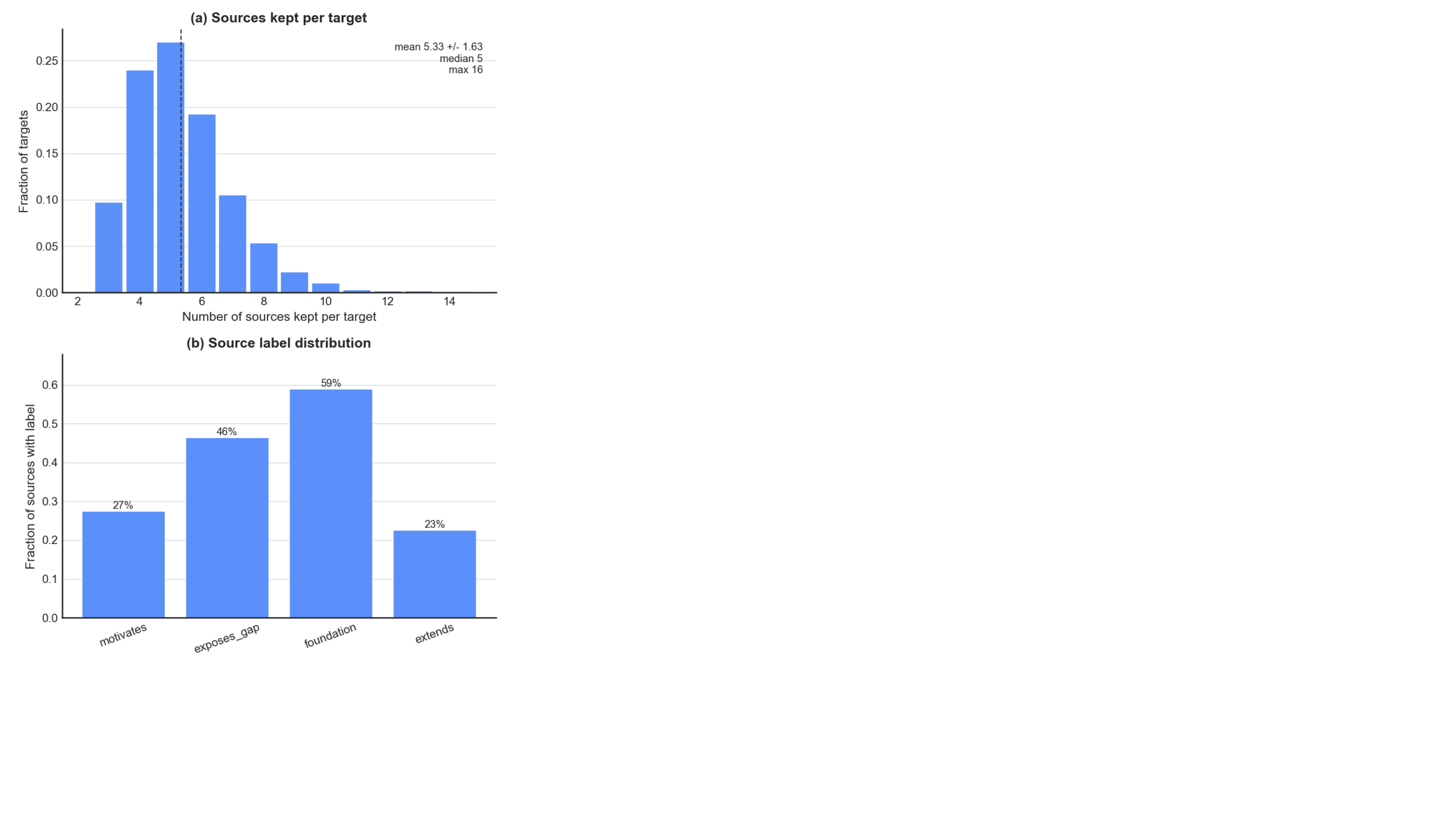}
\caption{Distribution of (a)  number of sources kept per-target post source filtering (b) the different citation/source types post-filtering. }
\label{fig:sparker_stats}
\vspace{-3em}
\end{wrapfigure}

We show examples of the source reference labeling in Table~\ref{tab:sparker_examples}. Figure~\ref{fig:sparker_stats} shows statistics about the number of references kept, and the distribution of the different source types amongst the selected documents.

\vspace{-1em}

\subsection{Training Data Curation}
\label{app:training_data}
As briefly discussed in Section~\ref{sec:data}, we use hindsight prompting to generate SFT data with thinking traces. Having collected the documents/instances from the steps in Appendix~\ref{app:document_collection}, we now build that into a training step. Since our training pipeline has two phases: (a) SFT and (b) GRPO, we first talk about the SFT dataset and the RL training follows from that. 

The SFT data curation process has two steps: (a) Generating $K$ ideas from a strong LLM (b) Generating thinking traces for the $K$ ideas. We use \texttt{claude-opus-4.8} as the teacher model. We first prompt it with Prompt~\ref{prompt:idea_generation} to generate 3 ideas per instance. Then we use the Prompt~\ref{prompt:trace_synth} to generate the thinking trace in hindsight. Example~\ref{example:sft_trace} shows an example generated with the above pipeline.  We see instances of the model exploring its own context, verifying its candidates, making them more concrete, making sure the ideas are not similar or the tags are correct etc.

\begin{figure}
    \centering
    \includegraphics[scale=0.25,trim=1cm 25cm 0 0]{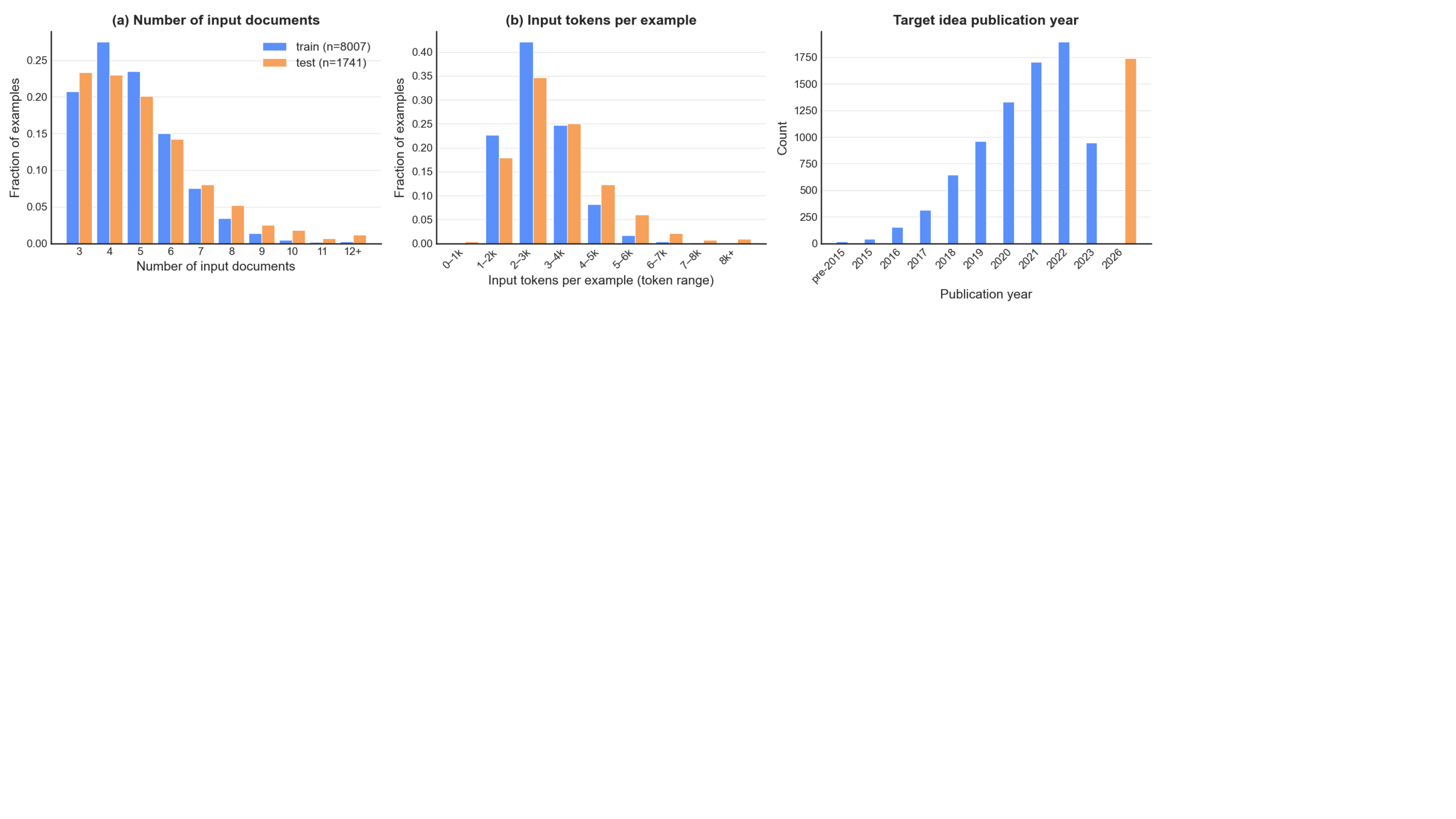}
    \caption{Statistics around the training and test datasets used in this work. Our training and test distributions contain varying number of source documents as inputs in order for the model to generate ideas, we also maintain a strict contamination boundary for the models we are training and evaluating. The legend counts (train $n{=}8{,}007$, test $n{=}1{,}741$) are the full pools after the source-count filter; the models are SFT'd on 7{,}172 of the training instances (after some filtering) and evaluated on 1{,}740 test instances. }
    \label{fig:data_distribution}
\end{figure}

\begin{wrapfigure}{r}{0.5\textwidth}
\vspace{-3em}
\centering
\includegraphics[scale=0.4]{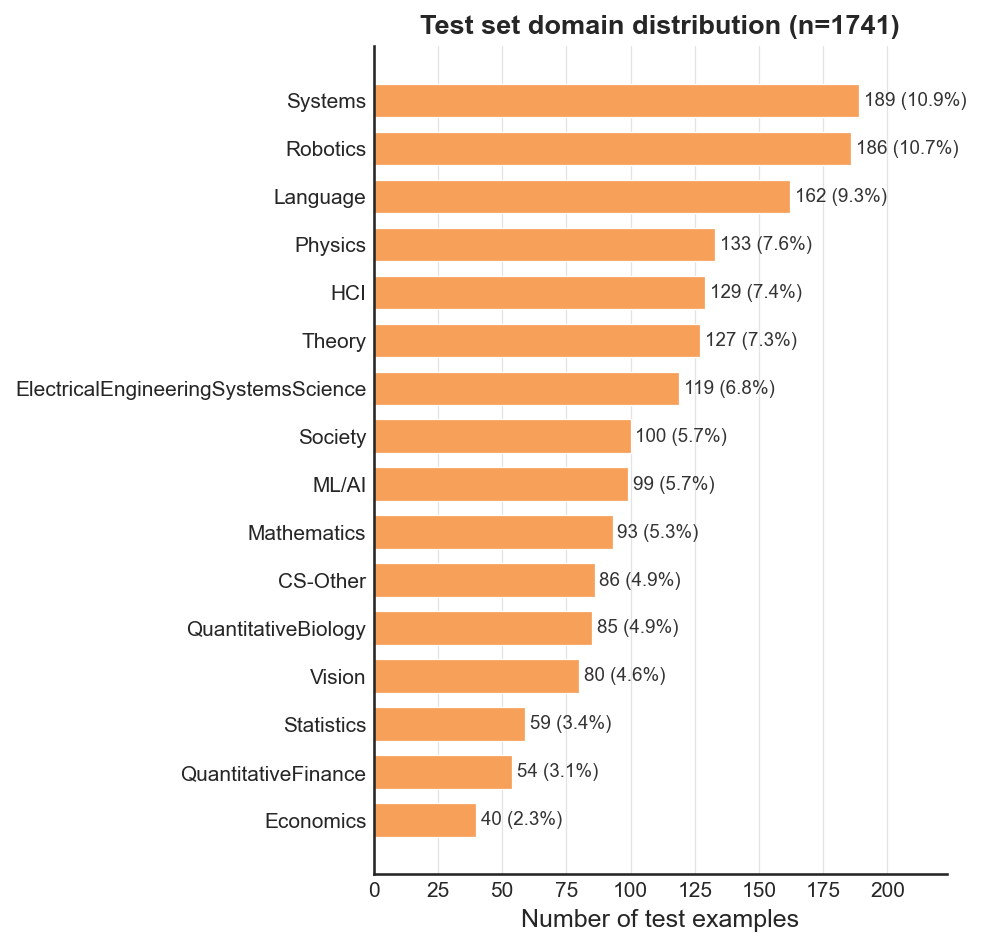}
\caption{Distribution of domains in the test dataset.}
\label{fig:domain_distribution}
\vspace{-3em}
\end{wrapfigure}

Figure~\ref{fig:data_distribution} shows the final distribution of (a) number of input documents (b) distribution of different input tokens (c) publications considered over the years, for the train and the test splits. Figure~\ref{fig:domain_distribution} also shows the distribution of the test set across different domains. On average, the completion is 4000 tokens, with 1300 tokens for the answer (3 ideas) and 2800 tokens for the thinking trace.

Note, that we train only from the language domain, but test across different domains for out of domain generalization.


\section{Training Details}
\label{app:training}
\paragraph{Training/Input Prompt} The same template is used to build both the live-inference user turn and the training input, so the model sees an identical instruction shape at train and test time. \texttt{\{num\_ideas\}} is substituted with $K$ (3 in the main runs) and \texttt{\{documents\}} with the numbered source-document block of Appendix~\ref{app:data_format}.

\paragraph{Reward implementation details.} Along with Equation~\ref{eq:reward} we have two terms that the training reward adds around the set-level score to make it operational.

\textbf{Schema gate and bonus.} A rollout is schema-valid only if it contains both a \texttt{<think>} and an \texttt{<answer>} block, the answer parses into exactly $K$ idea cards, every card has all four fields non-empty, and the \texttt{problem\_gap} and \texttt{method} fields open with a valid taxonomy label. An invalid rollout receives reward $0$ and no judge is called. A valid rollout additionally receives a small bonus $\epsilon=0.05$ on top of the set score.

\textbf{Length penalty.} To keep RL from lengthening the cards to raise the judged impact, we subtract a one-sided, dead-banded, capped penalty. Let $L$ be the number of Qwen3 tokens in the concatenated \texttt{problem\_gap}, \texttt{core\_claim}, and \texttt{method} fields of the $K$ cards, i.e.\ exactly the text the quality and distance judges read. The \texttt{inspiration} field and the \texttt{<think>} block are not measured, so trimming grounding cannot avoid the penalty. With target length $L_0$, tolerance $\tau$, slope $\lambda$, and cap $c$,
\begin{equation}
\label{eq:length_penalty}
\text{pen}(L)=\lambda\cdot\min\!\Big(\max\big(0,\ L/L_0-(1+\tau)\big),\ c\Big).
\end{equation}
We use $L_0=1{,}000$, $\tau=0$, $\lambda=1$, and $c=3$, so no penalty applies below $1{,}000$ tokens and the penalty never exceeds $\lambda c=3$ (reached at $4{,}000$ tokens). The full training reward is therefore:
\begin{equation}
\label{eq:full_reward}
R(x,U)=\mathbf{1}[\text{schema}]\Big(\epsilon+\sum_{i=1}^{K} q(u_i)\cdot\min_{j<i} d(u_i,u_j)-\text{pen}(L)\Big).
\end{equation}

\subsection{Hyperparameters}
\label{app:hyperparameters}
We train with two stages (SFT then GRPO) using \texttt{verl}~\citep{sheng2024hybridflow}. Table~\ref{tab:hyperparams} lists the settings for the main Qwen3-8B run; the Olmo-3 runs use the same recipe (Appendix~\ref{app:olmo}).


\begin{table}[t!]
\centering
\small
\footnotesize
\caption{Hyperparameters for the main Qwen3-8B run.}
\label{tab:hyperparams}
\begin{tabular}{lll}
\toprule
 & \textbf{SFT} & \textbf{GRPO} \\
\midrule
Init & \texttt{Qwen3-8B} & SFT checkpoint (after 1 epoch) \\
Training instances & \multicolumn{2}{l}{7,172 (same instances for SFT and GRPO), plus 100 validation} \\
Learning rate & $5\times10^{-6}$ & $1\times10^{-6}$ \\
Batch size & 64 sequences & 12 prompts $\times$ 4 rollouts \\
PPO mini-batch (prompts) & -- & 12 (one update per step) \\
Epochs / steps & 1 epoch (112 steps) & 600 steps / 1 epoch \\
Max prompt length & -- & 10,240 \\
Max response / sequence length & 16,384 (right-truncated) & 16,384 \\
Rollout temperature & -- & 1.0 \\
Advantage & -- & GRPO, no std norm (Dr.\ GRPO~\citep{liu2025understanding}) \\
PPO clip ($\epsilon_{\text{low}}/\epsilon_{\text{high}}$) & -- & 0.2 / 0.5 \\
KL loss & -- & 0.001 (\texttt{low\_var\_kl}) \\
Entropy bonus & -- & 0 \\
Ideas per rollout ($K$) & 3 & 3 \\
Schema bonus $\epsilon$ & -- & 0.05 \\
Length penalty $(L_0,\tau,\lambda,c)$ & -- & $(1000,\,0,\,1,\,3)$ \\
Rollout engine & -- & vLLM \\
\bottomrule
\end{tabular}
\end{table}

\paragraph{SFT.} We train on the hindsight-prompted traces of Appendix~\ref{app:training_data} with the loss on the full completion (thinking trace and answer) and none on the prompt. Sequences longer than 15,872 tokens are dropped when building the data, and 100 instances are held out for validation loss and format validity. We train the model for 1 epoch

\paragraph{GRPO.} Each step samples 12 prompts with 4 rollouts each. A rollout is a full set of $K{=}3$ ideas scored by the reward of Eq.~\ref{eq:full_reward}: quality from \texttt{SciJudge-30B} and distance from \texttt{gpt-oss-120b} (Appendix~\ref{app:metrics}). We remove the standard-deviation normalization of advantages, use asymmetric clipping, and keep a small KL term to the SFT policy. We evaluate checkpoint 600 for both versions of our proposed model (no think and think).

\subsection{SFT Variants: With and Without Thinking}
\label{app:sft_variants}
We SFT two variants of every policy, which differ only in what appears between the \texttt{<think>} tags. Both are followed by the same GRPO stage.
\begin{itemize}
    \item \textbf{With thinking.} The target is the hindsight thinking trace generated by the teacher (Appendix~\ref{app:training_data}), followed by the answer containing the $K$ idea cards.
    \item \textbf{Without thinking (empty think).} The same answers, but the \texttt{<think>} block is empty.
\end{itemize}
Both variants share the same 7,172 training instances and the same 100 validation instances, so the answers are identical and only the thinking span differs. 

\subsection{DARLING} \label{app:darling} 
DARLING~\citep{li2025darling} optimizes quality and diversity \emph{across the rollouts} of one prompt, whereas our reward enforces diversity \emph{within} a single generated set. We re-implement it in our setting so that the only intended difference from our method is where diversity is enforced. We keep DARLING's group-partition diversity term and its multiplicative reward, and we replace its learned equivalence classifier and quality reward with the judges used elsewhere in this paper.

\textbf{Setup (no SFT).} DARLING is trained with $K{=}1$: each rollout is a single idea card, and the group of rollouts for a prompt provides the diversity signal. Following their recipe, we start GRPO directly from Qwen3-8B, with no SFT warm-start. The model therefore learns the card format only from the schema check in the reward and from the prompt (identical to Appendix~\ref{app:training}). This matches the original method, which also applies RL to an instruction-tuned model without a task-specific SFT stage. The training instances, prompt, and judges are the same as for our method.

\textbf{Reward.} For a group of $n{=}8$ rollouts $\{y_1,\dots,y_n\}$ for the same prompt, the reward of rollout $i$ is
\begin{equation}
r_i = q(y_i)\cdot \mathrm{div}_i, \qquad \mathrm{div}_i=\max\!\Big(0.1,\ \frac{n-|C(i)|}{n-1}\Big),
\end{equation}
where $q(y_i)$ is the same pairwise impact win against the gold idea as in Appendix~\ref{app:metrics-quality} (SciJudge-30B), and $C(i)$ is the cluster containing $i$. Clusters come from union-find over a pairwise equivalence test, as in DARLING's released code. Two rollouts are merged if the distance between their ideas is below a threshold of 0.5. Instead of DARLING's trained Qwen3-Embedding classifier, we test equivalence with our distance judge (\texttt{gpt-oss-120b}, Appendix~\ref{app:metrics}). A rollout in a large group of near-duplicates therefore gets a low diversity score, and a unique rollout gets a score close to 1. With $K{=}1$ there is one idea per rollout, so no set-level aggregation is needed. An invalid rollout receives reward 0.

\textbf{Optimization.} We follow DARLING's recipe: $n{=}8$ rollouts per prompt, 12 prompts per step, advantages without standard-deviation normalization, symmetric PPO clipping at 0.2, no entropy bonus, KL coefficient 0.001, learning rate $1\times10^{-6}$, and rollout temperature 1.0 for training and validation. 


\textbf{Evaluation.} At test time we sample three independent generations per instance and treat them as the set, since the policy was trained to produce one idea per generation. All metrics in Section~\ref{sec:main_results} are computed on this set with the same judges as for the other models.

\subsection{Olmo-3 Details}
\label{app:olmo}
To test that the recipe is not specific to one model family, we repeat it on Olmo-3-7B \citep{olmo2025olmo3}. We initialize from \texttt{allenai/Olmo-3-7B-Think-SFT}, and run the same SFT (with and without thinking) and GRPO settings as Table~\ref{tab:hyperparams}. The length penalty is measured with the Olmo tokenizer. The two tokenizers count the card text within about $0.5\%$, so $L_0$ is unchanged. Results are in Appendix~\ref{app:results} (Transfer of recipe to other models).

\subsection{Compute}
\label{app:compute}
All runs use a single node with 8 NVIDIA H200 GPUs. SFT uses all 8 GPUs and takes about 46 minutes per epoch on 7,172 instances. For GRPO, 6 GPUs hold the policy and the vLLM rollouts, and 2 GPUs serve \texttt{SciJudge-30B} (tensor parallel size 2); the \texttt{gpt-oss-120b} distance judge is served through an external API and needs no local GPUs. A GRPO step takes about one minute, so one epoch (about 600 steps at 12 prompts per step) takes roughly 10 hours.


\section{Metrics}
\label{app:metrics}

In this section we describe the metrics: (a) quality i.e. impact, (b) distance, (c) grounding and (d) creative utility. We show the prompts along with examples.

\subsection{Quality}
\label{app:metrics-quality}

We score the quality of an idea $q(u)$ via a pairwise reward model that compares a generated idea against the gold idea ($g$) for the same source set and returns which one is likely to have higher scientific impact. The resulting win/loss is the impact term $I(u)$ of Section~\ref{sec:instantiation}. The judge is called with both orderings of the pair to cancel position bias~\citep{zheng2023judging}, and $I(u)$ is the fraction of orderings won ($\in \{0, 0.5, 1\}$). Where if $u$ wins in both orderings, we reward it 1, if $u$ loses in both orderings we reward 0, and if $u$ wins in one and loses in the other, we call it a tie and reward 0.5.

\begin{wraptable}{r}{0.6\textwidth}
\centering
\small
\vspace{-3em}
\caption{Impact-judge win-rate. $^{\dagger}$ evaluation judge; $^{\ddagger}$ training-reward judge.}
\begin{tabular}{lc}
\toprule
Impact judge & win-rate  \\
\midrule
\texttt{SciJudge-30B}$^{\ddagger}$ & 0.65 \\
\texttt{gpt-5.4-mini}$^{\dagger}$ & 0.74  \\
\texttt{claude-sonnet-5} & 0.74  \\
\bottomrule
\end{tabular}
\label{tab:impact-citation-validation}
\end{wraptable}

\paragraph{Length-controlled win rate.} To avoid any length biases in the preferences, we report a length-controlled win rate, following Length-Controlled AlpacaEval \citep{dubois2404length}, computed after the fact without re-judging. We treat a strict win ($u\succ g$ in both orderings) as $1$ and a tie or loss as $0$, and fit one logistic regression over all ideas from all models in the evaluation pool, judged on the same instances by the same model. 


\paragraph{Validating Quality.} We validate the impact judge $q$
(Eq.~\ref{eq:quality}) two ways: (i) validation against real-world papers (ii) robustness to length.

\paragraph{(i) Agreement with real-world impact.} We draw from our training set, papers from 2022 and 2023 and form pairs the \emph{same} calendar year (within four months, to remove the age-of-paper confound). We only keep pairs that differ in their citation ratio (which we get from Semantic Scholar). We filter for a citation ratio of 3 with absolute gap of at least 10 citations.  To ensure the gap reflects meaningful use, we add an \emph{influential-citation} gate (both papers' influential-citation counts $>0$, winner above loser by a margin $\ge 5$, we get this from Semantic Scholar). We rewrite each instance in the pair into idea-only summaries that reveal no results, metrics, or titles; the judge then compares \emph{ideas}, and the higher-uptake paper is the gold winner. This yields $122$ labeled pairs ($244$ papers).  We ask the judge which idea is higher impact. For validation we only prompt the model once and report the strict win rate (no ties). On the idea-card format used throughout the paper, the evaluation judge \texttt{GPT-5.4-mini} picks the higher-impact idea $74\%$ of the time. The training-reward judge \texttt{SciJudge-30B}, scores $65\%$ (Table \ref{tab:impact-citation-validation}). Between Claude and GPT, we went ahead with the latter for cost reasons. 

Because the reward is evaluated on every rollout during RL, its judge is queried orders of magnitude more often than the evaluation judge, which is applied once per test instance. We therefore optimize against a smaller, cheaper judge and reserve a stronger judge for evaluation.

\paragraph{(ii) Robustness to length.} To rule out that the judge merely prefers longer ideas, we summarize each candidate idea down to the gold idea's length (around 250 tokens) without changing its content and re-judge (Table~\ref{tab:impact-length-validation}). The candidate ideas come from assorted baselines and earlier training checkpoints, chosen to span short to highly verbose ideas so the test covers the regime where length bias would show. We poll around 2000 idea pairs. The win rates in Table~\ref{tab:impact-length-validation} are raw strict win rates (the candidate wins in both orderings), not length-controlled. Shortening the ideas to the gold length does not lower the win rate; it raises it for all three judges (e.g.\ 0.253 to 0.284 for \texttt{gpt-5.4-mini}), so the judges do not simply reward longer ideas. The win rate is nevertheless sensitive to length, which is why we also report a length-controlled win-rate. 

\textbf{Inference parameters.} Default judge model for training is \texttt{SciJudge-30B}; temperature $0.0$ (greedy), max output tokens $2048$, \texttt{top\_p}$=0.8$, \texttt{top\_k}$=20$. For evaluation we use \texttt{GPT-5.4-mini} as the judge, with the same prompt and both orderings. We use the prompt mentioned in Prompt~\ref{prompt:impact} for estimating impact.

\begin{wraptable}{r}{0.6\textwidth}
\centering
\small
\caption{Impact-judge performance with length changes. $^{\dagger}$ evaluation judge.}
\renewcommand{\tabcolsep}{1.2mm}
\begin{tabular}{lcccc}
\toprule
Impact judge & \multicolumn{2}{c}{Length (tokens)} & \multicolumn{2}{c}{Strict win-rate}  \\
\cmidrule(lr){2-3}\cmidrule(lr){4-5}
 & orig & matched & orig & matched \\
\midrule
\texttt{gpt-5.4-mini}$^{\dagger}$ & 642 & 258 & 0.253 & 0.284  \\
\texttt{claude-sonnet-5} & 642 & 258 & 0.090 & 0.133  \\
\texttt{claude-sonnet-4.6} & 642 & 258 & 0.165 & 0.205 \\
\bottomrule
\end{tabular}
\label{tab:impact-length-validation}
\end{wraptable}

\subsection{Distance}
\label{app:metrics-distance}

We score the pairwise distance $d(u_i, u_j)$ between two idea cards via an LLM judge that rates similarity, field by field, over the \texttt{problem\_gap}, \texttt{core\_claim}, and \texttt{method} fields (Section~\ref{sec:instantiation}, Eq.~\ref{eq:distance}); the \texttt{inspiration} field is excluded since it carries grounding rather than content.

\textbf{From similarity to distance.} Each 1-5 similarity is inverted to a per-field distance $d_{\text{field}} = (5 - \text{sim})/4 \in [0,1]$, and the three fields are averaged with equal weight (each $1/3$) to give the pairwise distance $d(u_i, u_j)$ used in Eq.~\ref{eq:distance}.

\paragraph{Validating Distance}  We draw within-set idea pairs across a few different systems, so the pairs span a range of similarities the reward is supposed to see. Each pair is presented with the same idea format and every rater scores all three fields with the identical scoring rubric. Two annotators (also authors on the paper), independently score 100 pairs. We then run the LLM judges on the same pairs through the same prompt. In Table~\ref{tab:distance-validation}, we report the human annotator agreement as well as agreement with the LLM as a judge. Humans achieve a moderate agreement of around 0.36 (Spearman). The training-reward judge i.e. \texttt{gpt-oss-120b} matches the human-human ceiling on the aggregated distance ($\rho{=}0.37$ vs.\ a human--human $\rho{=}0.36$), so the reward reproduces the human ordering of idea pairs about as well as a second annotator does. The stronger reference judge which is used for the final evaluation i.e. \texttt{gemini-3.5-flash}, agrees more closely ($\rho{=}0.50$ on $d$).

\begin{wraptable}{r}{0.6\textwidth}
\centering
\small
\renewcommand{\tabcolsep}{1mm}
\caption{Distance-metric validation on $n{=}100$ within-set idea pairs rated by both human annotators, over the three scored fields and their aggregated distance $d\in[0,1]$ (Eq.~\ref{eq:distance}). All entries are Spearman $\rho$.}
\begin{tabular}{lccc}
\toprule
Field & HH  & gpt-oss-120b & gemini-3.5-flash \\
 & $\rho$ & $\rho$ & $\rho$ \\
\midrule
problem gap & 0.30 & 0.31 & 0.44 \\
core claim & 0.36 & 0.29 & 0.30 \\
method & 0.28 & 0.38 & 0.53 \\
\midrule
Distance $d\in[0,1]$ & 0.36 & 0.37 & 0.50 \\
\bottomrule
\end{tabular}

\label{tab:distance-validation}
\end{wraptable}

\textbf{Inference parameters.} The training reward uses \texttt{gpt-oss-120b} as the judge and evaluation uses \texttt{gemini-3.5-flash}; temperature $0.0$ (greedy); max output tokens $1024$. In the training reward, an idea's diversity credit is its distance to the nearest of up to $2$ earlier sampled/generated ideas in the set. In the offline diversity evaluation, all $\binom{K}{2}$ pairs within a generated set are scored and averaged. We use the prompt mentioned in Prompt~\ref{prompt:distance_prompt} for estimating this metric.

\subsection{Grounding}
\label{app:metrics-grounding}
For grounding, we measure if an idea's declared source support actually carries the idea. We use this metric only for reporting (Section~\ref{sec:main_results}).

\textbf{Source dependence score.} The score we report is \emph{source-dependence}. Would the idea collapse if the cited documents were removed? The judge rates this on a 1--5 rubric, from 1 (decorative: the idea stands completely without the cited documents) to 5 (the idea strictly requires the cited findings), and looks for whether the declared support names concrete findings, methods or results that the idea needs, or whether it is template rhetoric (``we extend prior work on X'') that would read the same with the citations deleted. The grounding score of an idea is the min--max rescaling
\begin{equation}
\label{eq:grounding}
g(u)=\frac{\text{dependence}(u)-1}{4}\in\{0,\,0.25,\,0.5,\,0.75,\,1\},
\end{equation}
and the grounding of a model is the mean of $g(u)$ over all its generated ideas. The full prompt is in Prompt~\ref{prompt:grounding}.


\textbf{Inference parameters.} Grounding is judged with \texttt{gemini-3.5-flash}; temperature $0.0$; the thinking budget is set to $0$.

\paragraph{Validating Grounding.} On $100$ generated ideas ($50$ from \texttt{qwen3-8b} and $50$ from \texttt{claude-opus-4-8}) the \texttt{gemini-3.5-flash} and annotator scores are moderately correlated (Spearman $\rho{=}0.46$, quadratic-weighted $\kappa{=}0.26$; $62\%$ within one point). 

\paragraph{Why not reward grounding?} We do not put grounding in the training reward, for two reasons. First, optimizing three objectives in one run proved unstable. In our initial runs, grounding entered the reward as a graded multiplier and later as a binary gate, and the policy gamed the judge's category preferences (for example, the grounding-type distribution collapsed onto a single type). Second, grounding has no natural direction in which higher is always better. Our score measures how much an idea depends on its cited sources, so a highly grounded idea can be a simple, incremental extension of them, while a less grounded one can be a promising departure that is only loosely motivated by the documents. We therefore instruct the model at the prompt level to name the sources each idea draws on (Appendix~\ref{app:data_format}), and track grounding separately as a diagnostic (Section~\ref{sec:main_results}). This is a limitation of our objective: some drop in grounding is expected when ideas move away from their sources, but the drop after RL (Table~\ref{tab:final-si-n1-all}) is large, because nothing in the reward penalizes drifting, and finding a stable way to enforce a floor is future work.

\subsection{Creative Utility} CU orders a set greedily: the first idea $u_{(1)}$ is the one with the highest quality, and each later idea is the one that maximizes its marginal value $m_t=q(u_{(t)})\cdot\min_{j<t}d(u_{(t)},u_{(j)})$, with $m_1=q(u_{(1)})$. With patience $\gamma\in(0,1]$, CU is the patience-weighted sum of these marginals, maximized over orderings $\tau$ of the set,

 
\begin{equation}
\label{eq:cu}
CU(U)=\max_{\tau}\sum_{i=1}^{|U|}\gamma^{\,i-1}\,q\big(u_{\tau(i)}\big)\min_{j<i} d\big(u_{\tau(i)},u_{\tau(j)}\big),
\end{equation}

where $\tau$ ranges over orderings of the set, $q$ is quality, $d$ is distance (with $\min_{j<1}d:=1$), and $\gamma\in(0,1]$ is a patience discount on later ideas. We use $\gamma=0.9$, and divide by $\sum_{i}\gamma^{i-1}$ so that scores lie in $[0,1]$ (the normalized CU we report). 

\textbf{Why greedy reporting is sufficient.} Maximizing over orderings is combinatorial, so, following CREATE, we approximate it with the greedy (farthest-point) order: take the highest-quality idea first, then repeatedly take the idea with the largest marginal value $q(u)\min_{j\in\text{picked}}d(u,u_j)$. Because the greedy order is one particular ordering, its value is a lower bound on $CU(U)$, so it can only understate a set's CU. It also depends only on the set and not on the order in which a model happened to write its ideas, which is what we want when comparing models, especially since we have baselines where we \textit{sample} from a model instead of generating a set. 

CU differs from the training reward $R_{set}$ in three ways: (i) $R_{set}$ takes the generation order instead of the best ordering; (ii) it applies no patience discount ($\gamma=1$) and is unnormalized; and (iii) CU is computed with the evaluation judges rather than the training judges.

\section{Benchmarks}
\label{app:benchmarks}
\paragraph{GiantsBench}: For GiantsBench, we run our models with our training prompt, and GIANTS-4B with its own in-distribution prompt. Because the two prompts share only the \textit{claim} field, we compare the two systems at the \textit{claim} level.

\paragraph{IdeaSeed}: IdeaSeed contains 11,683 source papers from two domains. Each paper comes with a proposal-style statement of its idea and 1–7 related works that directly influenced it. We use a 500-paper subset. We first drop papers with fewer than three related works, so that every instance requires synthesizing several source documents. We then sample 250 papers per domain, stratified by related-work count. For this dataset, we report (i) the normalized entropy $H_{\mathrm{norm}}=H(p)/\log_2 |\mathcal{L}|$, where $p$ is the pooled label distribution and $|\mathcal{L}|=8$ is the number of labels on the axis. It lies in $[0,1]$, and higher values mean ideas are spread more evenly across types. We also report (ii) the total variation distance (TVD) between the model's label distribution and the human proposals' distribution, and, (iii) Jensen–Shannon divergence (JSD) \citep{lin1991divergence}. For TVD and JSD, lower values mean the model's spread of idea types is closer to the human one.



\section{Results}
\label{app:results}

\paragraph{All results} Table~\ref{tab:final-si-n1-all} has all the results along with ``number of instances with at least one winning instance''. We see that our proposed model has the highest win-rate and number of instances with at least one win. While some of the competing objectives also achieve similar performance on the win-rate, they collapse on the diversity metric, thus pulling down the overall utility of the set.

\begin{table}[t]
\centering
\small
\renewcommand{\tabcolsep}{1mm}
\caption{Results on our contamination-free test set (1{,}740 instances), with two extra columns: number of instances which at least one win in their set and grounding. Values in parentheses are standard deviations across instances.}
\label{tab:final-si-n1-all}
\begin{tabular}{lccccc}
\toprule
Model & Utility (CU) $\uparrow$ & $\geq$1 Win $\uparrow$ & Grounding ($g$) $\uparrow$ & Win Rate ($q$) $\uparrow$ & Avg. Distance ($d$) $\uparrow$ \\
\midrule
\multicolumn{6}{c}{General models} \\
\midrule
Claude Opus 4.8 & $0.38_{(0.22)}$ & $0.38_{(0.48)}$ & $\mathbf{0.98_{(0.08)}}$ & 0.21 & $0.80_{(0.07)}$ \\
\rowcolor{gray!20}
GPT-6 Astra & $0.28_{(0.24)}$ & $0.27_{(0.44)}$ & $0.94_{(0.11)}$ & 0.15 & $0.86_{(0.06)}$ \\
Nemotron 120B & $0.14_{(0.18)}$ & $0.08_{(0.28)}$ & $0.90_{(0.18)}$ & 0.05 & $0.79_{(0.10)}$ \\
\rowcolor{gray!20}
Qwen3-8B & $0.03_{(0.10)}$ & $0.02_{(0.13)}$ & $0.91_{(0.21)}$ & 0.01 & $0.80_{(0.11)}$ \\
\midrule
\addlinespace[2pt]
\multicolumn{6}{c}{Competing objectives} \\
\midrule
DARLING & $0.52_{(0.17)}$ & $0.59_{(0.49)}$ & $0.09_{(0.20)}$ & 0.38 & $0.67_{(0.13)}$ \\
\rowcolor{gray!20}
Impact-only RL & $0.38_{(0.18)}$ & $0.63_{(0.48)}$ & $0.74_{(0.23)}$ & 0.43 & $0.38_{(0.17)}$ \\
Impact+diversity, no SFT & $0.51_{(0.26)}$ & $0.52_{(0.50)}$ & $0.19_{(0.28)}$ & 0.30 & $\mathbf{0.92_{(0.07)}}$ \\
\midrule
\addlinespace[2pt]
\multicolumn{6}{c}{SFT only} \\
\midrule
SFT (no think) & $0.21_{(0.19)}$ & $0.14_{(0.34)}$ & $0.90_{(0.17)}$ & 0.07 & $0.77_{(0.09)}$ \\
\rowcolor{gray!20}
SFT (think) & $0.16_{(0.18)}$ & $0.11_{(0.32)}$ & $0.91_{(0.17)}$ & 0.06 & $0.76_{(0.10)}$ \\
\midrule
\addlinespace[2pt]
\multicolumn{6}{c}{Proposed models (SFT + RL)} \\
\midrule
Proposed (no think) & $\mathbf{0.64_{(0.23)}}$ & $\mathbf{0.78_{(0.42)}}$ & $0.61_{(0.31)}$ & \textbf{0.50} & $0.88_{(0.07)}$ \\
\rowcolor{gray!20}
Proposed (think) & $0.51_{(0.23)}$ & $0.62_{(0.49)}$ & $0.81_{(0.22)}$ & 0.34 & $0.83_{(0.08)}$ \\
\bottomrule
\end{tabular}
\end{table}

\paragraph{Sampling vs Set Level Prompting} 

\textbf{Gains comes from training.} To separate gains from training and set-level prompting, we compare each general model's set of $K=3$ ideas against $n=3$ independently sampled single-idea responses (Table~\ref{tab:sampled-vs-set}). Set-level prompting raises distance sharply (0.24--0.53 to 0.79--0.86) but lowers quality for the closed-source models (e.g.\ win rate 0.33 to 0.21 for Claude Opus 4.8), so utility barely changes (0.34 to 0.38). Our model beats both modes on both axes, with a win rate of 0.50 and a distance of 0.88, so the balance comes from the set-level reward.

Table~\ref{tab:sampled-vs-set} reports the prompted sampled vs set level results. We see that prompting for a set mostly enables the models to increase their diversity/average distance, but this generally comes at a quality cost, indicating a quality-diversity trade off. For example, for Claude, the distance goes up from 0.24 to 0.80, but the win-rate drops from 0.33 to 0.21, even though the overall utility goes up (because of the big jump in distance). But this table is also to show that the major gains in the results for our proposed trained model, come from the reward training and not only the set-level prompting.

\begin{table}[t]
\centering
\small
\renewcommand{\tabcolsep}{1mm}
\caption{Results on our contamination-free test set (1{,}740 instances). \emph{Set} draws all $K=3$ ideas in one response (as in Table~\ref{tab:final-si-n1-all}); \emph{sampled} draws them as three independent single-idea responses. We can see most models trade off quality for diversity when switching from sampling to set-based generation. Values in parentheses are standard deviations across instances.}
\label{tab:sampled-vs-set}
\begin{tabular}{lccccc}
\toprule
Model & Utility (CU) $\uparrow$ & $\geq$1 Win $\uparrow$ & Grounding ($g$) $\uparrow$ & LC Win Rate ($q$) $\uparrow$ & Avg. Distance ($d$) $\uparrow$ \\
\midrule
\multicolumn{6}{l}{Claude Opus 4.8} \\
Sampled ($k{=}1,n{=}3$) & $0.34_{(0.14)}$ & $0.59_{(0.49)}$ & $0.99_{(0.06)}$ & 0.33 & $0.24_{(0.21)}$ \\
\rowcolor{gray!8}
Set ($k{=}3,n{=}1$) & $0.38_{(0.22)}$ & $0.38_{(0.48)}$ & $0.98_{(0.08)}$ & 0.21 & $0.80_{(0.07)}$ \\
\midrule
\addlinespace[2pt]
\multicolumn{6}{l}{GPT-6 Astra} \\
Sampled ($k{=}1,n{=}3$) & $0.25_{(0.17)}$ & $0.37_{(0.48)}$ & $0.96_{(0.09)}$ & 0.23 & $0.28_{(0.23)}$ \\
\rowcolor{gray!8}
Set ($k{=}3,n{=}1$) & $0.28_{(0.24)}$ & $0.27_{(0.44)}$ & $0.94_{(0.11)}$ & 0.15 & $0.86_{(0.06)}$ \\
\midrule
\addlinespace[2pt]
\multicolumn{6}{l}{Nemotron 120B} \\
Sampled ($k{=}1,n{=}3$) & $0.15_{(0.16)}$ & $0.12_{(0.32)}$ & $0.93_{(0.16)}$ & 0.07 & $0.53_{(0.19)}$ \\
\rowcolor{gray!8}
Set ($k{=}3,n{=}1$) & $0.14_{(0.18)}$ & $0.08_{(0.28)}$ & $0.90_{(0.18)}$ & 0.05 & $0.79_{(0.10)}$ \\
\midrule
\addlinespace[2pt]
\multicolumn{6}{l}{Qwen3-8B} \\

Sampled ($k{=}1,n{=}3$) & $0.04_{(0.08)}$ & $0.02_{(0.13)}$ & $0.92_{(0.18)}$ & 0.01 & $0.29_{(0.17)}$ \\
\rowcolor{gray!8}
Set ($k{=}3,n{=}1$) & $0.03_{(0.10)}$ & $0.02_{(0.13)}$ & $0.91_{(0.21)}$ & 0.01 & $0.80_{(0.11)}$ \\
\bottomrule
\end{tabular}
\end{table}

\begin{figure}
    \centering
    \includegraphics[scale=0.22,trim=1cm 5cm 0 0]{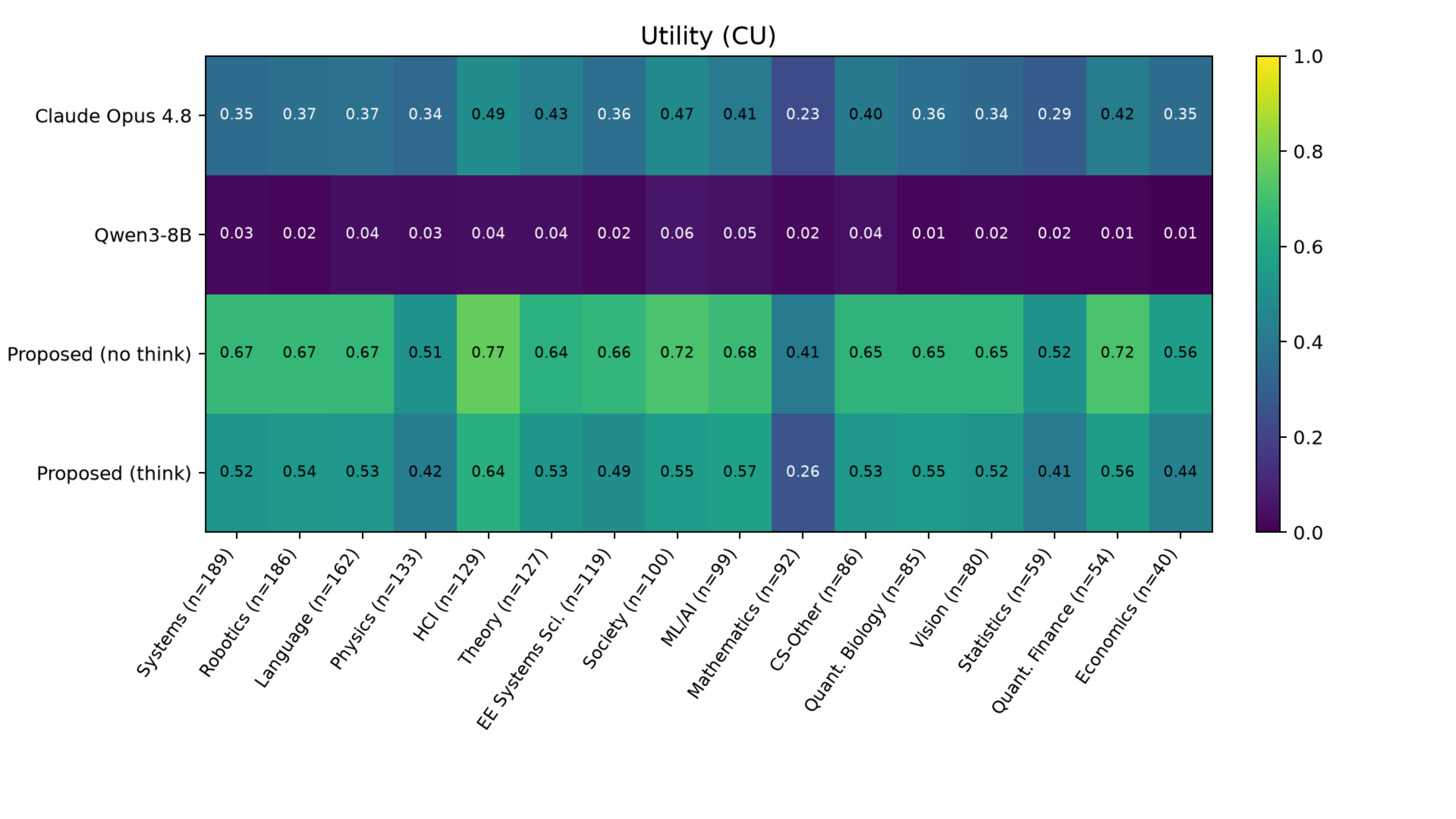}
    \caption{Creative Utility(CU) for two baselines and our proposed models, across 16 domains in our test-set.}
    \label{fig:per_domain_cu}
\end{figure}

\paragraph{Performance across different domains} Figure~\ref{fig:per_domain_cu} reports the split of performance (creative utility, CU) across the 16 domains in our constructed test set. We see that our model is able to maintain high utility across domains (with the exception of a dip for Mathematics for our proposed model with thinking).

\begin{table}[t]
  \centering
  \small
  \renewcommand{\tabcolsep}{1mm}
  \caption{Human study results broken down by the ACL Rolling Review (ARR) subarea of the target paper (\S\ref{sec:human}).}
  \begin{tabular}{lcccc}
    \toprule
    ARR Area & $n$ papers (\%) & Promising ideas $\uparrow$ & Unique ideas $\uparrow$ & Unique promising ideas $\uparrow$ \\
    & & \multicolumn{3}{l}{\footnotesize (Qwen3-8B / Claude Opus 4.8 / Ours per column)} \\
    \midrule
    \makecell[l]{Human-Centered NLP\\and Human-AI Interaction} & 18 (30\%) & $1.67/\textbf{2.39}/\textbf{2.33}$ & $2.56/2.78/\textbf{2.83}$ & $1.61/2.28/2.11$ \\
    LLM agents & 11 (18\%) & $1.36/\textbf{2.00}/\textbf{2.00}$ & $2.36/\textbf{2.82}/\textbf{2.82}$ & $1.18/1.91/\textbf{2.00}$ \\
    Machine Learning for NLP & 7 (12\%) & $2.14/2.14/\textbf{2.43}$ & $\textbf{2.86}/2.71/\textbf{2.86}$ & $2.14/2.14/\textbf{2.29}$ \\
    Resources and Evaluation & 5 (8\%) & $1.40/\textbf{2.20}/1.40$ & $2.20/\textbf{3.00}/2.80$ & $1.20/\textbf{2.20}/1.40$ \\
    Ethics, Bias, and Fairness & 5 (8\%) & $0.80/\textbf{2.80}/2.40$ & $2.40/2.40/\textbf{2.80}$ & $0.60/1.80/\textbf{2.20}$ \\
    Other & 14 (23\%) & $1.43/1.93/\textbf{2.14}$ & $2.57/2.57/\textbf{2.86}$ & $1.36/1.79/\textbf{2.07}$ \\
    \bottomrule
  \end{tabular}
  
  \label{tab:human-area}
\end{table}

\paragraph{Transfer of recipe to other models} In Table ~\ref{tab:final-si-n1-generalization} we report that our proposed objective and training recipe generalizes to other model families as well, specifically Olmo-3-7B-Think-SFT.

\begin{table}[t]
\centering
\small
\renewcommand{\tabcolsep}{1mm}
\caption{Results on our contamination-free test set (1{,}740 instances), showing the recipe applied to Olmo-3-7B-Think-SFT. Values in parentheses are standard deviations across instances.}
\label{tab:final-si-n1-generalization}
\begin{tabular}{lccccc}
\toprule
Model & Utility (CU) $\uparrow$ & $\geq$1 Win $\uparrow$ & Grounding ($g$) $\uparrow$ & LC Win Rate ($q$) $\uparrow$ & Avg. Distance ($d$) $\uparrow$ \\
\midrule
Base & $0.10_{(0.16)}$ & $0.06_{(0.25)}$ & $0.73_{(0.27)}$ & 0.03 & $0.79_{(0.12)}$ \\
\rowcolor{gray!20}
SFT (no think) & $0.17_{(0.18)}$ & $0.11_{(0.31)}$ & $0.84_{(0.22)}$ & 0.05 & $0.77_{(0.10)}$ \\
Proposed (no think) & $0.56_{(0.23)}$ & $0.67_{(0.47)}$ & $0.66_{(0.27)}$ & 0.39 & $0.83_{(0.09)}$ \\
\rowcolor{gray!20}
SFT (think) & $0.12_{(0.15)}$ & $0.06_{(0.23)}$ & $0.86_{(0.22)}$ & 0.03 & $0.75_{(0.10)}$ \\
Proposed (think) & $0.54_{(0.24)}$ & $0.63_{(0.48)}$ & $0.73_{(0.25)}$ & 0.34 & $0.83_{(0.08)}$ \\
\bottomrule
\end{tabular}
\end{table}


\paragraph{Grounding Trade-Off} Table~\ref{tab:position} shows how win-rate and grounding change for ideas as you go down a generated-set of ideas. For our trained models, the impact doesn't seem to have that much of a change (especially for the no-think model). The think model is more instruction following and can explicitly check for the instructed ordering when generating so we see that it generates the most impactful idea first. For grounding, we see the models get less grounded as they generate the next idea. The models choose to do a \textit{simpler} way of getting different ideas by moving away from the source. This trend is more prominent for the no think model compared to the think model. 

\begin{table}
\small
\centering
\caption{How the Impact/Win-rate and grounding changes with the position at which an idea is generated. We report the raw win-rate (instead of the strict length-controlled) here instead of length normalized }
\begin{tabular}{lcccccc}
\toprule
 & \multicolumn{3}{c}{Win-Rate} & \multicolumn{3}{c}{Grounding ($g$)} \\
\cmidrule(lr){2-4}\cmidrule(lr){5-7}
Model & 1 & 2 & 3 & 1 & 2 & 3 \\
\midrule
Proposed (no think) & 0.69 & 0.72 & 0.65 & 0.79 & 0.56 & 0.48 \\
Proposed (think) & 0.62 & 0.54 & 0.52 & 0.88 & 0.80 & 0.75 \\
\bottomrule
\end{tabular}
\label{tab:position}
\end{table}

\paragraph{Think vs no think}
From Table~\ref{tab:final-si-n1-main} we see that Proposed (no think) has a higher utility than Proposed (think). To see if this is because of the data curation, we use Claude-Sonnet-4.6 to generate \textit{forward thinking traces} for the entire training set (forward-full). This model achieves a CU of 0.50, a length controlled win-rate of 0.30 and an average distance of 0.87. This also doesn't beat our no-think baseline. Any filtering on the full-dataset for high quality and diverse set lead to substantially less instances for the SFT stage (around 1k) and there the size of the training data became another factor and it was difficult to isolate just training data structuring as a factor.


\paragraph{Human study}
\label{app:human_assignment}

Each of our four annotators first selects the ACL Rolling Review (ARR) subareas they are comfortable reviewing (e.g.\ LLM agents, Human-Centered NLP and Human-AI Interaction, Ethics, Bias, and Fairness). Independently, every paper in our test set is tagged with its ARR subarea by prompting an LLM classifier with the target paper's title and abstract. We then filter the test set down to the papers whose tag falls within at least one annotator's chosen subareas, and assign each of these papers to an annotator by matching the paper's tag to that annotator's selected subareas, so every rated portfolio falls within its annotator's stated area of expertise. 

Table~\ref{tab:human-area} reports the resulting per-area breakdown. Each cell reports the three systems' means as Qwen3-8B / Claude Opus 4.8 / Ours. All three metrics range from 0 to 3. $n$ is the number of source papers tagged with that area (out of 60), and its share of the evaluation set. Areas with fewer than 5 papers are pooled into \emph{Other} (Computational Social Science and Cultural Analytics, Dialogue and Interactive Systems, Generation, NLP Applications, NLP and Code Models, Question Answering, Sentiment Analysis, Stylistic Analysis, and Argument Mining). We observe that ours has the highest or tied-highest unique ideas in five of the six areas (Claude Opus 4.8 leads on Resources and Evaluation), but Claude Opus 4.8 has more promising ideas in Human-Centered NLP, Resources and Evaluation, and Ethics, Bias, and Fairness.

\begin{table}[t]
\centering
\small
\caption{Human annotations per annotator. For each portfolio of $K=3$ ideas, annotators marked the ideas that are worthwhile, the distinct representatives, and the distinct representatives that are also worthwhile. We report the mean number per portfolio (standard deviation in parentheses); $n$ is the number of portfolios.}
\label{tab:human-pointwise-per-annotator}
\begin{tabular}{ll|c|ccc}
\toprule
Annotator & System & $n$ & Promising & Unique & Unique \& Promising \\
\midrule
Annotator 1 & Qwen3-8B & 15 & $1.73_{(0.88)}$ & $2.67_{(0.49)}$ & $1.67_{(0.90)}$ \\
 & Claude Opus 4.8 & 15 & $1.93_{(0.59)}$ & $2.67_{(0.49)}$ & $1.87_{(0.64)}$ \\
 & Proposed & 15 & $2.20_{(0.56)}$ & $2.87_{(0.35)}$ & $2.20_{(0.56)}$ \\
\midrule
Annotator 2 & Qwen3-8B & 15 & $1.27_{(0.46)}$ & $2.40_{(0.74)}$ & $1.13_{(0.35)}$ \\
 & Claude Opus 4.8 & 15 & $1.93_{(0.80)}$ & $2.80_{(0.56)}$ & $1.87_{(0.83)}$ \\
 & Proposed & 15 & $1.67_{(0.49)}$ & $2.73_{(0.46)}$ & $1.53_{(0.52)}$ \\
\midrule
Annotator 3 & Qwen3-8B & 15 & $1.07_{(0.96)}$ & $2.53_{(0.64)}$ & $0.87_{(0.74)}$ \\
 & Claude Opus 4.8 & 15 & $2.33_{(0.72)}$ & $2.53_{(0.52)}$ & $1.80_{(0.68)}$ \\
 & Proposed & 15 & $2.40_{(0.83)}$ & $2.87_{(0.35)}$ & $2.13_{(0.92)}$ \\
\midrule
Annotator 4 & Qwen3-8B & 15 & $2.00_{(0.65)}$ & $2.47_{(0.64)}$ & $2.00_{(0.65)}$ \\
 & Claude Opus 4.8 & 15 & $2.60_{(0.63)}$ & $2.87_{(0.52)}$ & $2.60_{(0.63)}$ \\
 & Proposed & 15 & $2.40_{(0.74)}$ & $2.87_{(0.35)}$ & $2.33_{(0.72)}$ \\
\midrule
\midrule
All annotators & Qwen3-8B & 60 & $1.52_{(0.83)}$ & $2.52_{(0.62)}$ & $1.42_{(0.81)}$ \\
 & Claude Opus 4.8 & 60 & $2.20_{(0.73)}$ & $2.72_{(0.52)}$ & $2.03_{(0.76)}$ \\
 & Proposed & 60 & $2.17_{(0.72)}$ & $2.83_{(0.38)}$ & $2.05_{(0.75)}$ \\
\bottomrule
\end{tabular}
\end{table}

\begin{figure}
    \centering
    \includegraphics[scale=0.33,trim=1cm 21cm 10cm 0]{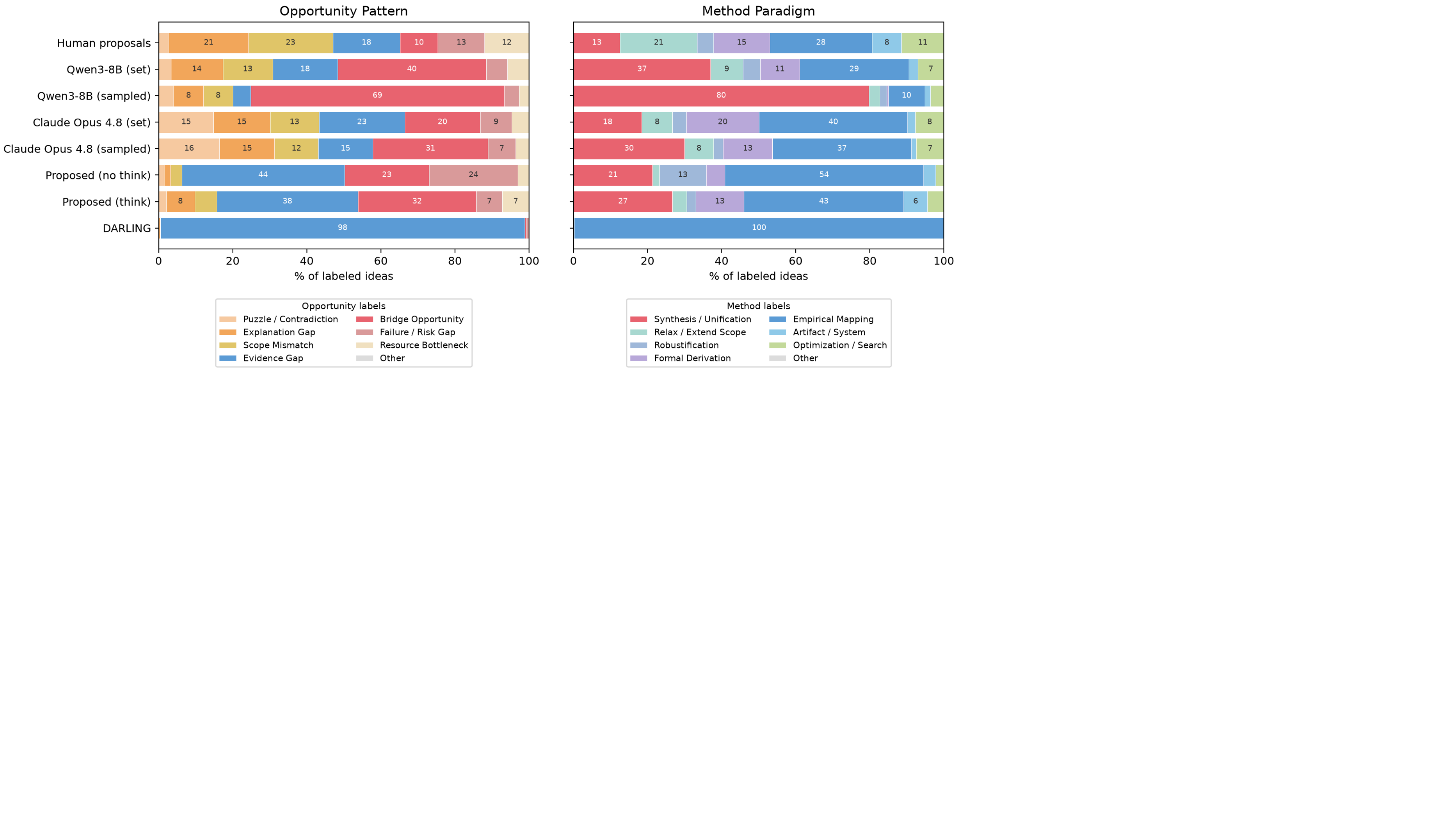}
    \caption{Distribution of `opportunity' and `method' labels on the IdeaSeed benchmark, as generated by different models compared to the proposed model.}
    \label{fig:ideaseed}
\end{figure}

\paragraph{IdeaSeed Benchmark}  We report the full performance on the IdeaSeed Benchmark in Table~\ref{tab:ideaseed-entropy} and show the distribution of labels across the population of ideas in Figure~\ref{fig:ideaseed}. Similar to the results in their original paper, we see the model anchoring to a few categories. The prompted baselines also have higher entropy on the Opportunity axis (0.89 for Claude Opus 4.8 and 0.80 for Qwen3-8B, against 0.67 for Proposed (no think) and 0.75 for Proposed (think)), which suggests RL may have reduced population-level type diversity on both axes. Further work is needed to see how to translate some of the diversity training measures to population level diversity while maintaining quality. 

\begin{table}[t]
  \centering
  \small
  \footnotesize
  \renewcommand{\tabcolsep}{1mm}
  \caption{Taxonomy diversity on IdeaSeed (500 papers: 250 ML, 250 natural
  science). Every generated idea and every human proposal is labeled by
  \texttt{GPT-5.4-mini} with one Opportunity Pattern and one Method Paradigm
  from the TasteGap taxonomy~\citep{chen2026measuring}.}
  \begin{tabular}{l|ccc|cc}
    \toprule
    Method & $H_{\mathrm{norm}}$ $\uparrow$ & TVD vs h $\downarrow$ & JSD vs h $\downarrow$ & \% unique/instance $\uparrow$ & Ideas \\
    \midrule
    \multicolumn{6}{l}{\textit{Opportunity types}} \\
    Claude Opus 4.8 (sampled, $k=1,n=3$) & 0.86 & 0.35 & 0.12 & 63.4\% (n=498) & 1494 \\
    Qwen3-8B (sampled, $k=1,n=3$) & 0.56 & 0.60 & 0.30 & 54.6\% (n=500) & 1500 \\
    Claude Opus 4.8 (set, $k=3,n=1$) & 0.89 & 0.27 & 0.08 & 89.0\% (n=499) & 1497 \\
    Qwen3-8B (set, $k=3,n=1$) & 0.80 & 0.30 & 0.10 & 89.3\% (n=499) & 1275 \\
    DARLING (sampled, $k=1,n=3$) & 0.05 & 0.80 & 0.58 & 35.0\% (n=500) & 1500 \\
    Proposed (no think) (set, $k=3,n=1$) & 0.67 & 0.50 & 0.23 & 87.0\% (n=500) & 1500 \\
    Proposed (think) (set, $k=3,n=1$) & 0.75 & 0.42 & 0.14 & 83.0\% (n=500) & 1502 \\
    Human proposals & 0.88 & -- & -- & -- & 500 \\
    \addlinespace[2pt]
    \midrule
    \multicolumn{6}{l}{\textit{Method types}} \\
    Claude Opus 4.8 (sampled, $k=1,n=3$) & 0.74 & 0.27 & 0.08 & 58.0\% (n=498) & 1494 \\
    Qwen3-8B (sampled, $k=1,n=3$) & 0.38 & 0.67 & 0.38 & 47.8\% (n=500) & 1500 \\
    Claude Opus 4.8 (set, $k=3,n=1$) & 0.77 & 0.23 & 0.05 & 84.2\% (n=499) & 1497 \\
    Qwen3-8B (set, $k=3,n=1$) & 0.77 & 0.27 & 0.08 & 88.7\% (n=499) & 1275 \\
    DARLING (sampled, $k=1,n=3$) & 0.01 & 0.72 & 0.51 & 33.5\% (n=500) & 1500 \\
    Proposed (no think) (set, $k=3,n=1$) & 0.65 & 0.43 & 0.17 & 79.0\% (n=500) & 1500 \\
    Proposed (think) (set, $k=3,n=1$) & 0.73 & 0.30 & 0.09 & 81.0\% (n=500) & 1502 \\
    Human proposals & 0.87 & -- & -- & -- & 500 \\
    \bottomrule
  \end{tabular}
  \label{tab:ideaseed-entropy}
\end{table}

\section{AutoResearch}
\label{app:autoresearch}

In this section we expand on the autoresearch experiment from Section~\ref{sec:autoresearch}. 

We walk through one run in full. The idea is a diagnostic benchmark for memory-augmented LLMs, generated by our think model. Its core claim has two parts: the suite \emph{reveals which memory mechanisms genuinely store information}, and it \emph{exposes a failure mode shared across many designs}. The card is shown below.

\begin{outputbox}[Idea Card for AutoResearch: Diagnostic Benchmark for Memory-Augmented LLMs]
\textbf{Problem gap} \textit{[Evidence Gap]} The field of memory augmentation for
LLMs is growing rapidly, but there is no standardized, systematic way to evaluate
whether memory mechanisms actually store semantic information, support effective
retrieval, or remain stable across models and tasks, so claims about a mechanism's
quality are largely unverified and the community lacks a shared benchmark for comparison.

\medskip
\textbf{Core claim} A diagnostic benchmark measuring whether memory representations
preserve semantic content, support contrastive retrieval, are interpretable, and remain
stable across a suite of existing memory mechanisms will reveal which mechanisms
genuinely store information and expose a common failure mode shared across many proposed
designs, enabling the field to audit and advance itself with measurable evidence rather
than untested claims.

\medskip
\textbf{Method} \textit{[Empirical Mapping]} Construct an open, modular diagnostic
benchmark for memory-augmented LLMs, consisting of tasks that probe:
\begin{itemize}
  \item \textbf{content fidelity}: do memory vectors encode task-relevant semantics?
  \item \textbf{contrastive retrieval quality}: does retrieval match ground-truth memory?
  \item \textbf{interpretability}: can the stored content be inspected?
  \item \textbf{stability}: does memory survive parameter perturbation and cross-model transfer?
\end{itemize}
Evaluate leading memory mechanisms (DPR-style dense retrievers, SBERT-style
bi-encoders, latent memory, compressed soft-prompt memory, retraining-free long-term
memory) on this benchmark and release it as a public tool. Map which mechanisms succeed
on which subtasks and quantify the gap between them, turning anecdotal claims into
comparable, reproducible evidence.

\end{outputbox}

Our coding agent comes up with the following proof of concept. The agent sharpens the claim's four named properties (content fidelity, contrastive retrieval, interpretability, stability) into metrics, and chooses seven CPU-runnable mechanisms from the families named in the method, plus two null controls. It supplies the datasets (HotpotQA and NQ-Swap) and the thresholds. Its first plan is:

\begin{inputbox}[PoC Design]
\promptlabel{poc:memory_benchmark}
\raggedright

\textbf{Data} (both real)
\begin{itemize}\itemsep0pt\topsep2pt
\item \textbf{HotpotQA (distractor)} -- \texttt{hotpotqa/hotpot\_qa}: each example gives a question and 10 paragraphs (2 gold, 8 topically related distractors), forming a hard-negative memory pool. Target $\sim$300 questions / $\sim$3000 paragraphs, subsampled to fit CPU.
\item \textbf{NQ-Swap} -- \texttt{pminervini/NQ-Swap}: original vs.\ entity-substituted contexts for the same question, a counterfactual content-fidelity probe (does the memory encode \emph{which} entity was stored?).
\end{itemize}

\vspace{4pt}
\textbf{Mechanisms} (memory = what the design stores for one paragraph; reader = its own retrieval op)

\vspace{2pt}
{\setlength{\tabcolsep}{3pt}\renewcommand{\arraystretch}{1.15}
\begin{tabular}{@{}>{\raggedright\arraybackslash}p{0.19\linewidth}
                   >{\raggedright\arraybackslash}p{0.46\linewidth}
                   >{\raggedright\arraybackslash}p{0.31\linewidth}@{}}
\toprule
\textbf{Arm} & \textbf{Memory (per paragraph)} & \textbf{Reader} \\
\midrule
dpr & \texttt{facebook/dpr-ctx\_encoder} embedding & \texttt{dpr-question\_encoder} (asymmetric dual encoder) \\
sbert-minilm & \texttt{all-MiniLM-L6-v2} embedding & bi-encoder cosine \\
sbert-mpnet & \texttt{all-mpnet-base-v2} embedding & bi-encoder cosine \\
latent-lm & mean-pooled last hidden layer of a frozen decoder LM (Qwen2.5-0.5B / GPT-2) & cosine \\
softprompt-k4 & same LM states segment-pooled into $k{=}4$ slots (ICAE-style) & slot-max over query \\
text-bm25 & the paragraph text (retraining-free long-term memory) & TF-IDF / BM25 \\
text-llmsummary & one-sentence LLM summary (nova-micro or Haiku); lossy text memory & TF-IDF \\
\midrule
null-random & random vector & (control) \\
rand-init-encoder & untrained BERT-architecture mean pool & (control) \\
\bottomrule
\end{tabular}}

\vspace{6pt}
\textbf{Axes and metrics} (each with a floor)

\vspace{2pt}
{\setlength{\tabcolsep}{3pt}\renewcommand{\arraystretch}{1.15}
\begin{tabular}{@{}>{\raggedright\arraybackslash}p{0.19\linewidth}
                   >{\raggedright\arraybackslash}p{0.50\linewidth}
                   >{\raggedright\arraybackslash}p{0.27\linewidth}@{}}
\toprule
\textbf{Axis} & \textbf{Metric} & \textbf{Floor} \\
\midrule
Fidelity-A & Ridge/logistic linear probe from memory $\to$ multi-label indicator of the stored paragraph's content words (top-$V$ vocab); held-out recall@$k$ & label-frequency prior \\
Fidelity-B (counterfactual) & NQ-Swap: accuracy of picking the original vs.\ entity-swapped passage memory given a query naming the true answer entity & 0.5 \\
Retrieval & Recall@1 and MRR of a gold paragraph within its 10-paragraph hard-negative pool & 0.2 (2 gold of 10) \\
Interpretability & Training-free top-20 vocabulary decoding of the memory vector vs.\ the stored paragraph's content words & frequency-matched random word list \\
Stability-1 & Recall@1 retention after Gaussian noise on encoder weights ($\sigma\in\{0.005,0.01,0.02\}\times$ per-tensor std), as ratio to clean & -- \\
Stability-2 & Cross-model portability: ridge map fit on 500 held-out paragraphs from mechanism A's memory space to B's, then retrieve with B's reader; ratio to B's native recall@1 & -- \\
\bottomrule
\end{tabular}}

\vspace{6pt}
\textbf{Supports} if the controls sit at floor on Fidelity-A/Retrieval while real mechanisms are clearly above, real mechanisms' per-axis scores separate beyond bootstrap 95\% CIs on $\geq$2 axes, cross-split Spearman $\rho \geq 0.7$, and $\geq$4/7 real mechanisms sit at floor on some axis.

\textbf{Refutes} if the suite cannot separate null from real memory, the ranking is split-unstable, or no axis is commonly failed.

\textbf{Secondary analysis.} Rank correlation \emph{between} axes: if retrieval recall@1 predicts every other axis, the suite adds nothing beyond the retrieval-only evaluation the \texttt{problem\_gap} complains about.

\end{inputbox}

The first run the agent did answered one conjunct and left the other open. The tests separated real mechanisms from null memory, but the NQ-Swap probe did not show a failure. The dense encoders scored 0.87--0.96, and only the untrained latent arms fell to chance. A weight-noise sweep at $\sigma=0.01$--$0.02$ was also invisible. The agent therefore revised its plan, adding three probes (entity-order swap, negation, multi-item capacity) to be run over all arms, and
wrote its decision rules down before running them:

\begin{inputbox}[PoC Plan v2: Pre-registered Search for a Shared Failure Mode]
\promptlabel{poc:memory_benchmark_v2}
\raggedright

\textbf{Learnings from v1} (250 HotpotQA pools, 2479 stored paragraphs, 300 NQ-Swap pairs)

\vspace{2pt}
{\setlength{\tabcolsep}{3pt}\renewcommand{\arraystretch}{1.15}
\begin{tabular}{@{}>{\raggedright\arraybackslash}p{0.27\linewidth}
                   >{\centering\arraybackslash}p{0.14\linewidth}
                   >{\centering\arraybackslash}p{0.17\linewidth}
                   >{\centering\arraybackslash}p{0.15\linewidth}
                   >{\centering\arraybackslash}p{0.17\linewidth}@{}}
\toprule
\textbf{Mechanism} & \textbf{Retrieval r@1} & \textbf{Fidelity probe (floor .12)} & \textbf{Interp.} & \textbf{NQ-Swap acc (chance .5)} \\
\midrule
dpr & 0.712 & 0.429 & 0.257 & 0.870 \\
sbert-minilm & 0.796 & 0.437 & 0.301 & 0.955 \\
sbert-mpnet & 0.792 & 0.450 & 0.282 & 0.947 \\
softprompt-k4 (smoke, 24 ex) & 0.25 & 0.441 & 0.037 & 0.55 \\
text-tfidf (smoke) & 0.83 & 0.557 & 1.00 & 1.00 \\
null-random (smoke) & 0.13 & 0.092 & 0.032 & 0.57 \\
\bottomrule
\end{tabular}}

\vspace{4pt}
Two things changed the design:
\begin{enumerate}\itemsep0pt\topsep2pt
\item \textbf{NQ-Swap is not a shared failure.} DPR/SBERT score 0.87--0.96 because the substituted entity is a literal surface token that lexical-ish dense matching picks up. Only the untrained latent/compressed arms (softprompt-k4 $\approx$ chance) fail, i.e.\ 2 of 7 arms, which is not a ``common failure mode shared across many designs.'' The claim's second conjunct is still untested and needs probes that can plausibly break \emph{all} designs.
\item \textbf{Weight noise at $\sigma$=0.01--0.02 is invisible} (retrieval ratio 1.00 for MiniLM and softprompt). The stability sweep is widened to $\sigma\in\{0.01,0.05,0.2\}$ plus a memory-drift cosine.
\end{enumerate}

\vspace{4pt}
\textbf{What v2 adds} (pre-registered before running). Three candidate shared-failure probes over \textbf{real HotpotQA passages} with programmatic minimal edits, run over \textbf{all 9 arms} (7 mechanisms + 2 controls). All three are reported regardless of outcome.

\vspace{2pt}
{\setlength{\tabcolsep}{3pt}\renewcommand{\arraystretch}{1.15}
\begin{tabular}{@{}>{\raggedright\arraybackslash}p{0.15\linewidth}
                   >{\raggedright\arraybackslash}p{0.53\linewidth}
                   >{\raggedright\arraybackslash}p{0.28\linewidth}@{}}
\toprule
\textbf{Probe} & \textbf{Construction} & \textbf{Metric / chance} \\
\midrule
P1 entity-order swap & Swap two entity mentions A, B throughout the stored passage; query is a real sentence naming them in one order; the correct memory is the passage whose order matches. Both directions per item; exact ties scored 0.5. Bag-of-words text memory is at chance by construction (identical BoW); dense encoders are known to be weak on argument order. & 2AFC accuracy, chance 0.5 \\
P2 negation & Negate the first copular sentence of the stored passage (``is'' $\to$ ``is not''); query is the affirmative or negated sentence; the correct memory is the polarity-matching one. ``not'' is a stop word for the TF-IDF arms, so text memory should also sit at chance. & 2AFC accuracy, chance 0.5 \\
P3 multi-item capacity & A memory unit stores $m\in\{1,2,4\}$ real passages; retrieve the unit containing the gold passage out of 10 units (gold position shuffled). Encoder max\_len grows with $m$, so this measures dilution, not truncation. & r@1 per $m$, retention r@1($m{=}4$)/r@1($m{=}1$); chance 0.1 \\
\bottomrule
\end{tabular}}

\vspace{4pt}
Also adds the \texttt{text-llmsummary} arm (nova-micro one-sentence memory notes, 3075 passages already summarised), run through the full v1 axis suite, so all five mechanism families in the idea's method are covered.

\vspace{4pt}
\textbf{Unchanged.} \texttt{falsifying\_observation}, \texttt{tested\_hypothesis} and \texttt{specificity} are as in plan v1. \texttt{claim\_scope} is unchanged except that the shared-failure conjunct is now addressed by P1--P3 rather than NQ-Swap alone. Coverage stays \texttt{most}: production-scale trained compression/latent-memory checkpoints are out of reach without a GPU, and ``the field will adopt it'' is unfalsifiable.

\vspace{4pt}
\textbf{Decision rules}
\begin{itemize}\itemsep0pt\topsep2pt
\item \textbf{Discriminative validity:} real arms beat both controls on retrieval and fidelity probe beyond bootstrap 95\% CIs.
\item \textbf{Ranking is signal, not noise:} Spearman $\rho \geq 0.7$ between the two disjoint data halves.
\item \textbf{Common failure mode:} $\geq$4/7 non-control arms within $1.5\times$ chance on the \emph{same} probe (P1, P2, or P3 retention $\leq$ 0.5).
\item \textbf{Added value over retrieval-only evaluation} (secondary): cross-axis Spearman over the real arms. If every axis is rank-identical to retrieval r@1, the diagnostic adds nothing beyond the retrieval-only evaluation the \texttt{problem\_gap} complains about.
\end{itemize}

\end{inputbox}

The final verdict was \emph{supports}, at coverage \emph{most}. The agent reports that the suite's rankings agree across disjoint data halves (Spearman 0.96 on retrieval), that content fidelity is nearly orthogonal to retrieval accuracy (Spearman 0.18 over the seven real arms), and that insensitivity to argument order is shared by every design family it tested. We read these claims qualitatively against the agent's own decision rules (an informal check, not a systematic audit), and in places the evidence does not seem to fully meet the verdict. The agent found that its own first probe was invalid, replaced it, and committed to its decision rules in advance. It does mean the \emph{supports} verdict should be read as ``the benchmark's design is worth building'', not ``the claim is established''.

We also show examples of some claims, agent generated verdicts, proof of concept summary and the results from the agents executions in Table~\ref{tab:autoresearch_examples}.


\begin{table*}[t]
\centering
\footnotesize
\renewcommand{\arraystretch}{1.25}
\caption{Example autoresearch cards, one per verdict. Each card is an idea's frozen claim, a scaled-down proof of concept run by the coding agent under a 90-minute, \$30, CPU-only budget, and the agent's structured verdict. Summaries are condensed from the agent's own report.}
\newcolumntype{Y}[1]{>{\raggedright\arraybackslash}p{#1}}
\begin{tabular}{Y{0.10\textwidth} Y{0.19\textwidth} Y{0.25\textwidth} Y{0.34\textwidth}}
\toprule
\textbf{Verdict} & \textbf{Claim} & \textbf{Proof of concept (summary)} & \textbf{Results (summary)} \\
\midrule

\textbf{Supports}\newline
Per-form robustness map for AI-text detectors\newline
{\scriptsize think model; coverage: most}
&
A benchmark of many text-evasion forms shows detectors resist some attacks and fail on others, and the perform map lets practitioners choose a detector per setting.
&
6 detectors $\times$ 12 conditions on human vs.\ ChatGPT text (HC3), 150 texts per class per condition. Four zero-shot statistical scores, a RoBERTa detector and a char n-gram classifier. Forms: paraphrase, style shift, typos and homoglyphs at three strengths, domain shift, regeneration by two other LLMs. TPR at 5\% FPR. Oracle per-form choice replaced by a held-out split-half test (400 splits).
&
Near ceiling on clean text (TPR $0.92$--$0.99$) but divergent by form: log-likelihood $1.00$ under domain shift, $0.11$ under paraphrase, $0.00$ under typos and homoglyphs (non-overlapping CIs). Rankings barely agree across forms (Kendall $\tau=0.095$). Held-out per-form selection beats the best single detector by $+0.107$ TPR (95\% CI $[0.064, 0.137]$, 400/400 splits). Not reached: commercial or watermark detectors, white-box attacks, larger domain grid. \\
\midrule

\textbf{Refutes}\newline
One cross-task faithfulness evaluator\newline
{\scriptsize no-think model; coverage: most}
&
Faithfulness is task-agnostic and measurable uniformly by one lightweight evaluator; models are far more unfaithful than accuracy suggests, with a larger gap for reasoning than retrieval.
&
6 task conditions in one evidence format (GSM8K, KG-QA, HotpotQA, SQuAD, two TAT-QA), two Llama solvers, 430 traces. One fixed evaluator prompt on four lightweight evaluators plus a Claude-Sonnet-5 reference. Ground truth from paired corruption: a value inserted that appears nowhere in the evidence (329 traces).
&
Correctness and grounding are separable, and all evaluators beat chance in all six conditions. But the measurement is not uniform: recall $26.6$--$94.2\%$, false-positive rate $0$--$22\%$, and the headline contrast flips sign between two evaluators. Reasoning-vs-retrieval gap is null within a modality (table $+9.3$ [$0.0$, $20.7$] vs.\ $+1.1$ [$-10.2$, $14.0$]); the effect is concentrated in KG-QA ($43.5\%$ of correct answers unsupported vs.\ $0$--$11.5\%$). A universal claim falsified by a subset. \\
\midrule

\textbf{Inconclusive}\newline
Content-level citation auditing\newline
{\scriptsize no-think model; coverage: most; blocker: noise}
&
Many citation errors are content-level, and an auditor reasoning over the claim and cited text catches them far more reliably than a metadata-only verifier, with no index needed.
&
Citation-Integrity corpus (biomedical, expert labels): 770 test pairs (abstract-level cited text), 399-pair stratified slice, precision reweighted to natural prevalence. Four LLM auditors (sub-claim decomposition then grounding) vs.\ a metadata-only verifier (OpenAlex/Crossref) and TF-IDF.
&
Direction holds: content errors are $21.0\%$ of pairs, and the metadata verifier has no content signal (precision $0.26$ vs.\ $0.21$ base rate). Magnitude does not: AUROC $0.62$--$0.67$ (all errors), $0.73$--$0.79$ (core), best F1 $0.37$--$0.43$, recall $0.50$ only at precision $0.30$; between the pre-registered bars. Auditors agree with each other ($\kappa$ $0.42$--$0.73$) more than with gold ($0.15$--$0.25$), within the published expert band. Not reached: full-text papers, other domains, fine-tuned auditors. \\
\bottomrule
\end{tabular}

\label{tab:autoresearch_examples}
\end{table*}

\section{Prompts}
\label{app:prompts}

\begin{outputbox}[Example Idea Card]
\promptlabel{example:idea_card}
\raggedright
\{

\textbf{"problem\_gap"}: "[Scope Mismatch] Prior work on Persian ezafe recognition relied on CRF and other traditional statistical methods with limited context windows, leaving unclear whether pretrained transformer models can substantially advance ezafe recognition accuracy and whether the established benefit of ezafe information for downstream NLP tasks (parsing, POS tagging) still holds when the downstream model itself is a transformer that may already implicitly capture such syntactic information.",

\textbf{"core\_claim"}: "Pretrained transformer models (BERT, XLMRoBERTa) substantially surpass CRF-based methods for Persian ezafe recognition, but their contextual representations and attention mechanisms already implicitly encode ezafe-like syntactic boundary information, rendering explicit ezafe features redundant for transformer-based POS tagging—unlike for CRF and RNN models, which still benefit from such information.",

\textbf{"method"}: "[Empirical Mapping] To test this, the approach is to frame ezafe recognition and POS tagging as sequence labeling tasks and systematically compare CRF, BiLSTM+CNN, and pretrained transformer models (multilingual BERT, XLMRoBERTa) on the 10-million-word Bijankhan corpus under controlled train/validation/test splits. For POS tagging, ezafe information is injected in two ways—as an input feature from the best ezafe recognizer and as an auxiliary task in a multi-task setting—across all model families, enabling a controlled comparison of which architectures benefit from explicit ezafe and which do not, supplemented by per-POS error analysis and annotation correction to diagnose remaining failure modes.",

\textbf{"inspiration"}: "Document 2 established CRF-based sequence labeling with a context window of 5 as the state-of-the-art for ezafe recognition (98.04\% accuracy), but its reliance on fixed local windows and hand-crafted Persian-specific features left open whether models with global contextual representations could overcome the long-distance dependency limitations it acknowledged. Document 1 demonstrated that ezafe-enriched POS tags improve dependency parsing and chunking across multiple parser types, but tested only traditional parsers (MaltParser, TurboParser, CRF chunker) and never examined whether the benefit of explicit ezafe features persists when the downstream model already possesses rich contextual representations. Connecting these is non-trivial because Document 1's finding that ezafe universally helps syntactic tasks creates an expectation that it should also help transformer-based POS tagging, yet the self-attention mechanism of transformers—which implicitly captures dependency-like relations—may render this explicit signal redundant, a tension that neither document addresses since neither considers pretrained transformers."

\}
\end{outputbox}

\begin{promptbox}[Prompt \thetcbcounter: Reference-Importance Prompt]
\promptlabel{prompt:source_select}
System:

You are an expert research scientist. You will read a target paper and a numbered bibliography in which each reference is pre-annotated with all of its in-text mentions. Identify the SPARKER references -- the small subset that motivates, exposes a gap, supplies the foundation, or is directly extended -- by reading every mention of each ref together. Return only the sparkers in detail; list the IDs of all remaining refs as `other'. Output a single JSON object and nothing else.

User:

TARGET PAPER

Title: \{title\}

Abstract:
\{abstract\}

Body (non-appendix):
\{body\}

REFERENCES (numbered, with all in-text mentions pre-collected)

Each entry below lists a reference followed by every place the target paper
mentions it (section + a snippet of the surrounding paragraph). Citations in
the body appear EITHER as numeric markers (e.g. ``[3]'', ``[3, 7]'') that match
the integer prefix below, OR as author-year strings (e.g. ``Smith et al.
(2020)'', ``Smith and Jones, 2020'') that match the author/year header.

YOUR TASK is selection, not exhaustive scoring:

\begin{itemize}
\item Pick out the SPARKER references -- the small subset of refs that drove the new paper into being.
\item For every other ref (datasets, tools, generic related work, background citations) just record its ID in \texttt{other\_ids}. Do not write a rationale for these.
\end{itemize}

Read EVERY mention of a ref together before judging -- some refs are cited
decoratively in Related Work and load-bearingly in Methods; the load-bearing
role wins.

A ref is a SPARKER if AT LEAST ONE of these labels fits. Tag every label
that applies; multi-label is allowed and encouraged when honest:

\begin{itemize}
\item \texttt{motivates} -- the cited work establishes that the problem matters or exists; the new paper's reason-for-being traces back to it. Test: would the reader still understand why this paper was written if the citation were removed?
\item \texttt{exposes\_gap} -- the cited work has a specific limitation, failure, or open question that the new paper directly targets. Test: does the new contribution map onto a flaw of the cited work? The flaw counts whether the cited work declares it itself (``future work needed...'') OR the new paper empirically shows the cited method/benchmark is saturated, insufficient, or wrong. This can be in the introduction, results sections or the discussion.
\item \texttt{foundation} -- the cited work provides the conceptual frame, formalism, or method the new paper starts from. Test: would the method section need to be rewritten if this cited work didn't exist?
\item \texttt{extends} -- the new paper is a direct continuation, adaptation, or generalization of the cited work -- positioned as ``the next step after X.'' Test: does the paper frame itself as building forward from this specific work?
\end{itemize}

A ref is NOT a sparker ($\rightarrow$ goes in \texttt{other\_ids}) when it is:
\begin{itemize}
\item a dataset / tool / model / implementation the paper merely consumes (PyTorch, scikit-learn, an embedding model, a judge LLM, etc.),
\item generic context-setting / related-work-list padding with no stance.
\end{itemize}

SELF-CONSISTENCY CHECK. Before finalizing a sparker, re-read your own
rationale. If it concedes any of:
\begin{itemize}
\item ``X provides theoretical grounding but the method does not build on X''
\item ``name-checked in related work; not load-bearing''
\end{itemize}
then the ref belongs in \texttt{other\_ids}, not \texttt{sparkers}. The rubric tests are
strict -- a ref must PASS one of them, not merely gesture at it.

For each SPARKER ref, also set INFLUENCE (orthogonal to the labels):
\begin{itemize}
\item \texttt{central} -- load-bearing for the paper's argument; removing it would damage the contribution, framing, or method.
\item \texttt{incidental} -- acknowledged but peripheral; paper would survive without it.
\end{itemize}

Provide one short rationale per sparker ref and a list of \texttt{context} quotes
drawn verbatim from the mention snippets or body (one per supporting
passage; usually 1--3).

WORKED EXAMPLES (from past papers; abridged):

Ex. 1 -- exposes\_gap, central (empirical):
Cited work: ``RAT-100, a remote-associates benchmark for creativity (1968).''
Mentions in new paper: ``Frontier models score $>$95\% on RAT-100; the test no
longer discriminates.'' [Experiments] ``Our Table 2 shows RAT-100 saturates
while our new benchmark does not.''
Verdict: SPARKER, exposes\_gap, central. The new paper's experiments
establish the gap even though the cited paper does not declare it.

Ex. 2 -- motivates, central (load-bearing motivation):
Cited work: ``Boden 1991, on combinatorial creativity.''
Mentions: ``We take motivation from Boden's account of combinatorial
creativity.'' (cited once, in the introduction's motivation paragraph.)
Verdict: SPARKER, motivates, central. The paper's stated reason-for-being
traces explicitly to this work -- the test ``would the reader still
understand WHY?'' fails without it.

Ex. 3 -- NOT a sparker (single-sentence formalism mention):
Cited work: ``Kulesza \& Taskar, Determinantal Point Processes.''
Mentions: ``Diverse selection has been studied via DPPs (Kulesza \& Taskar)
and submodular methods (Lin \& Bilmes).'' (one sentence, related work
only, not used in the method.)
Verdict: other\_ids. Fails the foundation test -- the method section would
not need rewriting if these refs didn't exist.

Ex. 4 -- NOT a sparker (thin motivation):
Cited work: ``Chakrabarty et al., on AI creative writing evaluation.''
Mentions: ``Recent work has questioned whether AI can be creative
(Chakrabarty et al.).'' (half-sentence in related work, no follow-up.)
Verdict: other\_ids. Fails the motivates test -- the new paper's
reason-for-being does not depend on this specific cite; removing it
would not change the reader's understanding of why the paper exists.

References:
\{refs\_block\}

COVERAGE RULE: every numbered ref above must appear in EXACTLY ONE place --
either in \texttt{sparkers} (with labels) or in \texttt{other\_ids}. Do not omit any id.

Return ONLY a JSON object of the form:
\begin{verbatim}
{
  "sparkers": [
    {
      "id": 7,
      "context": ["verbatim quote 1", "verbatim quote 2"]
      "rationale": "...",
      "labels": ["foundation", "extends"],
      "influence": "central",
    },
    ...
  ],
  "other_ids": [1, 2, 3, 4, 5, 6, 8, 9, 10, ...]
}
\end{verbatim}

\texttt{labels} must be a non-empty subset of
\texttt{["motivates", "exposes\_gap", "foundation", "extends"]}.
Most refs belong in \texttt{other\_ids}; sparkers are typically a minority.

Read through each section and gather context for each reference. Then think step by step through every (reference:str, context:List[str]) on what role(s) the reference fits
the best.
\end{promptbox}

\begin{promptbox}[Prompt \thetcbcounter: Source-Document Summarization Prompt]
\promptlabel{prompt:source_summ}
System:

You are an expert research scientist. You read a paper (title plus full text or abstract) and distill it into a compact, structured card in a fixed SI schema (problem\_gap / core\_claim / method) plus its empirical outcomes (results / takeaways). The spine (problem\_gap / core\_claim / method) captures the IDEA -- the conceptual approach -- stated independent of how well it turned out; results / takeaways carry what the work actually found. Be faithful to the source: report only what the paper claims, and use the numbers it states. Return only a single JSON object.

User:

Summarize the paper below into a structured card with five fields -- \texttt{problem\_gap}, \texttt{core\_claim}, \texttt{method}, \texttt{results}, \texttt{takeaways}. The spine fields (problem\_gap / core\_claim / method) play DIFFERENT roles; a card whose fields collapse into restatements of each other is not useful. Read the definitions carefully.

\begin{itemize}
\item \texttt{problem\_gap}: the gap this paper addresses -- the open question, tension, or unresolved difficulty in the area that motivates the work. State it as one clear gap: a contradiction, a missing explanation, a scope mismatch, an untested assumption, a disconnect between lines of work, or a bottleneck holding the area back. Name the gap itself, NOT the paper's specific solution to it (1--2 sentences).
\item \texttt{core\_claim}: the paper's load-bearing, non-obvious claim -- the observation, mechanism, or reframing that gives the problem a specific, newly tractable angle and distinguishes this work from others on the same problem. It is a claim about how the world \emph{is} (an asymmetry, a hidden dependency, a failed assumption, a reinterpretation), stated as the idea itself, independent of how well it turned out. WHAT approach/mechanism/hypothesis does the work introduce, and WHY would it work? (1--2 sentences).
\item \texttt{method}: at a high level, HOW the paper turns the claim into research -- the \emph{kind} of move it makes (combining ideas, extending a result to a new setting, making a fragile result robust, deriving something formally, mapping a space empirically, building an artifact) and how that move addresses the gap. Keep it at the level of an approach and setup, not a step-by-step protocol (2--3 sentences).
\item \texttt{results}: the main findings -- key quantitative/qualitative outcomes, comparisons against baselines, notable ablations. Include negative or null results and approaches the paper reports did NOT work, not only the wins. Report the concrete numbers the paper states where available (2--4 sentences). Do NOT report the paper's own untested conjectures as if they were findings -- those belong in takeaways.
\item \texttt{takeaways}: implications -- what worked and what did not, limitations, and what the result enables or rules out going forward. If the paper makes speculative claims or conjectures it does not test, note them here and mark them as speculative (1--3 sentences).
\end{itemize}

IMPORTANT split between the spine and the outcomes:
\begin{itemize}
\item \texttt{problem\_gap}, \texttt{core\_claim}, and \texttt{method} must read the SAME whether or not the idea has been tested. State the gap, the mechanism/claim, and the approach; do NOT put results, metrics, benchmark numbers, or evaluation verbs (``show'', ``achieve'', ``outperform'', ``improves accuracy by X\%'') in them.
\item \texttt{core\_claim} is NOT a prediction awaiting test (that's a hypothesis), NOT a plan of action (that's the method), and NOT a restatement of the gap.
\item Keep \texttt{problem\_gap} broad enough that \texttt{core\_claim} is doing real narrowing work; if the gap already contains the claim's commitments, state the gap more broadly.
\item \texttt{results} and \texttt{takeaways} are where the empirical story and the numbers live.
\end{itemize}

Be concise and specific. If a field is not stated in the source, infer the most likely answer from the rest of the paper; if the source is only an abstract, fill \texttt{results} with whatever outcomes it reports (it may be brief).

=== PAPER ===

Title: \{title\}

Abstract / Description:
\{body\}

Return ONLY a JSON object of the form:
\begin{verbatim}
{
  "problem_gap": "...",
  "core_claim": "...",
  "method": "...",
  "results": "...",
  "takeaways": "..."
}
\end{verbatim}
Do not output anything outside the JSON object.
\end{promptbox}

\begin{promptbox}[Prompt \thetcbcounter: Target Paper Summarization Prompt]
\promptlabel{prompt:target_summ}
System:

You are an expert research scientist. You are given (a) a numbered set of summarized inspiration documents and (b) the full text of a target paper that was written building on those inspirations. Your job is to recover the single core research idea of the target paper and express it in a fixed SI schema (problem\_gap / core\_claim / method / inspiration), grounding it in the numbered inspiration documents you were given -- NOT in the target paper's own bibliography. Be faithful to the target paper: do not invent contributions it does not make. Return only a single JSON object.

User:

You are given a set of inspiration documents (each an already-summarized paper) and the full text of a TARGET PAPER that builds on them. Recover the target paper's SINGLE core research idea and express it as one idea object in the SI schema below. The fields play DIFFERENT roles; a card whose fields collapse into restatements of each other is not useful.

\begin{itemize}
\item \texttt{problem\_gap}: the gap the target paper addresses, drawn from across the inspiration documents -- the open question, tension, or unresolved difficulty that connects them and that none resolves on its own. State it as one clear gap, NOT as the target paper's specific solution to it. BEGIN this field with the single best-fitting OPPORTUNITY PATTERN label in square brackets, chosen from the taxonomy below, then the prose gap (1--2 sentences). Pick the ONE dominant pattern even if several apply. If NONE genuinely fits, use ``[Other]'' rather than forcing a poor match -- a wrong label is worse than ``[Other]''. Example: ``[Bridge Opportunity] Work on X and on Y each \ldots\ but neither connects \ldots''.
\item \texttt{core\_claim}: the couple of sentences the whole idea lives or dies on -- ``we claim that X.'' It must be something that could turn out to be wrong: a definite, testable assertion/hypothesis/insight, not a topic or a direction. Do not just restate the gap, and do not describe what you would do about it. From these sentences alone a reader should know exactly what the target paper is betting on (1--2 sentences).
\item \texttt{method}: at a high level, HOW the claim WOULD BE turned into research -- the \emph{kind} of move (combining ideas, extending to a new setting, making a fragile result robust, deriving formally, mapping a space empirically, building an artifact) and how that move addresses the gap. Write it as a PROPOSED approach, forward-looking, as if the idea has not yet been carried out: use present/infinitive framing (``to test this, X''; ``the approach is to X''), NOT past tense and NOT ``the authors/the paper do X''. Do not narrate what was done or built. Concrete specifics (a setting, a target system, a dataset) may be named, but state them as part of the plan, never as accomplished facts. Keep it at the level of an approach, NOT a step-by-step protocol. BEGIN this field with the single best-fitting METHOD PARADIGM label in square brackets, chosen from the taxonomy below, then the prose approach (2--3 sentences). Pick the ONE dominant paradigm even if several apply. If NONE genuinely fits, use ``[Other]''. Example: ``[Synthesis / Unification] To test this, the approach is to \ldots''.
\item \texttt{inspiration}: prose (three to five sentences) that ARGUES how the idea arises FROM the inspiration documents -- the relationship BETWEEN them, not a per-document restatement. This field must be CRITICAL, not a credit ledger. It must (1) name each cited source's limitation -- for every Document leaned on, say what it got wrong, missed, stopped short of, or left unresolved, so the idea reads as a response to a concrete shortcoming, not a recombination of things that already worked (``Document 3 measured X but only under assumption A''); and (2) justify the non-obvious leap -- argue why connecting these documents is non-trivial, so the idea does not read as the obvious next step anyone would take (if the connection IS routine, say that plainly rather than inflating it). The idea may combine sources (naming which supplied the gap vs.\ the method vs.\ the analogy), resolve a tension between them, reframe or invert what a source assumed, or rest mainly on an idea imported from outside the provided documents, named explicitly (it will carry no Document $\langle N\rangle$). Refer to the numbered documents by their exact ``Document $\langle N\rangle$'' labels wherever the idea genuinely builds on them; do not invent a document link.
\end{itemize}

Write \texttt{problem\_gap}, \texttt{core\_claim}, and \texttt{method} so they read the same whether or not the idea has been tested: state the gap, the claim/bet, and the approach, NOT the outcome. Do NOT report results, metrics, or use evaluation verbs (``show'', ``achieve'', ``outperform''). Do NOT narrate the work as already completed: avoid past tense and completed-action verbs (``designed'', ``built'', ``ran'', ``evaluated'', ``conducted'') and do not attribute actions to ``the authors'' or ``the paper'' -- the card describes a forward-looking idea, not a recounting of what a paper did. Keep \texttt{problem\_gap} broad enough that \texttt{core\_claim} is doing real narrowing work.

TAXONOMY (for the required leading [labels]).
\texttt{problem\_gap} MUST begin with exactly one of these OPPORTUNITY PATTERNS, in square brackets and written verbatim:
\begin{itemize}
\item \texttt{[Puzzle / Contradiction]} -- a paradox, tradeoff, surprising failure, or conflicting evidence.
\item \texttt{[Explanation Gap]} -- a missing causal, mechanistic, theoretical, or explanatory account.
\item \texttt{[Scope Mismatch]} -- unrealistic assumptions, narrow regimes, unclear transferability, or boundary conditions.
\item \texttt{[Evidence Gap]} -- missing ways to observe, measure, benchmark, audit, diagnose, validate, compare, or accumulate evidence.
\item \texttt{[Bridge Opportunity]} -- disconnected literatures, theories, evidence streams, communities, or methods that could be connected.
\item \texttt{[Failure / Risk Gap]} -- brittleness, unreliability, bias, uncertainty, safety/privacy/security risk, or reproducibility concerns.
\item \texttt{[Resource Bottleneck]} -- cost, compute, time, data access, sample scarcity, experimental burden, deployment friction, usability, or scalability.
\end{itemize}
\texttt{method} MUST begin with exactly one of these METHOD PARADIGMS, in square brackets and written verbatim:
\begin{itemize}
\item \texttt{[Synthesis / Unification]} -- bridges, integrates, reconciles, or unifies separate literatures, theories, evidence streams, mechanisms, or methods.
\item \texttt{[Relax / Extend Scope]} -- makes prior work function under weaker assumptions, broader scope, new regimes, noisier conditions, or realistic settings.
\item \texttt{[Robustification]} -- reduces failures, brittleness, risk, uncertainty, bias, unreliability, or trustworthiness problems.
\item \texttt{[Formal Derivation]} -- introduces a formal model, theorem, bound, objective, derivation, proof, ontology, taxonomy, conceptual distinction, or explanatory formulation.
\item \texttt{[Empirical Mapping]} -- builds or applies systematic measurement, benchmarks, diagnostics, datasets, empirical maps, comparative studies, or pattern analyses.
\item \texttt{[Artifact / System]} -- builds a concrete artifact, software system, platform, device, material, prototype, or deployment workflow as the central contribution.
\item \texttt{[Optimization / Search]} -- uses optimization, search, screening, tuning, active/adaptive design, scaling, resource allocation, or efficiency strategies to discover or improve a solution.
\end{itemize}
If, and only if, none of the seven listed labels on an axis genuinely fits, you may use ``[Other]'' for that axis -- do not stretch a listed label to force a fit, and do not invent any label outside this list plus ``[Other]''. Use the bracketed label ONLY at the very start of the field; do not repeat it in the prose or add labels to any other field.

INSPIRATION DOCUMENTS

\{documents\}

TARGET PAPER

Title: \{title\}

Full text:
\{body\}

Return ONLY a JSON object of the form:
\begin{verbatim}
{
  "problem_gap": "[<one opportunity pattern>] ...",
  "core_claim": "...",
  "method": "[<one method paradigm>] ...",
  "inspiration": "Document <N> did X but stopped short of / assumed / never addressed;"
  "Document <M> offered the method yet..; connecting them is non-obvious because.."
  "(name any external source explicitly).."
}
\end{verbatim}
Do not output anything outside the JSON object.
\end{promptbox}

\begin{promptbox}[Prompt \thetcbcounter: Reverse-Engineering (Hindsight) Trace Prompt]
\promptlabel{prompt:trace_synth}
System:

You reconstruct the private reasoning a strong research-ideation model would have produced on its way to a final set of ideas. You write in the first person, as that model thinking aloud before it commits to an answer.

User:

Below is a research-ideation task: a set of numbered source documents, and the FINAL set of \{num\_ideas\} research ideas a model committed to after reading them. Each idea is a card with four fields --- `problem\_gap` (opening with a bracketed OPPORTUNITY PATTERN label), `core\_claim`, `method` (opening with a bracketed METHOD PARADIGM label), and `inspiration` (prose that grounds the idea in the numbered "Document N" sources).

Your job: write the exploratory <think> reasoning trace that PLAUSIBLY LED to exactly this final set. You are reconstructing the messy thinking that happens *before* an answer is committed --- not a polished after-the-fact justification.

\textbf{THE TAGS THE REASONING WORKS TOWARD}

Each idea's `problem\_gap` is stamped with ONE opportunity pattern, and its `method` with ONE method paradigm --- written verbatim:

\textbullet{} OPPORTUNITY PATTERNS: [Puzzle / Contradiction] \textperiodcentered{} [Explanation Gap] \textperiodcentered{} [Scope Mismatch] \textperiodcentered{} [Evidence Gap] \textperiodcentered{} [Bridge Opportunity] \textperiodcentered{} [Failure / Risk Gap] \textperiodcentered{} [Resource Bottleneck]  (or [Other] only if none genuinely fits).

\textbullet{} METHOD PARADIGMS: [Synthesis / Unification] \textperiodcentered{} [Relax / Extend Scope] \textperiodcentered{} [Robustification] \textperiodcentered{} [Formal Derivation] \textperiodcentered{} [Empirical Mapping] \textperiodcentered{} [Artifact / System] \textperiodcentered{} [Optimization / Search]  (or [Other]).

You reason ABOUT which label fits as the idea takes shape --- you do not stamp them cleanly in one confident pass (see HARD CONSTRAINTS).

\textbf{HARD CONSTRAINTS}

\textbullet{} Do NOT reveal, hint, or imply that the final ideas were given to you. Never write "the given ideas", "the target", "as provided", "I was told", etc. The trace must read as genuine discovery, arriving at the ideas on its own.

\textbullet{} The trace must LAND on the final set: by the end you commit to ideas that match the \{num\_ideas\} below in substance (same gaps, same claims, same methods, same grounding). You may reach them in any order and rename freely while reasoning.

\textbullet{} Output ONLY the reasoning trace prose. No <think> tags (they are added later), no <answer> block, no JSON, no headers, no preamble like "Here is".

\textbullet{} Do NOT announce a candidate-generation phase. Never write a gear-shift line like "let me start floating candidates/ideas", "now let me brainstorm", "time to generate ideas", or any sentence whose only job is to declare you're moving from analysis to proposing. Candidates must surface from the analysis itself (a tension or gap you just hit), so the first one reads as falling out of what you were already chewing on --- not as the start of a new section.

\textbullet{} NO performative rejections. NEVER reject a candidate in order to look thorough, and NEVER write a framing line like "let me also consider some ideas I should reject", "to show I explored more broadly", "for completeness, some directions I'm not taking". Every rejection must be of a candidate you were actually weighing --- dropped the moment its own flaw becomes the reason, not assembled afterward as a list of strawmen.

\textbullet{} Do NOT enumerate observations, tensions, or candidates as an ordinal/numbered sequence ("First tension\ldots{}", "Second connection\ldots{}", "Idea 1 \ldots{} Idea 8"), and do NOT keep a running count or write a tally line ("that's exactly N", "I now have N", "let me count what I have"). Connections surface irregularly, out of order, interleaved with re-reading the docs. The final count is something you discover at the end, not a quota you fill.

\textbullet{} OVERSHOOT, then cut or merge. Take MORE real contenders than you keep --- develop two or three beyond your final set far enough to be genuine contenders (a stated gap + a candidate claim), then remove them: reject at least one for a flaw that only became visible once you'd developed it, and/or MERGE two threads you'd been treating separately into one stronger idea. One-line throwaways do NOT count --- a cut only lands if the candidate had earned real consideration first. The final \{num\_ideas\} should feel like what SURVIVED, not what you set out to list.

\textbullet{} RE-TAG at least once. For at least one idea, reach first for an opportunity pattern or method paradigm that turns out not to fit, notice why, and correct it --- e.g. "I keep calling this an [Evidence Gap], but the sources already measure it fine; what's actually missing is the mechanism, so it's an [Explanation Gap]", or "I want [Synthesis / Unification] here, but I'm really just making one fragile result hold under noise --- that's [Robustification]." Let at least one label stay genuinely unsettled while you argue it out; do NOT stamp all labels cleanly in a single confident pass.

\textbullet{} The inspiration grounding must be EARNED, and sometimes come up short. For each surviving idea, work out which specific "Document N"(s) seed it, what each of those documents got wrong / missed / stopped short of, and why combining them is non-obvious (the tension or mismatch that had to be resolved). This must not be a rubber stamp: in this trace, at least one candidate should turn out to lean on a connection that's actually routine (the obvious next step) or that can't be honestly grounded in the numbered documents --- and you then either sharpen its non-obvious leap or drop it. A grounding that always lands cleanly is theater.

\textbullet{} NO after-the-fact audit block. Do NOT end with a batched QA pass that re-checks every pair for distinctness or confirms the tag spread in one sweep. Those checks happen DISTRIBUTED: the moment two ideas threaten to collapse into one gap or one claim, you challenge them right there and resolve it before moving on.

\textbf{THE SHAPE OF THE REASONING (a backbone, NOT a template)}

Let the trace move through roughly these phases. They are the spine of the thinking, NOT labeled sections to fill in --- do not write headers, do not announce "Phase 1". The phases interleave and you revisit earlier ones as later steps force a rethink; that back-and-forth IS the exploration.

Do NOT march through them once in order. A real reasoner jumps around: you start floating a candidate, realize you need to go back and re-read what Document 2 actually claims, that re-reading kills the candidate but suggests a different gap, which makes you reconsider whether an idea you'd already kept is really distinct. Make this non-linearity visible --- at least a couple of times, let a later step send you BACK to re-interrogate a document or revise an earlier judgment, mid-stream, before you've finished sweeping all candidates. Interleave floating and interrogating rather than finishing one cleanly before the other.

1. FRAME the task --- but do NOT open the same way every time. Your opening move for THIS trace: \{entry\_point\} Hold in mind, without writing it as a checklist, what a good set needs: \{num\_ideas\} genuinely distinct ideas spanning multiple problem\_gaps, a mix of opportunity patterns and method paradigms (not all the same two labels), and each core\_claim a real bet that narrows its gap --- every idea grounded in the numbered documents. That's the bar you're working toward, not a quota to announce up front.

2. INTERROGATE the documents. Do NOT summarize or restate them. Instead surface the live material an idea could be built on or pushed against: the specific claims each doc commits to, the tensions or disagreements between them, the assumptions they lean on, the gaps they leave open, what each stops short of. Ask probing questions of the docs and mark what you can actually verify from the text vs. what is a hunch you'd be reaching for. This is where the "Document N" material for later inspiration grounding gets found.

3. FLOAT \& STRESS-TEST candidates. Float several more contenders than will survive --- the overshoot-then-cut/merge is a HARD CONSTRAINT, not optional. Each candidate must EMERGE from the interrogation --- a tension you just noticed, a gap a doc leaves open --- so it falls out of what you were already chewing on (see the no-announcement rule in HARD CONSTRAINTS). For each, state the gap it opens, the claim it bets on, and roughly how you'd pursue it --- then probe it and either keep it or drop it the moment its flaw shows ("this collapses the claim into the gap", "just restates Document 2", "the gap's too narrow to admit a real bet", "I can't honestly ground this in any document"). Never collect rejections into a tidy block (see the no-performative-rejection rule). As you settle a candidate, work out its opportunity-pattern label and its method-paradigm label --- and let at least one be genuinely unsettled or get corrected (see the RE-TAG constraint). When a new candidate lands near one you're already holding, challenge the pair on the spot --- really distinct gaps/claims, or the same lens twice? --- and resolve it THEN, by cutting one or merging the two, rather than saving it for a final sweep. Hedge honestly where the evidence is thin. For EVERY surviving candidate, do the inspiration work in the trace: name the specific "Document N"(s) that seed it, say what each got wrong or stopped short of, and argue why connecting them is non-obvious --- the tension that had to be resolved. Per HARD CONSTRAINTS, at least one candidate's connection should come up short (routine, or ungroundable) and get sharpened or dropped.

4. CONVERGE. The survivors ARE your set: distinctness, tag fit, and grounding were settled as you went (see HARD CONSTRAINTS), not in a closing audit. Land on the final \{num\_ideas\} briefly --- no recap checklist, no pairwise distinctness sweep, no tag-spread tally. Just commit to what made it through.

\textbf{WHAT THE TRACE MUST CONTAIN (exploration-heavy)}

This is the whole point --- make the reasoning genuinely exploratory:

\textbullet{} Overshoot then cut/merge: develop more real contenders than you keep, and remove two or three that had genuine traction --- by reasoned rejection the moment the flaw shows, or by merging two threads into one (HARD CONSTRAINT). No performative reject-lists; the final set is what survived (see step 3 above).

\textbullet{} Show real UNCERTAINTY and hedging where it's honest: "I can't verify this from the documents, but my hunch is Document 4's method carries over because..."; "not sure this is a [Scope Mismatch] or really a [Failure / Risk Gap] --- leaning the latter because..."; "this might be too speculative, but it's the most distinctive bet I have."

\textbullet{} Show VERIFICATION moves: checking a candidate claim against what the docs actually say, testing whether the gap is broad enough to admit a real bet, asking whether two ideas are really distinct or secretly the same gap, checking whether the inspiration honestly points at the numbered documents.

\textbullet{} Show dead-ends and course-corrections --- a thread you start, doubt, and drop or reshape. Let a later step send you back to revise an earlier judgment.

\textbullet{} Voice: terse, first-person, working-notes register. Not an essay. Fragments, self-interruption, and "wait ---" reversals are good. A thread you start, drop mid-sentence, and pick back up two moves later is realistic --- let the structure be a little ragged rather than marching cleanly. A bit of noise is fine; a rigid uniform skeleton is not. Length should reflect real exploration of \{num\_ideas\} ideas: thorough, not padded.

\textbf{SOURCE DOCUMENTS (what the model saw)}

\{documents\}

\textbf{FINAL IDEA SET (what the reasoning must arrive at)}

\{ideas\_digest\}

Now write the exploratory reasoning trace that arrives at these \{num\_ideas\} ideas. Remember: reasoning prose only, no tags, and never let on that the final set was given to you.

\textbf{The \{entry\_point\} opening move} is drawn per instance from one of the following six, so the SFT traces do not all open the same way:

\textbullet{} Open mid-thought on the ONE claim across these documents that nags at you most --- the surprising, suspicious, or under-defended one --- and let the rest of the analysis unspool from chasing why it bothers you.

\textbullet{} Open on a TENSION: two documents that don't sit comfortably together. Start from the friction between them and let candidates fall out of trying to resolve or exploit it.

\textbullet{} Open on a LIMITATION a document admits about itself (or visibly has but doesn't state). Start by taking that gap seriously and seeing what it opens.

\textbullet{} Open skeptically --- your first instinct is that there may be nothing genuinely novel to say here. Start by pressure-testing that doubt against the documents until something forces you to change your mind.

\textbullet{} Open from the FIELD's side: what would a researcher working near this area most want that these documents gesture at but don't deliver? Start from the want and work back into the docs.

\textbullet{} Open on an ASSUMPTION shared across the documents that nobody questions. Start by naming it and asking what happens if it doesn't hold.

\end{promptbox}

\begin{outputbox}(Example SFT Trajectory: Raw Model Completion ($K=3$; inputs omitted))
\promptlabel{example:sft_trace}
\raggedright
\textbf{Raw completion --- $\langle$think$\rangle$ 2760 tok, $\langle$answer$\rangle$ 1326 tok}\\(4pt)
\textless{}think\textgreater{}\\

Starting with what Document 1 admits about itself: the easy-to-hard task curriculum helps meta-learners, but they leave two-phase-vs-more-phases untested and they lock the schedule *direction* to easy-then-hard. They even ran reversed ordering and it hurt. Fine — but that's the whole schedule, not a phase-dependent thing. So the assumption they never question is that the *direction* of emphasis is stationary once you've committed to a curriculum.\\(5pt)
Now hold that next to Document 5. Focal loss does the exact opposite. It permanently *down-weights* easy examples so the model concentrates on hard ones — and it says so explicitly, that easy examples never deserve priority, they just drown training. And Document 2's SupCon has this intrinsic hard-positive/hard-negative mining baked into the gradient structure. So in the loss-weighting literature you have two flatly opposed prescriptions coexisting: curriculum says start easy, focal/hard-mining says always chase hard. And nobody reconciles them.\\(5pt)
That's a real contradiction, not just two methods. Curriculum: down-weight hard early. Focal: up-weight hard always. When is each right? My hunch — and this is what falls out of Document 1 running its curriculum over *training progress* while Document 5 assumes a stationary loss — is that emphasis should be non-monotonic in training progress. Early: down-weight hard (curriculum regime). Late: up-weight hard (focal regime). One schedule that flips.\\(5pt)
Let me see if that can be made concrete. Document 1 already gives me a continuous task hardness score — the set-to-set distinguishability (pairwise Euclidean, Hausdorff, HSIC, min inter-class distance). Call it s\_t. Document 5 gives me the focal exponent form $w_t = (1 - s_t)^{\gamma(\tau)}$. So define a task-level weight $w = (1 - s_t)^{\gamma(\tau)}$ where $\gamma$($\tau$) depends on training progress $\tau$. If $\gamma$ is a fixed positive, that's focal-style hard emphasis. If $\gamma$ is negative early, easy tasks get up-weighted — curriculum. So both prior policies are *limiting cases* of one modulation. The bet: a $\gamma$$(\tau)$ that's negative early and positive late dominates either fixed policy.\\(5pt)
Is this a real bet or am I just wrapping two things in a formula? The falsifiable content is the non-monotonicity claim — that the crossover exists and beats both fixed baselines. That's testable on miniImageNet/tieredImageNet with the same backbones Document 1 used. Good. And the derivation piece — showing fixed-positive $\gamma$ recovers focal and a monotone cutoff recovers curriculum via the gradient contributions — that's the (Formal Derivation) spine. The gap itself is a genuine (Puzzle / Contradiction): two prescriptions coexist with no account of when each is correct.\\(5pt)
Wait — is the method really (Formal Derivation) or is it (Synthesis / Unification)? I'm unifying curriculum and focal after all. But the *core move* isn't just "these are the same family, look" — it's deriving the gradient limits and then fitting $\gamma(\tau)$ empirically against baselines. The unification is a byproduct; the work is the derivation of limiting cases plus the schedule fit. Lean (Formal Derivation). If it were pure synthesis I'd have nothing to test beyond "here's a general form." The non-monotone schedule is the actual bet, and that needs the derivation to show why it should exist. Keep (Formal Derivation).\\(5pt)
Grounding check: Document 1 seeds the distinguishability axis and the easy-to-hard finding (plus the reversed-ordering-hurts result, which is exactly the fixed-direction assumption I'm attacking). Document 5 seeds the exponent and its explicit anti-easy stance. Document 2 is the third leg — intrinsic hard mining as a *fixed structural* property. The non-obvious leap: treating these as the same modulation evaluated at different phases, not competing truths. That's honest — neither literature tried making the focal exponent time-varying along a continuous hardness axis. Good, this one holds.\\(5pt)
Now, pivot to the relation-extraction cluster — Documents 3, 4, 6, and 2 again. Document 4 diagnoses that few-shot relation classification trails humans badly (69–70\% vs \textasciitilde{}92\% on 5-way-1-shot) and names *linguistic diversity of expression* as the core barrier — but proposes no representation remedy, just gestures at "commonsense knowledge or causal reasoning." Document 3 (MTB) pretrains relation encoders using shared entity pairs as the positive signal, deliberately label-free to stay task-agnostic, with the blank-replacement trick to stop entity-identity shortcuts.\\(5pt)
Here's the tension I keep circling: MTB uses at most one weak positive per anchor (statements sharing the same entity pair), and that proxy conflates entity identity with relation. Document 2 solved *exactly* the many-positives problem — pull together all same-class samples — but only for image labels. And Document 6 says class means make great prototypes when embeddings cluster unimodally. So: what if you pretrain relation embeddings with SupCon's multi-positive outside-the-log loss, where positives are *all in-batch statements of the same relation label*, not just co-occurring entity pairs? Then evaluate the embeddings prototypically on FewRel.\\(5pt)
The friction that makes this non-obvious: MTB *deliberately* refused to use relation labels — that was the point, task-agnostic pretraining. SupCon's power comes precisely from labels. So combining them isn't just "plug SupCon in"; it's betting that label-supervised multi-positive pretraining beats label-free distributional pretraining *for prototype quality*. That's a genuine open question, not a routine mashup. If it were just "apply SupCon to text," I'd drop it as obvious. But the specific claim — tighter, more separable relation prototypes closing part of the FewRel human gap — that's a real bet, and it directly addresses Document 4's diagnosed barrier.\\(5pt)
Let me make sure this isn't secretly the same idea as the first one. First idea: task-level loss weighting schedule, images, curriculum-vs-focal. This one: representation pretraining objective, relation text, contrastive-vs-distributional. Different gap, different claim, different subfield. Distinct. Good. The label here — this is a (Bridge Opportunity): three threads (SupCon, MTB, prototypes) each advanced their subfield but were never combined to exploit relation labels for prototype-quality embeddings. Method is (Synthesis / Unification) — fusing SupCon's loss + MTB's encoder/blank trick + prototypical eval on FewRel. That fits cleanly, because here the *whole* contribution is the fusion and testing whether it beats MTB and cross-entropy. Unlike idea 1, there's no derivation of limiting cases — it's genuinely a synthesis.\\(5pt)
Grounding: Document 3 (single-entity-pair positive, label-free by design), Document 2 (many positives but only images), Document 6 (means as prototypes under unimodality, but from episodic non-contrastive training), Document 4 (diagnosis, no remedy). Each stopped short in a way the combination addresses. Earned.\\(5pt)
Now what else is in Document 6 I haven't used? The self-admitted limitation: prototypical networks assume unimodal class distributions, and — this is the interesting part — they *tried* learning per-dimension variances / non-spherical Gaussians and got *no empirical gain*. They report that as a dead end. But they never asked *which* episodes actually violate unimodality. So they can't distinguish "variance modeling is useless" from "variance modeling was applied indiscriminately across episodes that mostly didn't need it." That's an unexplained failure sitting right there.\\(5pt)
And — here's the connection — Document 1's set-to-set distinguishability metrics (Hausdorff, HSIC, min inter-class distance) were built purely to *rank task hardness for scheduling*. But a low minimum inter-class distance is exactly the signature of an episode where a single mean is a bad representative — classes bleeding into each other, multimodal support. So the metric Document 1 uses to decide *what to train on* is exactly the signal that should predict *where Document 6's inductive bias fails at inference*.\\(5pt)
The bet: per-episode prototype failure is predictable from those distinguishability scores, and *only* the low-distinguishability episodes benefit from k-sub-prototypes — which would explain why global variance modeling failed (applied everywhere, helped nowhere) while targeted multimodality helps. That reframes Document 6's dead end as a targeting problem.\\(5pt)
Is this an (Evidence Gap) or an (Explanation Gap)? First instinct: (Evidence Gap) — they didn't measure which episodes violate unimodality. But wait — the documents don't lack a measurement they could easily take; the deeper thing is there's *no account* of the mechanism behind why variance modeling failed. The measurement (correlate hardness with error) is the *tool*, but the gap is explanatory: prototype failures are unexplained, and I'm proposing the mechanism (multimodality, detectable via the metric). So it's an (Explanation Gap), not (Evidence Gap). Correcting that — the missing thing isn't data, it's the account of *why* the unimodal bias breaks and *where*.\\(5pt)
Method: compute Document 1's hardness scores per episode on miniImageNet and FewRel, correlate with prototypical-net error, then compare single- vs k-prototype conditioned on measured multimodality. That's (Empirical Mapping) — mapping the regime where the Bregman-clustering assumption breaks and testing whether the metric is an actionable failure predictor. Not derivation, not a new artifact — it's characterizing a regime.\\(5pt)
Now I have to check this against idea 1, because both lean on Document 1's distinguishability metrics. Are they the same idea wearing two hats? Idea 1 uses the metric as a *training-time curriculum axis* — a knob for loss weighting during meta-training. This one uses the metric as an *inference-time diagnostic* — predicting per-episode prototype failure and deciding representation structure. Training vs inference, weighting vs diagnosis. The whole non-obvious move of this third idea is precisely that it *repurposes* a curriculum instrument as a failure diagnostic — bridging training-time and inference-time reasoning that both papers kept separate. So the shared document is fine; the reuse is the point, and the gaps are genuinely different. Distinct.\\(5pt)
Let me sanity-check whether the third idea's grounding is honest and not routine. Document 6 gives the unimodality assumption + the reported null result on variance modeling (this is the crux — without that null result the idea is weak). Document 1 gives the metrics, used only for scheduling. The leap is that the same metric predicts representation adequacy. That's not the obvious next step for either paper — Document 1 never framed its score as a diagnostic, Document 6 never used any per-episode signal. Non-obvious enough. Earned.\\(5pt)
Did I have anything else in play that's dying? I briefly wanted a fourth thread around Document 4's "commonsense / causal reasoning" future direction — some idea about injecting external knowledge into few-shot relation classification. But that can't be honestly grounded: none of the numbered documents supply a knowledge source, a mechanism, or anything to build the bridge from. It'd be me inventing scaffolding the documents don't provide — pure speculation dressed as grounding. Drop it. The FewRel human-gap concern it was chasing is better served by idea 2's contrastive-prototype route, which *is* grounded. So that thread folds into idea 2 rather than standing alone.\\(5pt)
I also flirted with a batch-size/memory idea off Document 1's and Document 2's shared limitation (both constrained by large-batch/GPU memory — Document 2 mitigates with a memory bank, Document 1 is stuck at batch size 2 for big backbones). Tempting: port SupCon's memory-bank trick to Document 1's per-task loss reweighting to lift the batch-size ceiling. But that's an engineering fix, and the gap collapses — it's "make an existing method cheaper," no real conceptual bet, and the claim would just restate "memory banks help memory." Not a research idea with a falsifiable surprise. Let it go; the more interesting use of Document 1's metric is already spoken for by ideas 1 and 3.\\(5pt)
So what survived: the non-monotone task-level focal-vs-curriculum schedule as a (Puzzle / Contradiction) resolved by (Formal Derivation); the SupCon-plus-MTB-plus-prototype fusion for relation embeddings as a (Bridge Opportunity) via (Synthesis / Unification); and the per-episode prototype-failure diagnostic as an (Explanation Gap) via (Empirical Mapping). Three distinct gaps, three different method paradigms, each anchored in specific documents' stated limitations. Committing to these.\\
\textless{}/think\textgreater{}\\
\textless{}answer\textgreater{}\\
(\\
\{\\
~~"problem\_gap": "(Puzzle / Contradiction) Curriculum-style training prescribes emphasizing easy examples/tasks first, while focal-loss and hard-example-mining approaches prescribe concentrating optimization on hard examples; these opposite prescriptions coexist in the loss-weighting literature with no account of when each is correct during training.",\\
~~"core\_claim": "We claim the optimal emphasis is non-monotonic in training progress: early meta-training benefits from down-weighting hard tasks (curriculum regime) while late training benefits from up-weighting them (focal regime), and a single progress-dependent modulation of a focal-style exponent applied at the task level reduces to both prior policies as limiting cases and outperforms either fixed policy.",\\
~~"method": "(Formal Derivation) The approach is to define a task-level loss weight $w = (1 - s_t)^{\gamma(\tau)}$ where s\_t is a task's inter-category distinguishability score (as in the set-to-set metrics of Document 1) and $\gamma(\tau)$ is a schedule over training progress $\tau$ that is negative early (favoring easy tasks) and positive late (favoring hard tasks). Derive the gradient contributions to show that fixed-positive $\gamma$ recovers focal-style hard emphasis and a monotone easy-to-hard cutoff recovers a curriculum, then fit $\gamma(\tau)$ empirically on few-shot benchmarks against both fixed baselines to test whether the non-monotone schedule dominates.",\\
~~"inspiration": "Document 1 established that easy-to-hard task ordering helps meta-learning and even showed reversed ordering hurts, but it treated the schedule direction as fixed and never asked whether late-stage training should invert the policy. Document 5 built focal loss to permanently down-weight easy examples, implicitly assuming a stationary emphasis and explicitly rejecting the idea that easy examples ever deserve priority. Document 2 showed SupCon's gradient intrinsically mines hard examples but again as a fixed structural property. The non-obvious leap is recognizing these are not competing truths but the same modulation evaluated at different training phases; reconciling them requires borrowing Document 1's continuous distinguishability score as the axis on which Document 5's exponent can be made time-varying, which neither literature attempted."\\
\},\\
\{\\
~~"problem\_gap": "((Bridge Opportunity) Supervised contrastive learning with many same-class positives, relation pretraining from entity-pair co-occurrence, and prototype-based few-shot classifiers each advanced their own subfield, yet few-shot relation classification still trails human performance and none of these threads has been combined to exploit relation labels for prototype-quality embeddings.",\\
~~"core\_claim": "We claim that a supervised contrastive pretraining objective that treats all statements sharing the same relation label as positives (generalizing the single-entity-pair positive of matching-the-blanks) produces relation embeddings whose class means are tighter, more separable prototypes, and that this closes a substantial part of the FewRel human-model gap more than either self-supervised distributional pretraining or cross-entropy fine-tuning alone.",\\
~~"method": "(Synthesis / Unification) The approach is to fuse SupCon's outside-the-log multi-positive loss (Document 2) with the entity-marker relation encoder and blank-replacement trick (Document 3), then evaluate the resulting embeddings under a prototypical-network protocol (Document 6) on FewRel (Document 4). Positives are all in-batch statements of the same relation rather than just co-occurring entity pairs, and we test whether prototype separability and N-way K-shot accuracy improve over MTB and cross-entropy baselines.",\\
~~"inspiration": "Document 3 pretrained relation encoders using only shared entity pairs as the positive signal, an intentionally label-free proxy that conflates surface entity identity with relation and offers at most one weak positive per anchor. Document 2 solved exactly the many-positives problem but only for image labels and never touched relational or textual semantics. Document 6 showed class means make excellent prototypes when embeddings cluster unimodally, yet its embeddings came from episodic cross-entropy-like training, not a contrastive objective engineered for tight clusters. Document 4 diagnosed linguistic diversity as the core barrier but proposed no representation-learning remedy. The friction is that MTB deliberately avoided using relation labels (to stay task-agnostic) while SupCon's power comes precisely from labels; resolving whether label-supervised multi-positive pretraining beats label-free distributional pretraining for prototypes is the non-obvious bet."\\
\},\\
\{\\
~~"problem\_gap": "(Explanation Gap) Prototypical networks assume each class is well summarized by a single mean, and attempts to add per-dimension variance yielded no gains, yet there is no account of which episodes actually violate this unimodality assumption or how to detect them, leaving prototype failures unexplained.",\\
~~"core\_claim": "We claim that per-episode prototype failure is predictable from the same set-to-set distinguishability metrics used to rank task hardness: episodes whose minimum inter-class Hausdorff/HSIC distance is low are the ones where a single mean is a poor representative, and only these episodes benefit from a small number of sub-class prototypes, explaining why global variance modeling failed while targeted multimodality helps.",\\
~~"method": "(Empirical Mapping) The approach is to compute Document 1's set-to-set hardness scores per episode on miniImageNet and FewRel, correlate them with prototypical-network error, and then compare single-prototype versus k-prototype (clustered support) representations conditioned on the measured multimodality. This maps out the regime where the unimodal Bregman-clustering assumption breaks and diagnoses whether hardness metrics are actionable failure predictors rather than only curriculum signals.",\\
~~"inspiration": "Document 6 proved the single-mean prototype is optimal under an exponential-family unimodal assumption and reported that adding per-class variance gave no benefit, but it never characterized which episodes violate unimodality, so it could not distinguish 'variance modeling is useless' from 'variance modeling was applied indiscriminately.' Document 1 built distinguishability metrics (Hausdorff, HSIC, pairwise Euclidean) purely to schedule training difficulty and never repurposed them as diagnostics of representation adequacy at inference. The non-obvious connection is that the metric Document 1 uses to decide what to train on is exactly the signal that should predict where Document 6's inductive bias fails; treating a curriculum instrument as a per-episode failure diagnostic bridges training-time and inference-time reasoning that both papers kept separate."\\
\}\\
)\\
\textless{}/answer\textgreater{}
\end{outputbox}

\begin{promptbox}[Prompt \thetcbcounter: Training / Inference Ideation Prompt ($K{=}$\texttt{num\_ideas})]
\promptlabel{prompt:idea_generation}
You are a research ideation assistant. Given a set of source documents, your job is to generate \texttt{\{num\_ideas\}} distinct research ideas that could inspire a scientist working in or adjacent to the area covered by these documents.

Each idea is a single card in the schema below. The four fields play DIFFERENT roles; a card whose fields collapse into restatements of each other is not useful. Two of the fields must begin with a bracketed \texttt{[label]} chosen from the taxonomy that follows the schema.

\textbf{SCHEMA (one card per idea)}
\begin{itemize}
\item \texttt{problem\_gap}: The gap the idea addresses, drawn from across the source documents -- an open question, tension, or unresolved difficulty that connects them and that none resolves on its own. State it as one clear gap, NOT as your specific solution to it. BEGIN this field with the single best-fitting OPPORTUNITY PATTERN label in square brackets (from the taxonomy below), then the prose gap (1--2 sentences). Pick the ONE dominant pattern even if several apply; if NONE genuinely fits, use ``[Other]'' rather than forcing a poor match. Example: ``[Bridge Opportunity] Work on X and on Y each \ldots\ but neither connects \ldots''.
\item \texttt{core\_claim}: The couple of sentences the whole idea lives or dies on -- ``we claim that X.'' It must be something that could turn out to be wrong: a definite, testable assertion / hypothesis / insight, not a topic or a direction. Do not just restate the gap, and do not describe what you would do about it. From these sentences alone a reader should know exactly what the idea is betting on (1--2 sentences).
\item \texttt{method}: At a high level, HOW the claim would be turned into research -- the \emph{kind} of move (combining ideas, extending to a new setting, making a fragile result robust, deriving formally, mapping a space empirically, building an artifact) and how it addresses the gap. It should be a PROPOSED approach, forward-looking: use present/infinitive framing (``to test this, X''; ``the approach is to X''). Keep it at the level of an approach, NOT a step-by-step protocol. BEGIN this field with the single best-fitting METHOD PARADIGM label in square brackets (from the taxonomy below), then the prose approach (2--3 sentences). Pick the ONE dominant paradigm even if several apply; if NONE genuinely fits, use ``[Other]''. Example: ``[Synthesis / Unification] To test this, the approach is to \ldots''.
\item \texttt{inspiration}: Prose (three to five sentences) that ARGUES how the idea arises FROM the source documents -- the relationship BETWEEN them, not a per-document restatement. This field must be CRITICAL, not a credit ledger. Two things it must do: (1) NAME EACH CITED SOURCE'S LIMITATION -- for every Document you lean on, say what it got wrong, missed, stopped short of, or left unresolved, so the idea reads as a RESPONSE to a concrete shortcoming (``Document 3 measured X but only under assumption A''; ``Document 5 proposed Y yet never addressed Z''); and (2) JUSTIFY THE NON-OBVIOUS LEAP -- argue WHY connecting these documents is non-trivial: the tension, friction, or mismatch that had to be resolved for the pieces to fit, so the idea does not read as the obvious next step. If the connection IS routine, say so plainly rather than inflating it. Refer to the numbered documents by their exact ``Document $\langle$N$\rangle$'' labels wherever the idea genuinely builds on them. If you draw on any idea or analogy from OUTSIDE the provided documents, name that external source explicitly (it carries no Document $\langle$N$\rangle$). Do not invent a document link.
\end{itemize}

Write \texttt{problem\_gap}, \texttt{core\_claim}, and \texttt{method} so they: state the gap, the claim/bet, and the approach, NOT the outcome. Keep \texttt{problem\_gap} broad enough that the \texttt{core\_claim} is doing real narrowing work.

\textbf{TAXONOMY (for the required leading [labels])}

\texttt{problem\_gap} MUST begin with exactly ONE of these OPPORTUNITY PATTERNS, in square brackets and written verbatim:
\begin{itemize}
\item \texttt{[Puzzle / Contradiction]} -- a paradox, tradeoff, surprising failure, or conflicting evidence.
\item \texttt{[Explanation Gap]} -- a missing causal, mechanistic, theoretical, or explanatory account.
\item \texttt{[Scope Mismatch]} -- unrealistic assumptions, narrow regimes, unclear transferability, or boundary conditions.
\item \texttt{[Evidence Gap]} -- missing ways to observe, measure, benchmark, audit, diagnose, validate, compare, or accumulate evidence.
\item \texttt{[Bridge Opportunity]} -- disconnected literatures, theories, evidence streams, communities, or methods that could be connected.
\item \texttt{[Failure / Risk Gap]} -- brittleness, unreliability, bias, uncertainty, safety/privacy/security risk, or reproducibility concerns.
\item \texttt{[Resource Bottleneck]} -- cost, compute, time, data access, sample scarcity, experimental burden, deployment friction, usability, or scalability.
\end{itemize}

\texttt{method} MUST begin with exactly ONE of these METHOD PARADIGMS, in square brackets and written verbatim:
\begin{itemize}
\item \texttt{[Synthesis / Unification]} -- bridges, integrates, reconciles, or unifies separate literatures, theories, evidence streams, mechanisms, or methods.
\item \texttt{[Relax / Extend Scope]} -- makes prior work function under weaker assumptions, broader scope, new regimes, noisier conditions, or realistic settings.
\item \texttt{[Robustification]} -- reduces failures, brittleness, risk, uncertainty, bias, unreliability, or trustworthiness problems.
\item \texttt{[Formal Derivation]} -- introduces a formal model, theorem, bound, objective, derivation, proof, ontology, taxonomy, conceptual distinction, or explanatory formulation.
\item \texttt{[Empirical Mapping]} -- builds or applies systematic measurement, benchmarks, diagnostics, datasets, empirical maps, comparative studies, or pattern analyses.
\item \texttt{[Artifact / System]} -- builds a concrete artifact, software system, platform, device, material, prototype, or deployment workflow as the central contribution.
\item \texttt{[Optimization / Search]} -- uses optimization, search, screening, tuning, active/adaptive design, scaling, resource allocation, or efficiency strategies to discover or improve a solution.
\end{itemize}

If, and only if, none of the seven listed labels on an axis genuinely fits, you may use ``[Other]'' for that axis -- do not stretch a listed label to force a fit, and do not invent any label outside this list plus ``[Other]''. Use the bracketed label ONLY at the very start of the field; do not repeat it in the prose or add labels to \texttt{core\_claim} or \texttt{inspiration}.

\textbf{DIVERSITY BAR (across the \texttt{\{num\_ideas\}} ideas)}

The \texttt{\{num\_ideas\}} ideas together should span:
\begin{itemize}
\item Multiple distinct \texttt{problem\_gap}s (not variants of the same question).
\item A mix of opportunity patterns and method paradigms -- do not produce ideas that all carry the same two [labels].
\item Different \texttt{core\_claim}s -- even when two ideas share a gap, they must offer materially different bets on it.
\end{itemize}

Avoid: paraphrasing the source documents, restating consensus findings as ``claims,'' and cards whose \texttt{core\_claim} collapses into the \texttt{problem\_gap}.

\textbf{ORDERING}

Put your strongest single idea first -- most impactful, best-grounded, most defensible. Each subsequent idea should be as distinct as possible from the ones already listed, but equally impactful (if possible).

\textbf{OUTPUT FORMAT}

Each idea object must have exactly this structure:
\begin{verbatim}
{
  "problem_gap": "[<one opportunity pattern>] ...",
  "core_claim": "...",
  "method": "[<one method paradigm>] ...",
  "inspiration": "... Document <N> did X but stopped short of ...;
    Document <M> offered the method yet ...; connecting them is
    non-obvious because ... (name any external source explicitly) ..."
}
\end{verbatim}

\textbf{SOURCE DOCUMENTS}

\texttt{\{documents\}}

\textbf{TASK}

Generate \texttt{\{num\_ideas\}} research ideas conforming to the schema, taxonomy, diversity bar, and ordering rule above. Think before writing -- sketch the space of possible gaps and claims internally, then reason through them, eliminating ideas that either have already been implemented, or do not make any logical sense. Then commit to \texttt{\{num\_ideas\}} that together cover that space well.

Structure your response in two parts:
\begin{enumerate}
\item \texttt{<think> ... </think>}: Use this block to sketch the space of possible gaps and claims before committing. Inside, work through: what are the main threads or open questions running through the documents; what candidate \texttt{problem\_gap}s could be posed, and which are broad enough to admit multiple distinct \texttt{core\_claim}s; for each promising gap, what \texttt{core\_claim} -- drawn from the sources or a related domain -- gives it a distinctive angle, and what method paradigm would carry it (note the two [labels]); and which \texttt{\{num\_ideas\}} cards together cover the space well and hit the diversity bar. This block is for your reasoning only and will not be parsed as output.
\item \texttt{<answer> ... </answer>}: Place the final JSON list of \texttt{\{num\_ideas\}} idea objects inside this block, and nothing else. The content of \texttt{<answer>} must be a valid JSON list parseable on its own -- no prose, no markdown code fences, no commentary before or after the list, no enclosing object around the list.
\end{enumerate}

Do not produce any text outside of these two blocks.
\end{promptbox}

\begin{promptbox}[Prompt \thetcbcounter: Impact / Quality Judge Prompt (SciJudge, pairwise)]
\promptlabel{prompt:impact}
System:

You are a helpful assistant. You first think about the reasoning process in your mind and then provide the user with the answer.

User:

Today is \texttt{\{today\}}. Below are two research ideas, A and B. Determine which idea is likely to have higher scientific impact (i.e.\ would attract more follow-up work and citations).

Show your reasoning process in \texttt{<reason> </reason>} tags. And return the final answer in \texttt{<answer> </answer>} tags. The final answer should contain only `A' or `B'.

Idea A:
\texttt{\{abstract\_a\}}

Idea B:
\texttt{\{abstract\_b\}}

\end{promptbox}

\begin{promptbox}[Prompt \thetcbcounter: Distance / Similarity Judge Prompt (pairwise)]
\promptlabel{prompt:distance_prompt}
System:

You are an expert research scientist judging how SIMILAR two research ideas are to each other. You are not judging quality, novelty, or impact -- only how much the two ideas overlap in substance. Two ideas can be phrased very differently yet propose the same thing, or share surface wording yet differ in their core contribution. Return only a single JSON object.

User:

Below are two research ideas, IDEA A and IDEA B. Judge how SIMILAR they are to each other, field by field.

Each idea is a card with three fields: a problem gap, a core claim, and a method. The problem gap and the method each OPEN WITH A BRACKETED \texttt{[label]} drawn from a short fixed taxonomy. Those labels are coarse categories -- there are only seven of each -- so two ideas sharing a label is common and means very little on its own. Judge the PROSE that follows the label, not the label itself.

Score three features, each on a 1--5 scale. USE THE FULL SCALE -- every point 1--5 is a distinct judgment; do not default to 1/3/5.

\begin{itemize}
\item \texttt{problem\_gap\_sim} -- do the two ideas target the SAME gap: the same open question, tension, or unresolved difficulty? 1 = unrelated gaps (different questions entirely); 2 = same broad area or task family, but a clearly different gap; 3 = related / adjacent gaps, where the framing or what is missing differs in a way that matters; 4 = nearly the same gap, differing only in scope, domain, or a secondary condition; 5 = exactly the same gap. An identical bracketed label with substantively different prose is at most a 2 -- the label alone is not evidence of similarity.
\item \texttt{core\_claim\_sim} -- do the two ideas make the SAME central claim: the testable assertion the idea lives or dies on? 1 = entirely different claims, no mechanistic overlap; 2 = only a loosely related claim or a generic shared building block (e.g.\ both assume attention matters), not the specific assertion; 3 = a partial overlap, where one is a weaker/narrower version of the other, or they share a premise but not the central assertion; 4 = mostly the same claim, where one adds a condition, generalization, or framing the other lacks; 5 = the same claim -- confirming one would largely confirm the other.
\item \texttt{method\_sim} -- do the two ideas propose the SAME approach for testing the claim: the same paradigm AND the same concrete instrument (what is built, measured, derived, or optimized, and on what)? 1 = unrelated approaches, different paradigm and different instrument; 2 = same broad paradigm but a different instrument, or a shared generic tool (e.g.\ both fine-tune a model) doing different work; 3 = related approaches, with an overlapping instrument but a different design, target, or evidence it would produce; 4 = largely the same approach, differing in scope, dataset, or one step; 5 = the same approach -- the two would be executed nearly identically. An identical bracketed label with substantively different prose is at most a 2.
\end{itemize}

=== IDEA A ===
\texttt{\{idea\_a\}}

=== IDEA B ===
\texttt{\{idea\_b\}}

\textbf{OUTPUT FORMAT}

Return ONLY a JSON object of the form:
\begin{verbatim}
{
  "problem_gap_sim": 1 | 2 | 3 | 4 | 5,
  "core_claim_sim": 1 | 2 | 3 | 4 | 5,
  "method_sim": 1 | 2 | 3 | 4 | 5,
  "rationale": "1-3 sentences citing what specifically does or does not overlap."
}
\end{verbatim}
Do not output anything outside the JSON object.
\end{promptbox}

\begin{promptbox}[Prompt \thetcbcounter: Grounding Judge Prompt (load-bearing head, no source documents)]
\promptlabel{prompt:grounding}
System:

You are an expert research scientist auditing whether a research idea's central insight is LOAD-BEARING on the grounding the author declared. Be skeptical of decorative citations (generic gap / connect / extend language with no concrete finding). Return only a single JSON object.

User:

Below is a research INSIGHT and the GROUNDING the author declared for it (the \texttt{inspiration} prose: which sources, what they are said to support).

Score ONE feature, 1--5. USE THE FULL SCALE.

\begin{itemize}
\item \texttt{load\_bearing} -- would the insight COLLAPSE if the cited documents were removed? (decorative citation test) 1 = decorative: the insight stands completely without the cited docs; 2 = the docs add color but the insight does not depend on them; 3 = the docs support the insight but it could be re-grounded elsewhere; 4 = the insight largely depends on the specific cited findings; 5 = the insight strictly requires the cited docs; remove them and it falls apart.
\end{itemize}

Judge from the text alone: does the declared grounding name concrete findings / methods / results that the insight needs, or is it template rhetoric that would read the same with the citations deleted?

=== INSIGHT ===
\texttt{\{insight\_claim\}}

=== DECLARED GROUNDING ===
\texttt{\{grounding\_block\}}

\textbf{OUTPUT FORMAT}

Return ONLY a JSON object of the form:
\begin{verbatim}
{
  "load_bearing": 1 | 2 | 3 | 4 | 5,
  "rationale": "1-3 sentences on whether the grounding is decorative or load-bearing."
}
\end{verbatim}
\end{promptbox}

\end{document}